\documentclass[conference]{IEEEtran}
\usepackage{times}

\usepackage[numbers]{natbib}
\usepackage{multicol}
\usepackage[bookmarks=true]{hyperref}

\usepackage{graphicx}
\usepackage{subfigure}
\usepackage{wrapfig}
\usepackage{times}
\usepackage{algorithmic}
\usepackage{algorithm}
\usepackage{amsmath}
\usepackage{amssymb}
\usepackage{graphics}
\usepackage{epsfig}
\usepackage{multirow}
\usepackage{wrapfig}
\usepackage{amsthm}
\usepackage{mathtools}
\usepackage{makecell}
\usepackage{bbm}
\usepackage{booktabs}
\usepackage{graphicx}
\usepackage{color,soul}

\usepackage{amsmath}
\usepackage{amssymb}
\usepackage{amsthm}

\newcommand{\SE}{\mathrm{SE}}
\newcommand{\SO}{\mathrm{SO}}

\DeclareMathOperator*{\argmax}{arg\,max}

\begin{document}

% \title{\hl{$\SE(2)$} Equivariant Grasp Learning in Real Time}

\title{Sample Efficient Grasp Learning Using Equivariant Models}

% You will get a Paper-ID when submitting a pdf file to the conference system
\author{\authorblockN{Xupeng Zhu, Dian Wang, Ondrej Biza, Guanang Su, Robin Walters, Robert Platt}
\authorblockA{Khoury College of Computer Sciences\\
Northeastern University,
Boston, Massachusetts, 02115 \\ Email:\texttt{\{zhu.xup, wang.dian, biza.o, su.gu, r.walters,  r.platt\}@northeastern.edu}}}

%\author{\authorblockN{Michael Shell}
%\authorblockA{School of Electrical and\\Computer Engineering\\
%Georgia Institute of Technology\\
%Atlanta, Georgia 30332--0250\\
%Email: mshell@ece.gatech.edu}
%\and
%\authorblockA{Twentieth Century Fox\\
%\and
%\authorblockN{James Kirk\\ and Montgomery Scott}
%\authorblockA{Starfleet Academy\\
%San Francisco, California 96678-2391\\

% avoiding spaces at the end of the author lines is not a problem with
% conference papers because we don't use \thanks or \IEEEmembership

%Homer Simpson\authorrefmark{2},
%Montgomery Scott\authorrefmark{3} and
%Eldon Tyrell\authorrefmark{4}}
%Telephone: (800) 555--1212, Fax: (888) 555--1212}
%\authorblockA{\authorrefmark{4}Tyrell Inc., 123 Replicant Street, Los Angeles, California 90210--4321}}

\maketitle

% \begin{abstract}
% Online $$\SE(2)$$ grasp learning is desired because minimal human effort requirement. The challenge of the learning is sam. Our approach encodes the $$\SE(2)$$ symmetry in the $$\SE(2)$$ grasping, decouples the action into translation and rotation, and balances between exploration and exploitation. The experimental evaluation suggests that our proposal outperforms the baselines in both simulation and on a physical robot. In particular, our model achieves significantly higher grasping success rate, compares to the baselines, in cluttered scene with novel object set after $500$ grasps.
% \end{abstract}

% \begin{abstract}
% Online grasp learning is desired because of the minimal level of human intervention. However, traditional methods are typically not sample efficient enough to learn within a decent amount of grasps. This paper investigates the $$\SE(2)$$ grasping problem and propose a sample-efficient method for solving cluttered grasping. Our method encodes the $$\SE(2)$$ symmetries of the problem, decouples the learning of translational and rotational actions, and balances between exploration and exploitation. The experimental evaluation suggests that our proposal outperforms the baselines in both simulation and on a physical robot. In particular, after only $500$ grasps of training, our model achieves $93$\% success rate in cluttered scene with novel objects.
% \end{abstract}

\begin{abstract}

%Visual grasp detection is a key problem in robotics where the agent must learn to model the grasp function, a mapping from an image of a scene onto a set of feasible grasp poses. In this paper, we recognize that the \hl{$\SE(2)$} grasp function is $\SE(2)$-equivariant and that it can be modeled using an equivariant convolutional neural network. As a result, we are able to significantly improve the sample efficiency of grasp learning to the point where we can learn a good approximation of the \hl{$\SE(2)$} grasp function within only 500 grasp experiences. This is fast enough that we can learn to grasp completely on a physical robot in about an hour. 

In planar grasp detection, the goal is to learn a function from an image of a scene onto a set of feasible grasp poses in $\SE(2)$. In this paper, we recognize that the optimal grasp function is $\SE(2)$-equivariant and can be modeled using an equivariant convolutional neural network. As a result, we are able to significantly improve the sample efficiency of grasp learning, obtaining a good approximation of the grasp function after only 600 grasp attempts. This is few enough that we can learn to grasp completely on a physical robot in about 1.5 hours. Code is available at \url{https://github.com/ZXP-S-works/SE2-equivariant-grasp-learning}.

% Visual grasp detection is a key problem in robotics.  The goal is to learn a grasp function from an image of a scene onto a set of feasible grasp poses. In this paper, we recognize that if the set of grasp poses is \hl{$\SE(2)$}, then the grasp function is $\SE(2)$-equivariant and can be modeled using an equivariant convolutional neural network. As a result, we are able to significantly improve the sample efficiency of grasp learning, obtaining a good approximation of the grasp function within only 600 experiences. This is few enough that we can learn to grasp completely on a physical robot in about an hour. 

% Online grasp learning facilitates proper policy learning with the minimal level of human intervention. However, traditional methods are typically not sample-efficient enough to learn within a decent amount of grasps. This paper investigates the $$\SE(2)$$ grasping problem and proposes a sample-efficient method for solving cluttered grasping. Our method encodes the $$\SE(2)$$ symmetries of the problem, decouples the learning of translational and rotational actions, and balances between exploration and exploitation. The experimental evaluation suggests that our proposal outperforms the baselines both in simulation and on a physical robot. In particular, after only $500$ grasps of training, our model achieves a $92$\% success rate in the cluttered scene with novel objects.
\end{abstract}

\IEEEpeerreviewmaketitle

\section{Introduction}

An important trend in robotic grasping is \emph{grasp detection} where machine learning is used to infer the positions and orientations of good grasps in a scene directly from raw visual input, i.e. raw RGB or depth images. This is in contrast to classical model-based methods that attempt to reconstruct the geometry and pose of objects in a scene and then reason geometrically about how to grasp those objects. 

% In grasp detection, we attempt to learn a neural network model that can infer the position and orientation of grasps in a scene directly from raw visual input, i.e. raw RGB or depth images.

Most current grasp detection models must be trained using large offline datasets. For example, \cite{mousavian20196} trains on a dataset consisting of over 7M simulated grasps, \cite{breyer2021volumetric} trains on over 2M simulated grasps, \cite{mahler2017dex} trains on grasp data drawn from over 6.7M simulated point clouds, and \cite{gpd} trains on over 700k simulated grasps. Some models are trained using datasets obtained via physical robotic grasp interactions. For example,  \cite{supersizing_self_supervision} trains on a dataset created by performing 50k grasp attempts over 700 hours, \cite{qt-opt} trains on over 580k grasp attempts collected over the course of 800 robot hours, and \cite{berscheid2021robot} train on a dataset obtained by performing 27k grasps over 120 hours. 

% Several works~\cite{chalvatzaki2020orientation,3d_generative_res_conv,morrison2018closing} are trained entirely using large on-line datasets like Jacquard~\cite{jacquard} and Cornell~\cite{cornell}. 

Such high data and time requirements motivate the desire for a more sample efficient grasp detection model, i.e. a model that can achieve good performance with a smaller dataset. In this paper, we propose a novel grasp detection strategy that improves sample efficiency significantly by incorporating equivariant structure into the model. Our key observation is that the target grasp function (from images onto grasp poses) is $\SE(2)$-equivariant. That is, rotations and translations of the input image should correspond to the same rotations and translations of the detected grasp poses at the output of the function. In order to encode the prior knowledge that our target function is $\SE(2)$-equivariant, we constrain the layers of our model to respect this symmetry. Compared with conventional grasp detection models that must be trained using tens of thousands of grasp experiences, the equivariant structure we encode into the model enables us to achieve good grasp performance after only a few hundred grasp attempts. 

This paper makes several key contributions. First, we recognize the grasp detection function from images to grasp poses to be $\SO(2)$-equivariant. Then, we propose a neural network model using equivariant layers to encode this property. Finally, we introduce several algorithmic optimizations that enable us to learn to grasp online using a contextual bandit framework. Ultimately, our model is able to learn to grasp well after only approximately 600 grasp trials -- 1.5 hours of robot time. Although the model we propose here is only for 2D grasping (i.e. we only detect top down grasps rather than all six dimensions as in 6-DOF grasp detection), the sample efficiency is still impressive and we believe the concepts can be extended to higher-DOF grasp detection models.

These improvements in sample efficiency are important for several reasons. First, since our model can learn to grasp in only a few hundred grasp trials, it can be trained easily on a physical robotic system. This greatly reduces the need to train on large datasets created in simulation, and it therefore reduces our exposure to the risks associated with bridging the sim2real domain gap -- we can simply do all our training on physical robotic systems. Second, since we are training on a small dataset, it is much easier to learn on-policy rather than off-policy, i.e. we can train using data generated by the policy being learned rather than with a fixed dataset. This focuses learning on areas of state space explored by the policy and makes the resulting policies more robust in those areas. Finally, since we can learn efficiently from a small number of experiences, our policy has the potential to adapt relatively quickly at run time to physical changes in the robot sensors and actuators.

\section{Related Work}

\subsection{Equivariant convolutional layers}

Equivariant convolutional layers incorporate symmetries into the structure of convolutional layers, allowing them to generalize across a symmetry group automatically. This idea was first introduced as G-Convolution~\citep{g_conv} and Steerable CNN~\citep{steerable_cnns}. E2CNN is a generic framework for implementing $\mathrm{E}(2)$ Steerable CNN layers~\citep{e2cnn}. In applications such as dynamics ~\citep{walters2020trajectory,wang2020incorporating} and reinforcement learning~\citep{van2020mdp,mondal2020group,wang2021equivariant,wang2022equivariant} equivariant models demonstrate improvements over traditional approaches. 

% This paper applies the equivariant learning in robotic grasping to substantially decreases the amount of sample required for learning a grasping policy.

\subsection{Sample efficient reinforcement learning}

Recent work has shown that data augmentation using random crops and/or shifts can improve the sample efficiency of standard reinforcement learning algorithms~\citep{rad,DrQ}. It is possible to improve sample efficiency even further by incorporating contrastive learning~\citep{oord2018representation}, e.g. CURL~\cite{curl}. The contrastive loss enables the model to learn an internal latent representation that is invariant to the type of data augmentation used. The FERM framework~\cite{FERM} applies this idea to robotic manipulation and is able to learn to perform simple manipulation tasks dirctly on physical robotic hardware. The equivariant models used in this paper are similar to data augmentation in that the goal is to leverage problem symmetries to accelerate learning. However, whereas data augmentation and contrastive approaches require the model to \emph{learn} an invariant or equivariant encoding, the equivariant model layers used in this paper \emph{enforce} equivariance as a prior encoded in the model. This simplifies the learning task and enables our model to learn faster (see Section~\ref{sect:experiments}).

\subsection{Grasp detection}

In grasp detection, the robot finds grasp configurations directly from visual or depth data. This is in contrast to classical methods which attempt to reconstruct object or scene geometry and then do grasp planning. 

\noindent
\underline{2D Grasping:} Several methods are designed to detect grasps in 2D, i.e. to detect the planar position and orientation of grasps in a scene based on top-down images. A key early example of this was DexNet 2.0, which infers the quality of a grasp centered and aligned with an oriented image patch~\cite{mahler2017dex}. Subsequent work proposed fully convolutional architectures, thereby enabling the model to quickly infer the pose of \emph{all} grasps in a (planar) scene~\cite{morrison2018closing,4Dof,jacquard,3d_generative_res_conv,zhou2018fully} (some of these models infer the $z$ coordinate of the grasp as well).

\noindent
\underline{3D Grasping:} There is much work in 3D grasp detection, i.e. detecting the full 6-DOF position and orientation of grasps based on TSDF or point cloud input. A key early example of this was GPD~\cite{tenpas_ijrr2017} which inferred grasp pose based on point cloud input. Subsequent work has focused on improving grasp candidate generation in order to improve efficiency, accuracy, and coverage~\cite{mousavian20196,sundermeyer2021contact,jiang2021synergies,graspnet-1b,breyer2021volumetric,berscheid2021robot}. 

\noindent
\underline{On-robot Grasp Learning:} Another important trend has been learning to grasp directly from physical robotic grasp experiences. Early examples of this include~\cite{supersizing_self_supervision} who learn to grasp from 50k grasp experiences collected over 700 hours of robot time and~\cite{levine2018learning} who learn a grasp policy from 800k grasp experiences collected over two months. QT-Opt~\citep{qt-opt} learns a grasp policy from 580k grasp experiences collected over 800 hours and~\cite{james2019sim} extend this work by learning from an additional 28k grasp experiences. \cite{song2020grasping} learns a grasp detection model from 8k grasp demonstrations collected via demonstration and~\cite{synergy} learns a online pushing/grasping policy from just 2.5k grasps.

\section{Background}

\subsection{Equivariant Neural Network Models}

% This work focus on learning equivariant grasping policies. We implement this using the equivariant neural networks, i.e., the Steerable CNNs~\citep{}. This section introduces the background of equivariant feature map and equivariant convolution.

% \textbf{$\SO(2)$ Equivariance:} 

% \vspace{0.15cm}
% \noindent
\subsubsection{The cyclic group $C_n \leq \SO(2)$}

In this paper, we are primarily interested in equivariance with respect to the group of planar rotations, $\SO(2)$. However, in practice, in order to make our models computationally tractable, we will use the cyclic subgroup $C_n$ of $\SO(2)$,  $C_n=\{ 2  \pi k / n  : 0\leq k < n\}$. $C_n$ is the group of discrete rotations by multiples of $2\pi/n$ radians.

% \vspace{0.15cm}
% \noindent
\subsubsection{Representation of a group}

Elements of the cyclic group $g \in C_n$ represent rotations.  To specify how these rotations apply to data, we define a \emph{representation} of the group. However, the type of representation needed depends upon the type of data to be rotated. There are two main representations relevant to this paper. The \emph{regular representation} acts on an $n$-dimensional vector $(x_1, x_2, \dots, x_n) \in \mathbb{R}^n$ by permuting its elements: $\rho_{reg}(g)x = (x_{n-m+1}, \dots, x_n, x_1, x_2, \dots, x_{n-m})$ where $g$ is the $m$th element in $C_n$. The \emph{trivial representation} acts on a scalar $x \in \mathbb{R}$ and makes no change at all: $\rho_0(g)x = x$. The \emph{standard representation} $\rho_1$ is a rotation matrix that rotates a vector $x$ in the standard way.

% $\rho_1(g)x = \begin{pmatrix}
% {\scriptstyle \cos{\theta}} & {\scriptstyle -\sin{\theta}} \\
% {\scriptstyle \sin{\theta}} & {\scriptstyle \cos{\theta}}
% \end{pmatrix} x$.

% There are three main representations relevant to this paper. The \emph{standard representation} of $g$ is a matrix $\rho_1(g) = \big(\begin{smallmatrix}
%       \cos g & -\sin g\\
%       \sin g & \cos g
% \end{smallmatrix}\big)$ which rotates a vector in $\mathbb{R}^2$. The \emph{regular representation} acts on an $m$-vector $(x_1, x_2, \dots, x_m) \in \mathbb{R}^m$ by permuting its elements: $\rho_{reg}(g)x = (x_m, x_1, x_2, \dots, x_{m-1})$. Finally, the \emph{trivial representation} acts on a scalar $x \in \mathbb{R}$ and makes no change at all: $\rho_0(g)x = x$.

% An equivariant convolutional layer is associated with a finite group and a group representation. It tracks a separate $m$-channel image for each group element.

% \vspace{0.15cm}
% \noindent
% \subsubsection{$\rho$-feature maps of equivariant convolutional layers} 
\subsubsection{Feature maps of equivariant convolutional layers} 

An equivariant convolutional layer maps between feature maps which transform by specified representations $\rho$ of the group. We add an extra channel to the input and output feature maps which encodes the group transformation on feature space. So, whereas the feature map used by a standard convolutional layer is a tensor $\mathcal{F} \in \mathbb{R}^{m \times h \times w}$, 
% \hl{($m$ chennels, $h$ height and $w$ width)}, 
an equivariant convolutional layer adds an extra dimension: $\mathcal{F} \in \mathbb{R}^{k \times m \times h \times w}$, where $k$ denotes the dimension of the group representation. This tensor associates each pixel $(x,y) \in \mathbb{R}^{h \times w}$ with a matrix $\mathcal{F}(x,y) \in \mathbb{R}^{k \times m}$.

% \vspace{0.15cm}
% \noindent
\subsubsection{Action of the group operator on the feature map} 

Given a feature map $\mathcal{F} \in \mathbb{R}^{k \times m \times h \times w}$ associated with group $G$ and representation $\rho$, a group element $g \in G$ acts on $\mathcal{F}$ via:
\begin{equation}
(g \mathcal{F})(x) = \rho(g) \mathcal{F} (\rho_1(g)^{-1} x),
\label{eqn:group_operator}
\end{equation} 
where $x \in \mathbb{R}^2$ denotes pixel position. In the above, $\rho_1(g)^{-1} x$ is the coordinates of pixel rotated by $g^{-1}$ and $\mathcal{F} (\rho_1(g)^{-1} x) \in \mathbb{R}^{k \times m}$ is the matrix associated with pixel $\rho_1(g)^{-1} x$. The element $g$ operates on $\mathcal{F} (\rho_1(g)^{-1} x)$ via the representation $\rho$ associated with the feature map. For example, if $\rho = \rho_0$ (the trivial representation), then $k=1$ and $g$ acts on $\mathcal{F}$ by rotating the image but leaving the $m$-vector associated with each pixel unchanged. In contrast, if $\rho = \rho_{reg}$ (the regular representation), then $k=|G|$ (the number of elements in $G$) and $g$ acts on $\mathcal{F}$ by performing a circular shift on the group dimension of $\mathcal{F} (\rho_1(g)^{-1} x)$. This last action (by the regular representation) is the one primarily associated with the hidden equivariant convolutional layers used in this paper.

% \vspace{0.15cm}
% \noindent
\subsubsection{The equivariant convolutional layer} 

An equivariant convolutional layer is a function $h$ from $\mathcal{F}_{in}$ to $\mathcal{F}_{out}$ that is constrained to represent only equivariant functions with respect to a chosen group $G$. The feature maps $\mathcal{F}_{in}$ and $\mathcal{F}_{out}$ are associated with representations $\rho_{in}$ and $\rho_{out}$ acting on feature spaces $\mathbb{R}^{k_{in}}$ and $\mathbb{R}^{k_{out}}$ respectively. 
%is associated with either  with the trivial representation or the regular representation: $\mathcal{F}_{in}$ is associated with $\rho_{in} \in \{\rho_0, \rho_{reg}\}$ and $\mathcal{F}_{out}$ is associated with $\rho_{out} \in \{\rho_0, \rho_{reg}\}$. 
Then the equivariant constraint for $h$ is~\cite{equi_theory}:
\begin{equation}
\label{eqn:equi_conv}
h(g \mathcal{F}_{in}) = g h(\mathcal{F}_{in}) = g \mathcal{F}_{out}.
\end{equation}
This constraint can be implemented by tying kernel weights $K(y) \in \mathbb{R}^{k_{out} \times k_{in}}$ in such a way as to satisfy the following constraint~\citep{equi_theory}:
\begin{equation}
K(gy) = \rho_{out}(g) K(y)\rho_{in}(g)^{-1}.
\end{equation}

\subsection{Augmented State Representation (ASR)} 
\label{sect:asr}

We will formulate $\SE(2)$ robotic grasping as the problem of learning a function from an $m$ channel image, $s \in S = \mathbb{R}^{m \times h \times w}$, to a gripper pose $a \in A = \SE(2)$ from which an object may be grasped. Since we will use the contextual bandit framework, we need to be able to represent the $Q$-function, $Q : \mathbb{R}^{m \times h \times w} \times \SE(2) \rightarrow \mathbb{R}$. However, since this is difficult to do using a single neural network, we will use the Augmented State Representation (ASR)~\citep{sharma2017learning,asr} to model $Q$ as a pair of functions, $Q_1$ and $Q_2$. 

We factor $\SE(2) = \mathbb{R}^2 \times \SO(2)$ into a translational component $X \subseteq \mathbb{R}^2$ and a rotational component $\Theta \subseteq \SO(2)$. The first function is a mapping $Q_1 : \mathbb{R}^{m \times h \times w} \times X \rightarrow \mathbb{R}$ which maps from the image $s$ and the translational component of action $X$ onto value. This function is defined to be: $Q_1(s,x) = \max_{\theta \in \Theta} Q(s,(x,\theta))$. The second function is a mapping $Q_2 : \mathbb{R}^{m \times h' \times w'} \times \Theta \rightarrow \mathbb{R}$ with $h' \leq h$ and $w' \leq w$ which maps from an image patch and an orientation onto value. This function takes as input a cropped version of $s$ centered on a position $x$, $\mathrm{crop}(s,x)$, and an orientation, $\theta$, and outputs the corresponding $Q$ value: $Q_2(\mathrm{crop}(s,x),\theta) = Q(s,(x,\theta))$.

\begin{figure}
    \centering
    \includegraphics[width=0.45\textwidth]{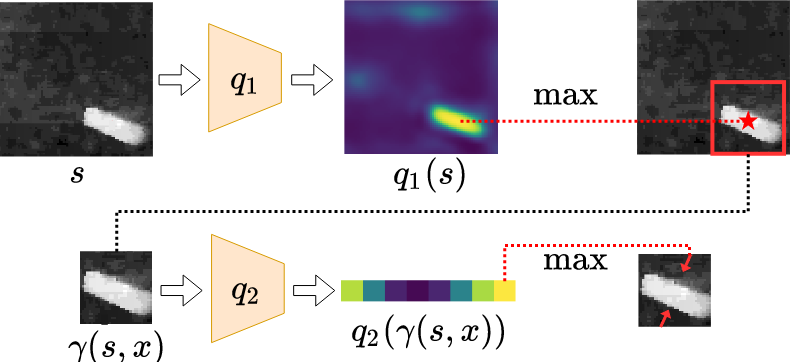}
    \caption{Illustration of the ASR representation. $Q_1$ selects the translational component of an action, $Q_2$ selects the rotational component.}
    \label{fig:asr}
\end{figure}

Inference is performed on the model by evaluating $x^* = \argmax_{x \in X} Q_1(s,x)$ first and then evaluating $Q_2(\mathrm{crop}(s,x^*),\theta)$. Since each of these two models, $Q_1$ and $Q_2$, are significantly smaller than $Q$ would be, inference is much faster. Figure~\ref{fig:asr} shows an illustration of this process. The top of the figure shows the action of $Q_1$ while the bottom shows $Q_2$. Notice that the semantics of $Q_2$ imply that the $\theta$ depends only on $\mathrm{crop}(s,x)$, a local neighborhood of $x$, rather than on the entire scene. This assumption is generally true for grasping because grasp orientation typically depends only on the object geometry near the target grasp point.

\section{Approach}
\label{sect:method}

\subsection{Problem Statement}

In \emph{planar grasp detection}, the goal is to estimate a grasp function $\Gamma : \mathbb{R}^{m \times h \times w} \rightarrow \SE(2)$ that maps from a top down image of a scene containing graspable objects, $s \in S = \mathbb{R}^{m \times h \times w}$, to a planar gripper pose, $a \in A = \SE(2)$, from which an object can be grasped. This is similar to the formulations used by~\cite{mahler2017dex,morrison2018closing}.

\subsection{Formulation as a Contextual Bandit}
\label{sect:bandit}

We formulate grasp learning as a contextual bandit problem where state is the image $s \in S = \mathbb{R}^{m \times h \times w}$ and the action $a \in A = \SE(2)$ is a grasp pose to which the robot hand will be moved and a grasp attempted, expressed in the reference frame of the image. After each grasp attempt, the agent receives a binary reward $R$ drawn from a Bernoulli distribution with unknown probability $r(s,a)$. The true $Q$ function denotes the expected reward of taking action $a$ from $s$. Since $R$ is binary, we have that $Q(s,a) = r(s,a)$. This formulation of grasp learning as a bandit problem is similar to that used by, e.g.~\cite{danielczuk2020exploratory,qt-opt,synergy}.

\subsection{Invariance Assumption}

We assume that the (unknown) reward function $r(s,a)$ that denotes the probability of a successful grasp is invariant to translations and rotations $g \in \SE(2)$. For an image $s \in S = \mathbb{R}^{m \times h \times w}$, let $gs$ denote the the image $s$ translated and rotated by $g$. Similarly, for an action $a \in A = \SE(2)$, let $ga$ denote the action translated and rotated by $g$. Therefore, our assumption is:
\begin{equation}
r(s,a) = r(gs,ga).
\label{eqn:grasp_invariance_assumption}
\end{equation}
Notice that for grasp detection, this assumption is always satisfied because of the way we framed the grasp problem: when the image of a scene transforms, the grasp poses (located with respect to the image) also transform.

\subsection{Equivariant Learning}
\label{sec:equi_learning}

% Since our agent is only rewarded for achieving a successful grasp, the grasp equivariance assumption (Assumption~\ref{assumption:grasp_equivariance}) ensures that both the expected reward function and the optimal $Q$ function are invariant with the group operator $g$: $r(s,a) = r(gs,ga)$ and $Q^*(s,a) = Q^*(gs,ga)$. 

% \vspace{0.15cm}
% \noindent
\subsubsection{Invariance properties of $Q_1$ and $Q_2$} 

Since we have assumed that the reward function $r$ is invariant to transformations $g \in \SE(2)$, the immediate implication is that the optimal $Q$ function is also invariant in the same way, $Q(s,a) = Q(gs,ga)$.
% Under the assumption of Equation~\ref{eqn:grasp_invariance_assumption}, the reward $r$ is invariant to transformations $g \in \SE(2)$. The immediate implication is that $Q$ is also invariant in the same way, $Q(s,a) = Q(gs,ga)$. 
In the context of the augmented state representation (ASR, see Section~\ref{sect:asr}), this implies separate invariance properties for $Q_1$ and $Q_2$:
\begin{align}
\label{eqn:q1_inv}
Q_1(gs, gx) &= Q_1(s, x) \\
\label{eqn:q2_inv}
Q_2(g_\theta (\mathrm{crop}(s, x)), g_\theta +\theta) &= Q_2(\mathrm{crop}(s, x), \theta),
\end{align}
% \begin{equation}
% \label{eqn:q1_inv}
% Q_1(gs, gx) = Q_1(s, x)
% \end{equation}
% \begin{equation}
% \label{eqn:q2_inv}
% Q_2(g_\theta (\mathrm{crop}(s, x)), g_\theta +\theta) = Q_2(\mathrm{crop}(s, x), \theta),
% \end{equation}
where $g_\theta \in \SO(2)$ denotes the rotational component of $g\in \SE(2)$, $gx$ denotes the rotated and translated vector $x \in \mathbb{R}^2$, and $g_\theta (\mathrm{crop}(s, x))$ denotes the cropped image rotated by $g_\theta$.

% where $g_\theta \in \SO(2)$ denotes the rotational component of $g\in \SE(2)$. In the equations above, $gs$ denotes the rotated and translated image $s$, $gx$ denotes the rotated and translated vector $x \in \mathbb{R}^2$, and $g_\theta (\mathrm{crop}(s, x))$ denotes the cropped image rotated by $g_\theta$. 

% Assumption~\ref{assumption:goal_inv} implies that the optimal $Q$-function is $\SE(2)$ invariant:
% \begin{equation}
% \label{eqn:q_inv}
% Q^*(s, a) = Q^*(gs, ga)
% \end{equation}
% because the optimal $Q$-function is identical to the reward function: $Q^*(s, a)=R(s, a)$. The main contribution of this paper is to implement equivariant grasping learning based on  Equation~\ref{eqn:q_inv} to improve the sample efficiency. With the $Q$-function decomposition in Equation~\ref{eqn:Q1} and~\ref{eqn:Q2}, Equation~\ref{eqn:q_inv} becomes two separate invariant properties:
% \begin{equation}
% \label{eqn:q1_inv}
% Q^*_1(gI, gx) = Q^*_1(I, x)
% \end{equation}
% and
% \begin{equation}
% \label{eqn:q2_inv}
% Q^*_2(\gamma(gI, gx), g \theta) = Q^*_2(g_\theta \gamma (I, x), g_\theta +\theta) = Q^*_2(\gamma(I, x), \theta)
% \end{equation}
% where $g_\theta\in\SO(2)$ is the rotational component of $g\in \SE(2)$. $g_\theta$ acts on $\gamma(I, x)$ through image rotation, and $g_\theta$ acts on $\theta$ through adding the rotational component of $g$ to $\theta$. 

% \vspace{0.15cm}
% \noindent
\subsubsection{Discrete Approximation of $\SE(2)$} 

To more practically implement the invariance constraints of Equation~\ref{eqn:q1_inv} and~\ref{eqn:q2_inv} using neural networks, we use a discrete approximation to $\SE(2)$.  We constrain the positional component of the action to be a discrete pair of positive integers $x \in \{1\dots h\}\times  \{1\dots w\}\subset \mathbb{Z}^2$, corresponding to a pixel in $s$, and constrain the rotational component of the action to be an element of the finite cyclic group $C_n = \{2\pi k/n : 0\leq k < n, i\in \mathbb{Z}\}$. This discretized action space will be written $\hat{\SE}(2) = \mathbb{Z}^2 \times C_n$. (Note that while $\mathbb{Z}^2$ and $C_n$ are subgroups of $\SE(2)$, the set $\hat{\SE}(2)$ is not; it is just a subsampling of elements from $\SE(2)$.)

% \vspace{0.15cm}
% \noindent
\subsubsection{Equivariant $Q$-Learning with ASR}

In $Q$-Learning with ASR, we approximate $Q_1$ and $Q_2$ using neural networks. We model $Q_1$ as a fully convolutional UNet~\citep{u_net} $q_1: \mathbb{R}^{m \times h\times w}\to \mathbb{R}^{1\times h\times w}$ that takes as input the state image and outputs a $Q$-map that assigns each pixel in the input a $Q$ value. We model $Q_2$ as a standard convolutional network $q_2: \mathbb{R}^{m \times h'\times w'}\to \mathbb{R}^n$ that takes the $h' \times w'$ image patch as input and outputs an $n$-vector of $Q$ values over $C_n$. The networks $q_1$ and $q_2$ thus represent the functions $Q_1$ and $Q_2$ by partially evaluating at the first argument and returning a function in the second.  As a result, the invariance properties of Equation~\ref{eqn:q1_inv} and~\ref{eqn:q2_inv} for $Q_1$ and $Q_2$ imply $q_1$ and $q_2$ are equivariant:
\begin{align}
\label{eqn:q1_net_equi}
q_1(gs) &= g q_1(s) \\
\label{eqn:q2_net_equi}
q_2(g_\theta \mathrm{crop}(s,x)) &= \rho_{reg}(g_\theta) q_2(\mathrm{crop}(s,x))
\end{align}
% \begin{equation}
% \label{eqn:q1_net_equi}
% q_1(gs) = g q_1(s)
% \end{equation}
% \begin{equation}
% \label{eqn:q2_net_equi}
% q_2(g_\theta \mathrm{crop}(s,x)) = \rho_{reg}(g_\theta) q_2(\mathrm{crop}(s,x))
% \end{equation}
where $g\in \hat{\SE}(2)$ acts on the output of $q_1$ through rotating the $Q$-map, and $g_\theta\in C_n$ acts on the output of $q_2$ by performing a circular shift of the output $Q$ values via the regular representation $\rho_{reg}$.

\begin{figure}
    \centering
    \subfigure[Illustration of Equation~\ref{eqn:q1_net_equi}]{\includegraphics[width=0.35\textwidth]{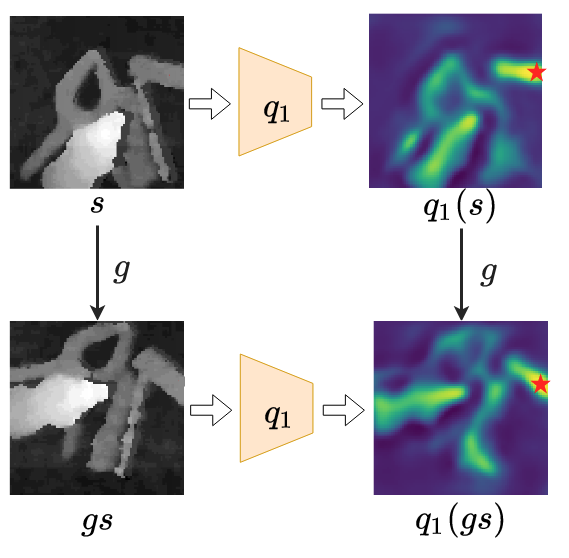}}
    \subfigure[Illustration of Equation~\ref{eqn:q2_net_equi}.]{\includegraphics[width=0.35\textwidth]{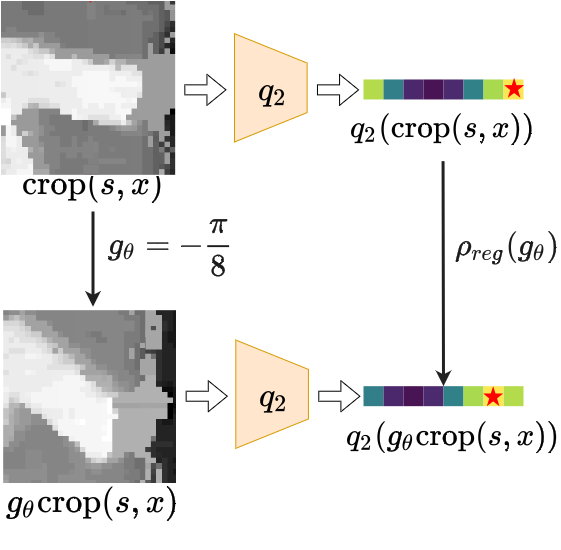}}
    \caption{Equivariance relations expressed by Equation~\ref{eqn:q1_net_equi} and Equation~\ref{eqn:q2_net_equi}.}
    \label{fig:q1_q2_net_equi}
\end{figure}

This is illustrated in Figure~\ref{fig:q1_q2_net_equi}. In Figure~\ref{fig:q1_q2_net_equi}a we are given the depth image $s$ in the upper left corner. If we rotate this image by $g$ (lower left of Figure~\ref{fig:q1_q2_net_equi}a) and then evaluate $q_1$, we arrive at $q_1(gs)$. This corresponds to the LHS of Equation~\ref{eqn:q1_net_equi}. However, because $q_1$ is an equivariant function, we can calculate the same result by first evaluating $q_1(s)$ and \emph{then} applying the rotation $g$ (RHS of Equation~\ref{eqn:q1_net_equi}). Figure~\ref{fig:q1_q2_net_equi}b illustrates the same concept for Equation~\ref{eqn:q2_net_equi}. Here, the network takes the image patch $\mathrm{crop}(s,x)$ as input. If we rotate the image patch by $g_\theta$ and then evaluate $q_2$, we obtain the LHS of Equation~\ref{eqn:q2_net_equi}, $q_2(g_\theta \mathrm{crop}(s,x))$. However, because $q_2$ is equivariant, we can obtain the same result by evaluating $q_2(\mathrm{crop}(s,x))$ and circular shifting the resulting vector to denote the change in orientation by one group element.

% \begin{figure}
%     \centering
%     \subfigure[]{\includegraphics[width=0.45\textwidth]{figure/equation7.png}}
%     \subfigure[]{\includegraphics[width=0.47\textwidth]{figure/equation8.png}}
%     \caption{Figures for equation\ref{} and equation\ref{}.}
%     \label{fig:my_label}
% \end{figure}

% Considering the fact that the $Q$-network is often modeled with state as input and the $Q$-values for all actions as output, we must define the neural networks for modeling $Q_1$ and $Q_2$. We model $Q_1$ using a fully convolutional network (FCN)~\citep{u_net} $q_1: \mathbb{R}^{n\times h\times w}\to \mathbb{R}^{1\times h\times w}$ that takes in the state image and outputs a $Q$-map with the same spatial size as the input, representing the $Q_1$ value for each pixel. We model $Q_2$ using a standard CNN $q_2: \mathbb{R}^{n\times h'\times w'}\to \mathbb{R}^m$ that takes in an image patch and outputs the $Q_2$ value for all $\theta\in C_m$. Then the invariant properties of Equation~\ref{eqn:q1_inv} and~\ref{eqn:q2_inv} become equivariant properties:
% \begin{equation}
% \label{eqn:q1_net_equi}
% q_1(gI) = gq_1(I)
% \end{equation}
% \begin{equation}
% \label{eqn:q2_net_equi}
% q_2(g_\theta \gamma(I, x)) = \pi(g_\theta)q_2(\gamma(I, x))
% \end{equation}
% where $g\in \hat{\SE}(2)$ acts on the output of $q_1$ through rotating the $Q$-map, $g_\theta\in C_m$ acts on the output of $q_2$ through circularly permutation. 

% \vspace{0.15cm}
% \noindent
\subsubsection{Model Architecture of Equivariant $q_1$} 

As a fully convolutional network, $q_1$ inherits the translational equivariance property of standard convolutional layers. The challenge is to encode rotational equivariance so as to satisfy Equation~\ref{eqn:q1_net_equi}. We accomplish this using equivariant convolutional layers that satisfy the equivariance constraint of Equation~\ref{eqn:equi_conv} 
% for the cyclic group $C_u$
where we assign $\mathcal{F}_{in}=s\in{\mathbb{R}^{1\times m\times h\times w}}$ to encode the input state $s$ and $\mathcal{F}_{out} \in \mathbb{R}^{1\times 1\times h\times w}$ to encode the output $Q$ map. Both feature maps are associated with the trivial representation $\rho_0$ such that the rotation $g$ operates on these feature maps by rotating pixels without changing their values. We use the regular representation $\rho_{reg}$ for the hidden layers of the network to encode more comprehensive information in the intermediate layers. We found we achieved the best results when we defined $q_1$ using the dihedral group $D_4$ which expresses the group generated by rotations of multiples of $\pi/2$ in combination with horizontal and vertical reflections.

% reflections over the lines $y=0$, $x=0$, or $y=\pm x$ \hl{in the coordinate with the origin at the center of $s$}.

% We empirically achieve the best performance when defining $q_1$ in the Dihedral group $D_4$ that encodes 4 rotations every 90 degrees and reflection.

% TALK ABOUT INTERMEDIATE LAYERS USING THE REGULAR REPRESENTATION?

% Since the Fully Convolutional Network of $q_1$ is already translational equivariant, we only need to encode the rotational equivariant property using the equivariant network defined in the cyclic group $C_N$. We implement Equation~\ref{eqn:q1_net_equi} using equivariant convolutional layers as in Equation~\ref{eqn:equi_conv} with an input $\mathcal{F}_{in}=I\in{\mathbb{R}^{1\times n\times h\times w}}$ and an output $\mathcal{F}_{out} \in \mathbb{R}^{1\times 1\times h\times w}$, both in the trivial representations ($g$ acts on the input and output through rotating the pixels exclusively). The number of rotations in the cyclic group $C_N$ for $q_1$ can be an arbitrary number. 

% \vspace{0.15cm}
% \noindent
\subsubsection{Model Architecture of Equivariant $q_2$} 

Whereas the equivariance constraint in Equation~\ref{eqn:q1_net_equi} is over $\hat{\SE}(2)$, the constraint in Equation~\ref{eqn:q2_net_equi} is over $C_n$ only. We implement Equation~\ref{eqn:q2_net_equi} using Equation~\ref{eqn:equi_conv} with an input of $\mathcal{F}_{in} = \mathrm{crop}(s, x) \in \mathbb{R}^{1\times m\times h'\times w'}$ as a trivial representation, and an output of $\mathcal{F}_{out} \in \mathbb{R}^{n\times1\times 1\times 1}$ as a regular representation. $q_2$ is defined in terms of the group $C_n$, assuming the rotations in the action space are defined to be multiples of $2\pi/n$.  

%where the number of rotations in $C_{n}$ must match the number of rotations in the action space.

% \vspace{0.15cm}
% \noindent
\subsubsection{$q_2$ Symmetry Expressed as a Quotient Group} 

It turns out that additional symmetries exist when the gripper has a bilateral symmetry. In particular, it is often the case that rotating a grasp pose by $\pi$ radians about its forward axis does not affect the probability of grasp success, i.e. $r$ is invariant to rotations of the action by $\pi$ radians. When this symmetry is present, we can model it using the quotient group $C_n / C_2 \cong \{ 2 \pi k /n : 0\leq k < n/2, k\in \mathbb{Z}, 0 \equiv \pi \}$ which pairs orientations separated by $\pi$ radians into the same equivalence class.

% Since the parallel jaw gripper is symmetric when rotated by \hl{$\pi$}, we introduce additional structure by replacing $C_{n}$ with the quotient group \hl{$C_n / C_2 = \{\frac{i\times \pi}{n}|0\leq i < n, i\in \mathbb{Z}\}$}.

% We implement Equation~\ref{eqn:q2_net_equi} using Equation~\ref{eqn:equi_conv} with an input of $\mathcal{F}_{in} = \gamma (I, x)\in \mathbb{R}^{1\times n\times h'\times w'}$ as a trivial representation, and an output of $\mathcal{F}_{out} \in \mathbb{R}^{m\times1\times 1\times 1}$ as a regular representation. The number of rotations in $C_N$ for $q_2$ must match the number of rotations in the action space, i.e., $N=m$. Moreover, since the parallel jaw gripper is symmetric when rotated by $\frac{\pi}{2}$, we in practice use the quotient group $C_m / C_2 = \{\frac{i\times 2\pi}{m}|0\leq i < m/2, i\in \mathbb{Z}\}$ to encode the gripper symmetry for $q_2$ as in~\cite{wang2021equivariant}. 

% \subsection{Improving Sample Efficiency in Online Grasping}

% \subsection{Formulating the Contextual Bandit Problem}

\subsection{Other Optimizations}

\label{sec:other_optimizations}

While our use of equivariant models to encode the $Q$ function is responsible for most of our gains in sample efficiency (Section~\ref{sect:ablations}), there are several additional algorithmic details that, taken together, have a meaningful impact on performance.

% \vspace{0.15cm}
% \noindent
\subsubsection{Loss Function}

In the standard ASR loss function, both $q_1$ and $q_2$ have a Monte Carlo target, i.e. the target is set equal to the transition reward~\citep{asr}:
\begin{align}
    \label{eqn:asr}
    \mathcal{L} &= \mathcal{L}_{1} + \mathcal{L}_{2} \\ 
    \label{eqn:asrl1}
    \mathcal{L}_{1} &= \tfrac{1}{2}(Q_1(s, x) - r)^2 \\
    \mathcal{L}_{2} &= \tfrac{1}{2}(Q_2(\mathrm{crop}(s, x), \theta) - r)^2.
\end{align}
However, in order to reduce variance resulting from sparse binary rewards in our bandit formulation, we modify $\mathcal{L}_1$:
% \begin{equation}
% \mathcal{L}_1' = \frac{1}{2}\left(Q_1(s, x) - \left(r + (1-r) \max_{\theta\in \bar\Theta} \left[ Q_2(\mathrm{crop}(s, x), \theta) \right] \right)  \right)^2,
% \end{equation}
\begin{equation}
\mathcal{L}_1' = \frac{1}{2}(Q_1(s, x) - (r + (1-r) \max_{\theta\in \bar\Theta} \left[ Q_2(\mathrm{crop}(s, x), \theta) \right] )  )^2,
\end{equation}
where $\bar{\Theta} = \{\bar{\theta} \neq\theta | \forall \bar{\theta}\in C_n/C_2\}$. For a positive sample ($r=1$), the $Q_1$ target will simply be 1, as it was in Equation~\ref{eqn:asrl1}. However, for a negative sample ($r=0$), we use a TD target calculated by maximizing $Q_2$ over $\theta$, but where we exclude the failed $\theta$ action component from $\bar{\Theta}$.

% over the $Q_2$ action component (but excluding the failed $\theta$ action component from $\bar{\Theta}$). 

In addition to the above, we add an off-policy loss term $\mathcal{L}_1''$ that is evaluated with respect to an additional $k$ grasp positions $\bar{X} \subset X$ sampled using a Boltzmann distribution from $q_1(s)$: 
\begin{equation}
\label{eqn:q1_offline}
\mathcal{L}_1'' = \frac{1}{k}\sum_{x_i\in \bar{X}} \frac{1}{2}\left(Q_1(s, x_i) -  \max_{\theta\in\Theta} \left[ Q_2(\mathrm{crop}(s, x_i), \theta) \right] \right)^2,
\end{equation}
where $q_2$ provide targets to train $q_1$. This off-policy loss minimizes the gap between $q_1$ and $q_2$. Our combined loss function is therefore $\mathcal{L} = \mathcal{L}_{1}' + \mathcal{L}_{1}'' + \mathcal{L}_{2}$.

\subsubsection{Prioritizing failure experiences in minibatch sampling}

In the contextual bandit setting, we want to avoid the situation where the agent selects the same incorrect action several times in a row. This can happen because when a grasp fails, the depth image of the scene does not change and therefore the $Q$ map changes very little. We address this problem by ensuring that following a failed grasp experience, that the failed grasp is included in the sampled minibatch on the next SGD step~\cite{synergy}, thereby changing the $Q$ function prior to reevaluating it on the next time step. This reduces the chance that the same (bad) action will be selected.

% \subsubsection{Prioritizing Online Failure}\label{sec:prioritizing_failure}
% One challenge of learning with few online transition is to prevent the agent from executing receptive sub-optimal actions, which wastes the transition to gather valueless data. To avoid this scenario, we force the agent to sample the previous failure to the mini-batch (similar to that deployed in \cite{synergy}, however, \cite{synergy} includes failed grasp during testing phase but not training phase). After performing a SGD step on a minibatch including the most-recent failed grasp, its corresponding $Q$ value decreases, thus the agent will avoid executing the same action again.

% \subsection{Adapting the Grasping Problem}

% To adapting the Contextual Bandit grasping problem (AGP), we adapt ASR in the following ways. First, to overcome the agent execute repetitive failure grasps, we sample the transition of the previous failure grasp in the mini-batch, similar to that deployed in \cite{synergy} (however, \cite{synergy} include failure grasp during testing phase, not training). After performing a SGD step on this failure grasp, its corresponding $Q$ value decreases thus avoiding execute this action again. Second, we use softmax layer as the last layer of the network. The layer produces two classes probability, correspond to $Q$ value for success and failure. We choose the class success of the output to model the binary reward.

% \vspace{0.15cm}
% \noindent
\subsubsection{Boltzmann exploration}

We compared Boltzmann exploration with $\epsilon$-greedy exploration and found Boltzmann to be better in our grasp setting. We use a temperature of $\tau_\text{training}$ during training and a lower temperature of $\tau_\text{test}$ during testing. Using a non-zero temperature at test time helped reduce the chances of repeatedly sampling a bad action.

% \begin{wrapfigure}{r}{0.2\textwidth}
%   \begin{center}
%     \includegraphics[width=0.2\textwidth]{figure/z_heuristic.png}
%   \end{center}
%     \caption{The red box is $rec$, the two blue boxes are $rec_\text{edge}$.}
%     \label{fig:z_heuristic}
% \end{wrapfigure}

% \vspace{0.15cm}
% \noindent
\subsubsection{Data augmentation}

Even though we are using equivariant neural networks to encode the $Q$ function, it can still be helpful to perform data augmentation as well. This is because the granularity of the rotation group encoded in $q_1$ ($D_4$) is coarser than that of the action space ($C_n/C_2$). We address this problem by augmenting the data with translations and rotations sampled from $\hat{\SE}(2)$. For each experienced transition, we add eight additional $\hat{\SE}(2)$-transformed images to the replay buffer.
% \textbf{Data augmentation:} Even though we are using equivariant neural networks to encode the $Q$ function, it can still be helpful to do data augmentation as well. This is because the equivariant network is expressed over a finite rotation group, e.g. ${C_{16}}/{C_2}$ or ${C_{32}}/{C_2}$, and there therefore exist ``gaps'' between discrete orientations in the representation. We address this problem by augmenting the data with translated and rotations sampled from a continuous region within $\SE(2)$. For each experienced transition, we add eight additional images to the replay buffer that have been transformed in this way.

% \vspace{0.15cm}
% \noindent
\subsubsection{Softmax at the output of $q_1$ and $q_2$} 

Since we are using a contextual bandit with binary rewards and the reward function $r(s,a)$ denotes the parameter of a Bernoulli distribution at $s,a$, we know that $Q_1$ and $Q_2$ must each take values between zero and one. We encode this prior using an entry-wise softmax layer at the output of each of the $q_1$ and $q_2$ networks.

% Since the reward function of the contextual bandit is either zero or one, the optimal $Q$ function at a given state can be viewed as the probability of a success at that state. We encoded this prior knowledge using an \hl{entry-wised} softmax layer at the output of the $q_1$ and $q_2$ networks. \hl{Thus each entry of $q_1$ and $q_2$ is in $(0, 1)$.}

% \vspace{0.15cm}
% \noindent
\subsubsection{Selection of the $z$ coordinate} 

In order to execute a grasp, we must calculate a full $x,y,z$ goal position for the gripper. Since our model only infers a planar grasp pose, we must calculate a depth along the axis orthogonal to this plane (the $z$ axis) using other means. In this paper, we calculate $z$ by taking the average depth over a $5 \times 5$ pixel region centered on the grasp point in the input depth image. The commanded gripper height is set to an offset value from this calculated height. While executing the motion to this height, we monitor force feedback from the arm and halt the motion prematurely if a threshold is exceeded. (In our physical experiments on the UR5, this force is measured using torque feedback from the joints.)

\section{Experiments in Simulation}
\label{sect:experiments}

\subsection{Setup}
\label{sec:simulation_experiments}

\begin{figure*}
    \centering
    \subfigure[86 GraspNet-1B objects used]{\includegraphics[height=0.25\textwidth]{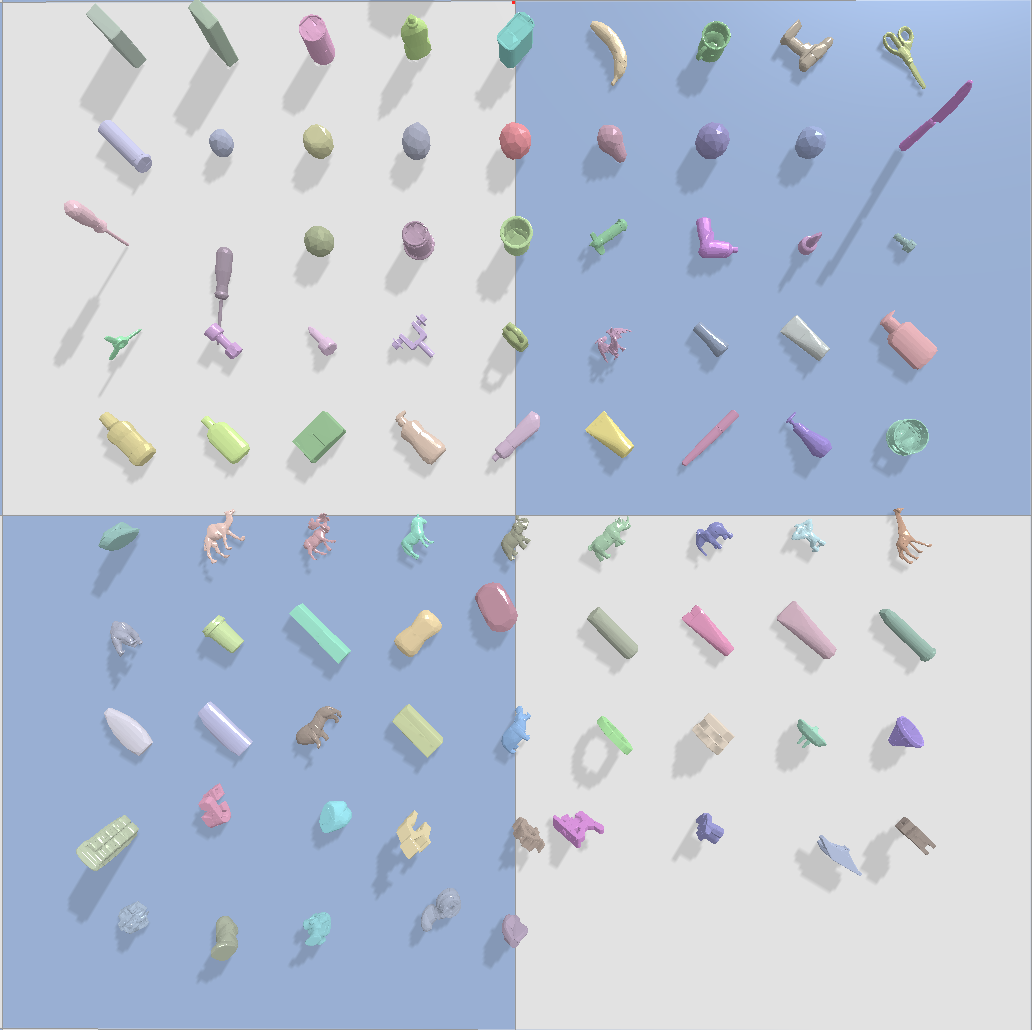}}
    \subfigure[Pybullet simulation]{\includegraphics[height=0.25\textwidth]{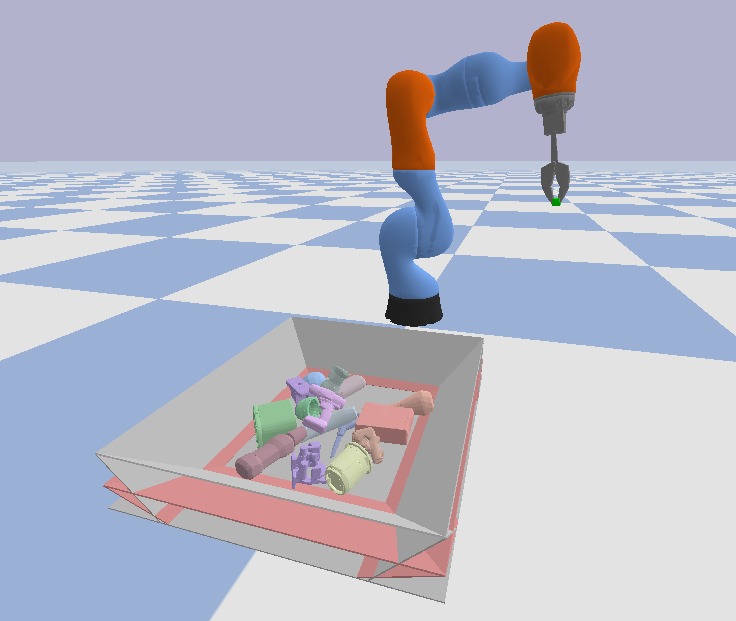}}
    \subfigure[Depth image]{\includegraphics[height=0.25\textwidth]{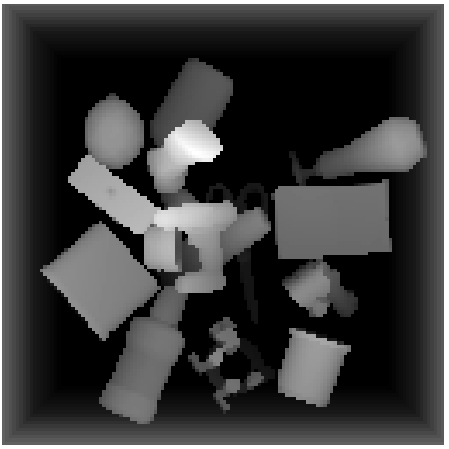}}
    \caption{(a) The 86 objects used in our simulation experiments drawn from the GraspNet-1Billion dataset~\cite{graspnet-1b}. (b) Phybullet simulation. (c) State is a top-down image of the grasp scene.}
    \label{fig:simulation_exps}
\end{figure*}

% \begin{wrapfigure}{r}{0.25\textwidth}
% \vspace{-0.4cm}
%   \begin{center}
%     \includegraphics[width=0.25\textwidth]{figure/experiment/GraspNet_obj.png}
%   \end{center}
%     \caption{The 86 objects used in our simulation experiments drawn from the GraspNet-1Billion dataset~\cite{graspnet-1b}.}
%     \label{fig:z_heuristic}
% \end{wrapfigure}

% \vspace{0.15cm}
% \noindent
% \textbf{Object Set:} 
\subsubsection{Object Set}

All simulation experiments are performed using objects drawn from the GraspNet-1Billion dataset~\cite{graspnet-1b}. This includes 32 objects from the YCB dataset~\cite{ycb}, 13 adversarial objects used in DexNet 2.0~\cite{mahler2017dex}, and 43 additional objects unique to GraspNet-1Billion~\cite{graspnet-1b} (a total of 88 objects). Out of these 88 objects, we exclude two bowels because they can be stably placed in non-graspable orientations, i.e. they can be placed upside down and cannot be grasped in that orientation using standard grippers. Also, we scale these objects so that they are graspable from any stable object configuration. we refer to these 86 mesh models as our simulation ``object set'', shown in Figure~\ref{fig:simulation_exps}a.

% \begin{wrapfigure}{r}{0.25\textwidth}
% % \vspace{-0.4cm}
%   \begin{center}
%     \includegraphics[width=0.25\textwidth]{figure/experiment/simulation_environment.png}
%   \end{center}
%     \caption{Pybullet simulation environment.}
%     \label{fig:pybullet_simulaiton}
% \end{wrapfigure}

% \vspace{0.15cm}
% \noindent
% \textbf{Simulation Details:} 

\subsubsection{Simulation Details}

Our experiments are performed in Pybullet~\citep{pybullet}. The environment includes a Kuka robot arm and a $0.3\text{m}\times0.3\text{m}$ tray with inclined walls (Figure~\ref{fig:simulation_exps}b). At the beginning of each episode, the environment is initialized with $15$ objects drawn uniformly at random from our object set and dropped into the tray from a height of $40$ cm so that they fall into a random configuration. State is a depth image captured from a top-down camera (Figure~\ref{fig:simulation_exps}c). On each time step, the agent perceives state and selects an action to execute which specifies the planar pose to which to move the gripper. A grasp is considered to have been successful if the robot is able to lift the object more than $0.1$m above the table. The episode continues until all objects have been removed from the tray or until $30$ grasp attempts have been made at which point the episode terminates and the environment is reinitialized.

\subsection{Comparison Against Baselines}
\label{sect:baselines}

\subsubsection{Baseline Model Architectures}

% Figure~\ref{fig:simulation_baseline} compares 

We compare our method against two different model architectures from the literature: VPG~\cite{synergy} and FC-GQ-CNN~\citep{4Dof}. Each model is evaluated alone and then with two different data augmentation strategies (soft equ and RAD). In all cases, we use the contextual bandit formulation described in Section~\ref{sect:bandit}. The baseline model architectures are: \underline{VPG:} Architecture used for grasping in~\citep{synergy}. This model is a fully convolutional network (FCN) with a single-channel output. The $Q$ value of different gripper orientations is evaluated by rotating the input image. We ignore the pushing functionality of VPG. \underline{FC-GQ-CNN:} Model architecture used in~\citep{4Dof}. This is an FCN with $8$-channel output that associates each grasp rotation to a channel of the output. During training, our model uses Boltzmann exploration with a temperature of $\tau = 0.01$ while the baselines use $\epsilon$-greedy exploration starting with $\epsilon = 50\%$ and ending with $\epsilon = 10\%$ over 500 grasps (this follows the original implementation in~\cite{synergy}).

\subsubsection{Data Augmentation Strategies}

The data augmentation strategies are: 
\underline{$n\times$ RAD:} The method from~\cite{rad} where we perform $n$ SGD steps after each grasp sample, where each SGD step is taken over a mini-batch of samples for which the observation and action have been randomly translated and rotated. \underline{$n\times$ soft equ:} same as $n \times$ RAD except that we produce a mini-batch by drawing $bs/n$ (where $bs$ is the batch size) samples and then randomly augmenting those samples $n$ times. Details can be found in Appendix~\ref{sec:aug_baseline}.

% Appendix~\ref{sec:aug_baseline}

% a data augmentation method that performs $n$ soft equivariant SGD steps per grasp, where each SGD step is taken over $n$ times randomly \SE(2) augmented mini-batch \hl{Specifically, we sample $1/n$ batch size of data, augment it $n\times$ and train on this mini-batch. This augmentation aims at achieve equivariance on the mini-batch.} (THIS SOUNDS EXACTLY THE SAME AS RAD). \hl{We choose the best augmentation results for these strategies, see appendix}\ref{sec:aug_baseline}.

% \cite{wang2021equivariant}

% \underline{$n\times$ Soft Equivariant} \cite{wang2021equivariant}: a data augmentation method that perform $n$ soft equivariant SGD steps per grasp, where each SGD step is taken over $n$ times randomly \SE(2) augmented mini-batch; 

% 4) \textit{$n\times$ RAD} \cite{rad}: a data augmentation method that perform $n$ times of RAD SGD steps per grasp, where each SGD step is taken over randomly \SE(2) perturbed mini-batch.

\subsubsection{Results}

The learning curves of Figure~\ref{fig:simulation_baseline} show grasp success rate versus number of grasp attempts. Figure~\ref{fig:simulation_baseline}a shows on-line learning performance. Our method uses Boltzmann exploration while the baselines use $\epsilon$-greedy as described above. Each curve connects data points evaluated every 150 grasp attempts. Each data point is the average success rate over the last 150 grasps (therefore, the first data point occurs at 150). Figure~\ref{fig:simulation_baseline}b shows near-greedy performance by stopping training every 150 grasp attempts and performing 1000 test grasps and reporting average performance over these 1000 test grasps. Our method tests at a lower test temperature of $\tau=0.002$ while the baselines test pure greedy behavior. 

% \textbf{Baseline methods:} 1) \textit{VPG} \citep{synergy}: an FCN with single-channel output that estimates the $Q$-map for each rotation in the action space by rotating the input image. We only use the grasping network and use DQN to solve the bandit problem; 

\subsubsection{Discussion of Results}

\begin{figure}
    \centering
    \subfigure[Training]{\includegraphics[width=0.24\textwidth]{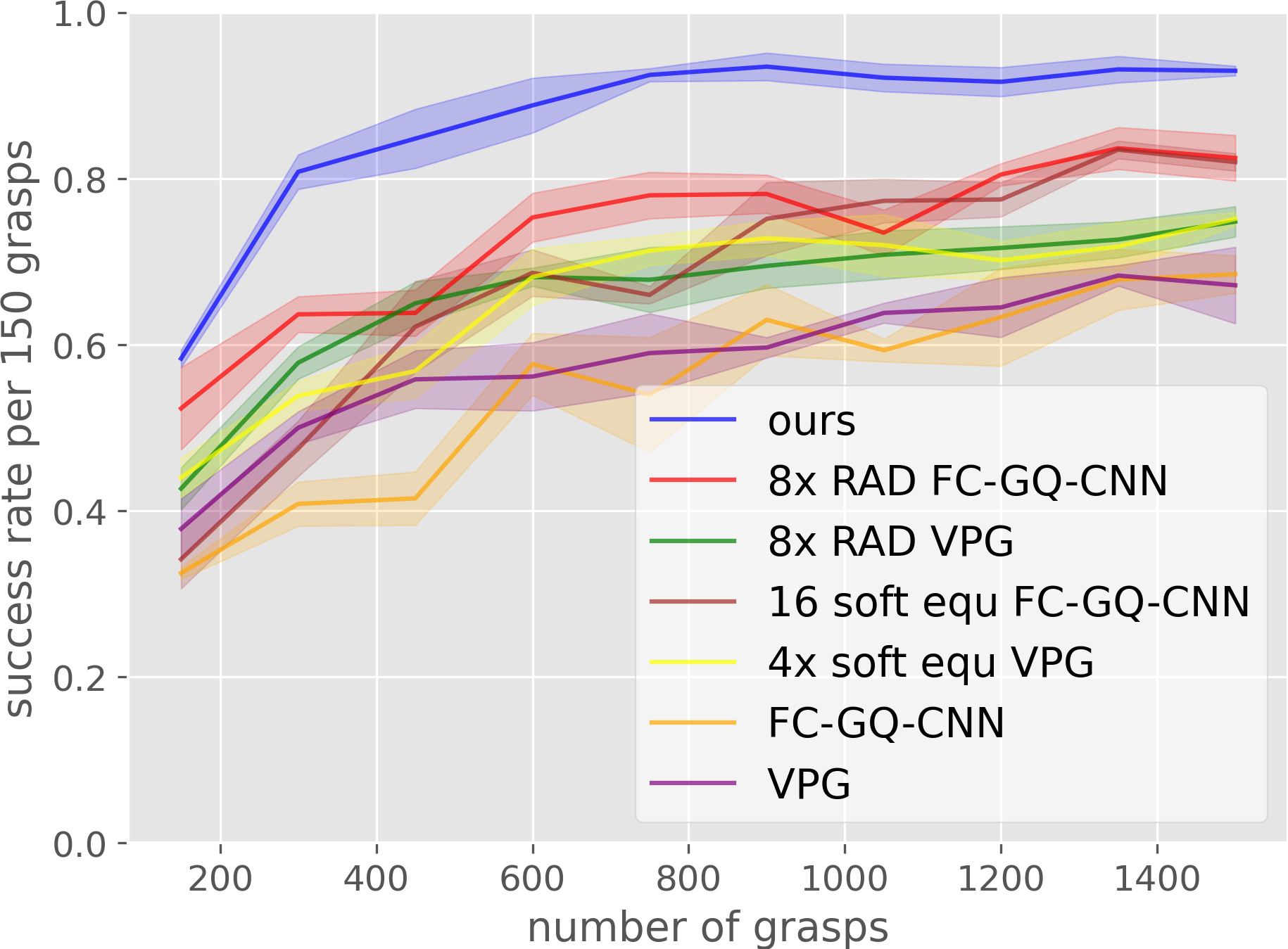}}
    \subfigure[Testing]{\includegraphics[width=0.24\textwidth]{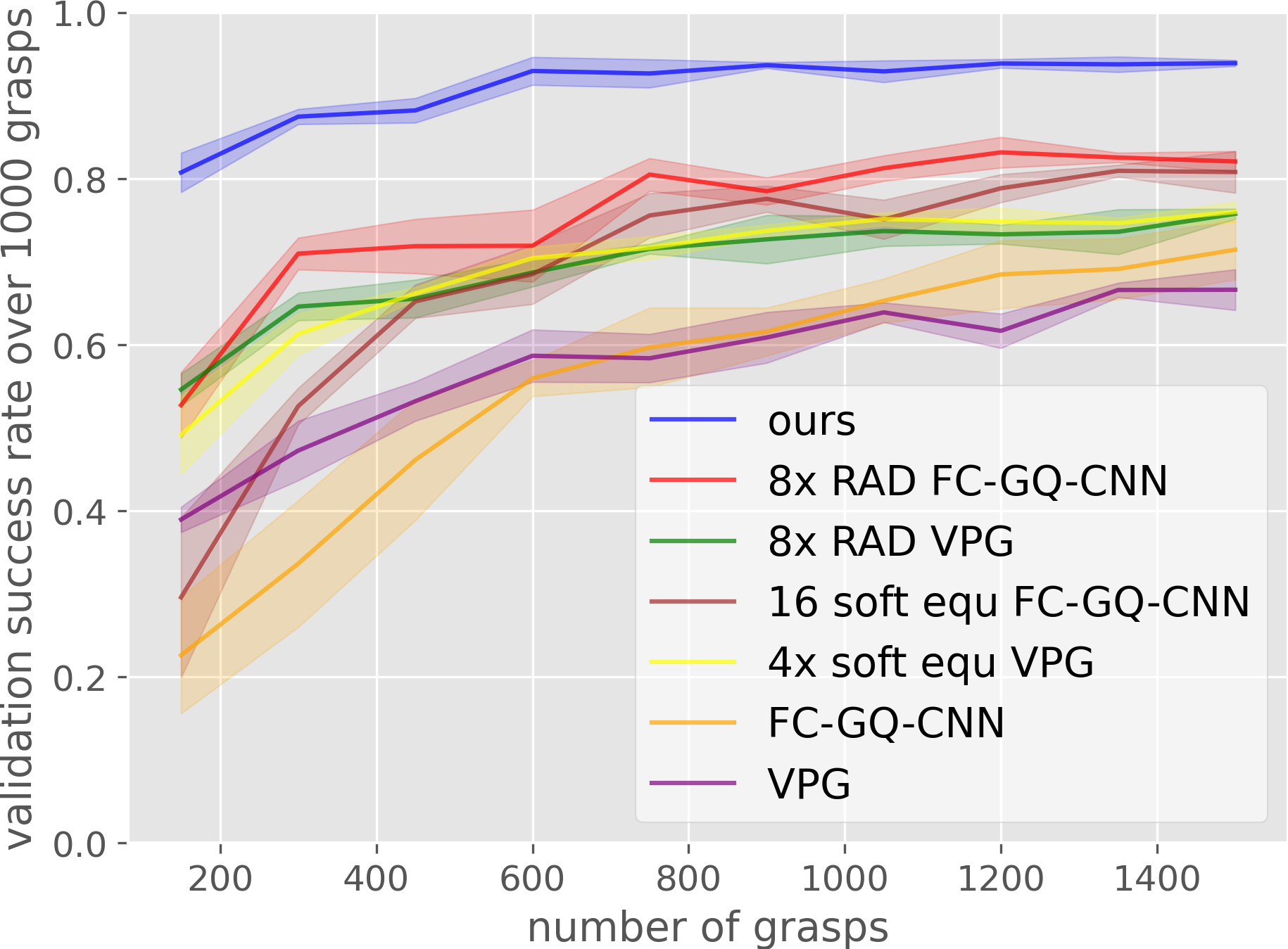}}
    \caption{Comparison with baselines. All lines are an average of four runs. Shading denotes standard error. (a) shows learning curves as a running average over the last 150 training grasps. (b) shows average near-greedy performance of 1000 validation grasps performed every 150 training steps.}
    \label{fig:simulation_baseline}
\end{figure}

Generally, our proposed equivariant model convincingly outperforms the baseline methods and data augmentation strategies. In particular, Figure~\ref{fig:simulation_baseline}b shows that the grasp success rate of the near-greedy policy learned by the equivariant model after 150 grasp attempts is at least as good as that learned by any of the other baselines methods after 1500 grasp attempts. Notice that each of the two data augmentation methods we consider (RAD and soft equ) have a positive effect on the baseline methods. However, after training for the full 1500 grasp attempts, our equivariant model converges to the highest grasp success rate ($93.9\pm0.4\%$).

% There are two findings in the baseline comparisons in simulation experiment. First, ours outperforms all baselines. Specifically, ours achieved as good test success rate at the $150$th grasp as the best baseline at the $1500$th grasp, lead to $\times10$ higher sample efficiency. Besides, ours converges to a better success rate - achieved $93.9\pm0.4\%$. Second, for each non-augmented baselines, its augmented version outperforms, for both $n\times$ soft equ or $n\times$ RAD augmentation methods.

\subsection{Ablation Study}
\label{sect:ablations}

There are three main parts to the approach described in this paper: 1) use of equivarant convolutional layers instead of standard convolution layers; 2) use of the augmentated state representation (ASR) instead of a single network; 3) the various optimizations described in Section~\ref{sec:other_optimizations}. Here, we evaluate performance of the method when ablating each of these three parts. 

\subsubsection{Ablations}

In \underline{no equ}, we replace all equivariant layers with standard convolutional layers. In \underline{no ASR}, we replace the equivariant $q_1$ and $q_2$ models described in Section~\ref{sect:asr} by a single equivariant network. In \underline{no opt}, we remove the optimizations described in Section~\ref{sec:other_optimizations}. In addition to the above, we also evaluated \underline{rot equ} which is the same as \underline{no ASR} except that we replace ASR with a U-net\cite{u_net} and apply $4\times$ RAD\cite{rad} augmentation. Detailed network architectures can be found in Appendix~\ref{sec:network_architecture}. 

% \hl{ as $4\times$ is the best parameter in appendix}\ref{sec:aug_baseline}

% \hl{The network architectures for these ablations are shown in figure}\ref{fig:q1q2_architecture} in appendix.

% We ablate ours three major contributions to validate their efficacy: 1) \textit{no equ}: replace equivariant networks by FCN; 2) \textit{no ASR}: replace $q_1$ and $q_2$ by one equivariant network; 3) \textit{no opt}: remove the optimizations for the bandit problem in Section~\ref{sec:other_optimizations}. Besides, we implemented a baseline to compare equivariant networks and a FCN that designed to achieve equivariance \textit{rot equ}: replace VPG's\cite{synergy} Densenet\cite{densenet} backbone with U-net\cite{u_net} and apply $4\times$ RAD\cite{rad} augmentation.

\subsubsection{Results and Discussion}

\begin{figure}
    \centering
    \subfigure[Training]{\includegraphics[width=0.24\textwidth]{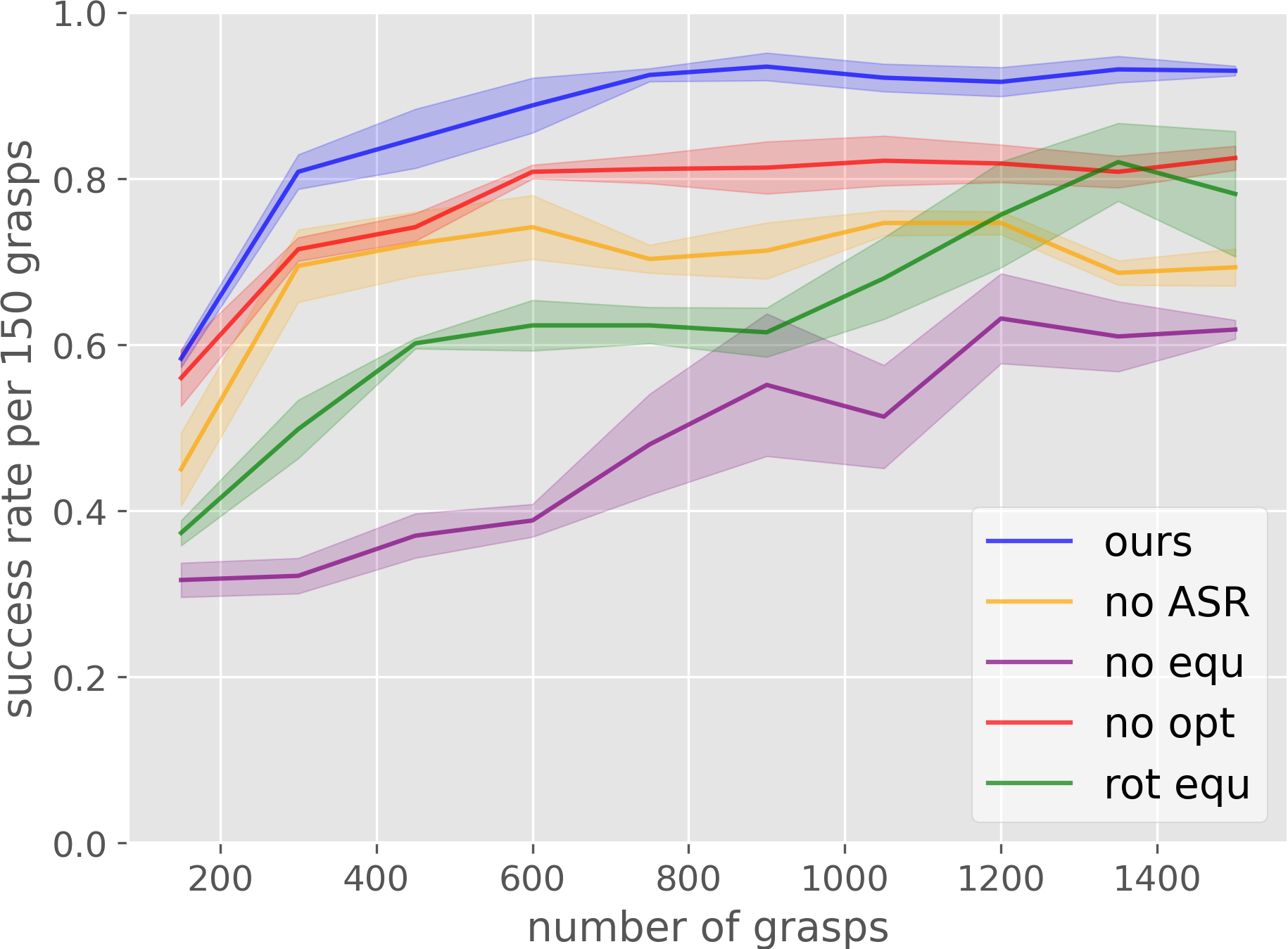}}
    \subfigure[Testing]{\includegraphics[width=0.24\textwidth]{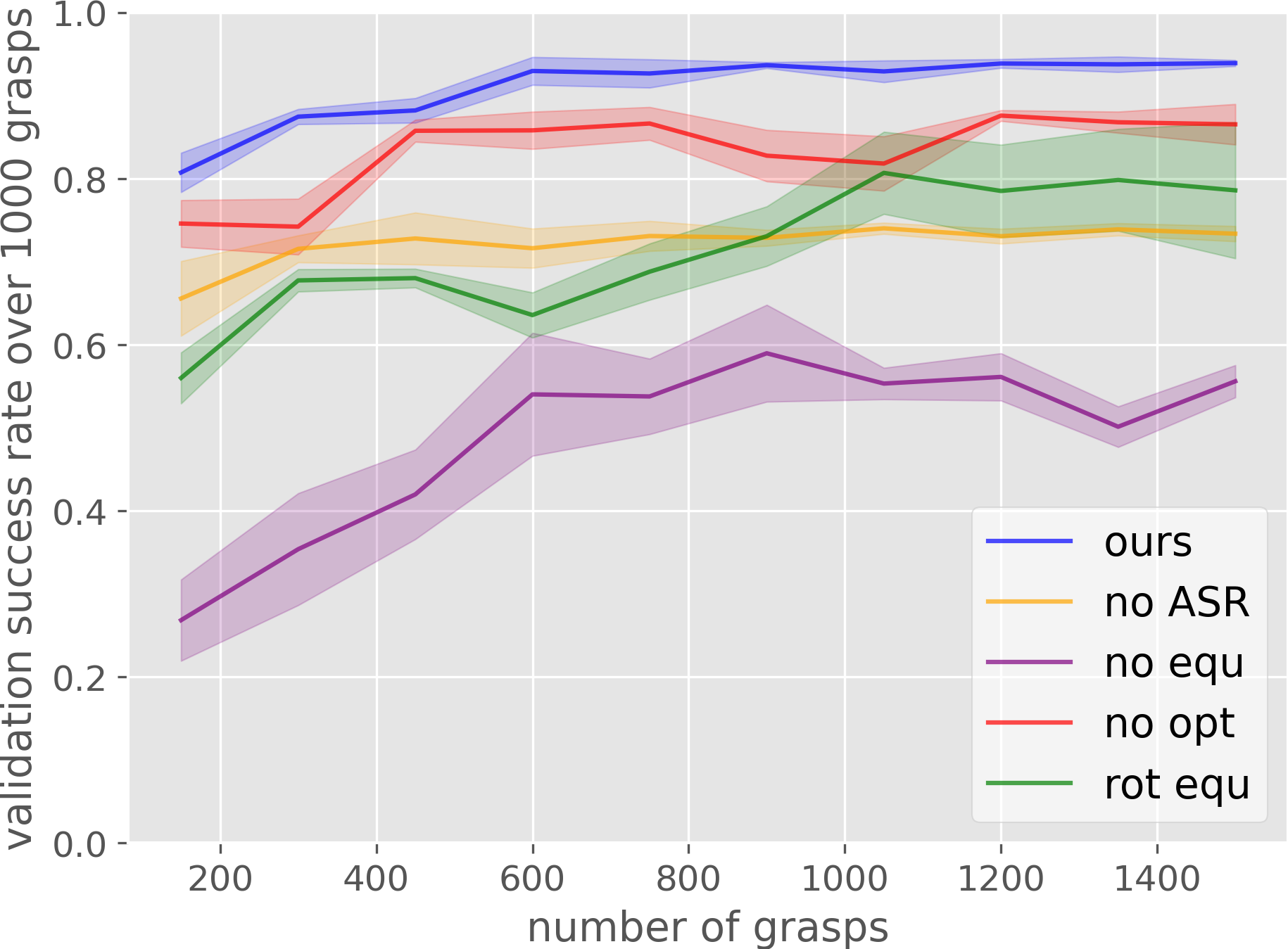}}
    \caption{Ablation study. Lines are an average over $4$ runs. Shading denotes standard error. (a) learning curves as a running average over the last 150 training grasps. (b) average near-greedy performance of 1000 validation grasps performed every 150 training steps.}
    \label{fig:simulaiton_ablation}
\end{figure}

Figure~\ref{fig:simulaiton_ablation} shows the results where they are reported exactly in the same manner as in Section~\ref{sect:baselines}. \underline{no equ} does worst, suggesting that our equivariant model is critical. We can improve on this somewhat by adding data augmentation (\underline{rot equ}), but this sill underperforms significantly. The other ablations, \underline{no ASR} and \underline{no opt} demonstrate that those parts to the method are also important.

% Interestingly, rot equ achieves compatible performance than its equivariant counterpart --no ASR. Rot equ's resembles no ASR, i.e., rot equ has $|a_\theta|$ of networks where the equivariant networks in no ASR has $|a_\theta|$ of trivial representation channels. Though rot equ requires more data augmentation, GPU buffer, and SGD steps.

% \begin{figure}
%     \centering
%     \subfigure[Robot setup]{\includegraphics[width=0.28\textwidth]{figure/experiment/UR5_setup.png}}
%     \subfigure[Observation]{\includegraphics[width=0.26\textwidth]{figure/RL experiment/ur5_obs.png}}
%     \caption{}
%     \label{tab:robot_setup}
% \end{figure}

\begin{figure}
    \centering
    \includegraphics[width=0.28\textwidth]{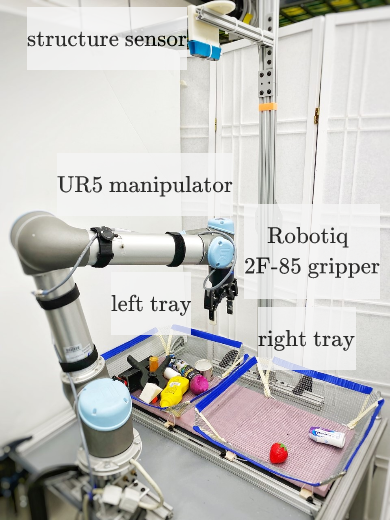}
    \caption{Setup for self-supervised training on the robot.}
    \label{fig:robot_setup}
\end{figure}

\section{Experiments in Hardware}
\label{sec:physical_robot_experiments}

\subsection{Setup}

\subsubsection{Robot Environment}

Our experimental platform is comprised of a Universal Robots UR5 manipulator equipped with a Robotiq 2F-85 parallel-jaw gripper, an Occipital Structure Sensor, and the dual-tray grasping environment shown in Figure~\ref{fig:robot_setup}. The work station is equipped with Intel Core i7-7800X CPU and NVIDIA GeForce GTX 1080 GPU.

\subsubsection{Objects}

\begin{figure}
    \centering
    \subfigure[Training set]{\includegraphics[width=0.145\textwidth]{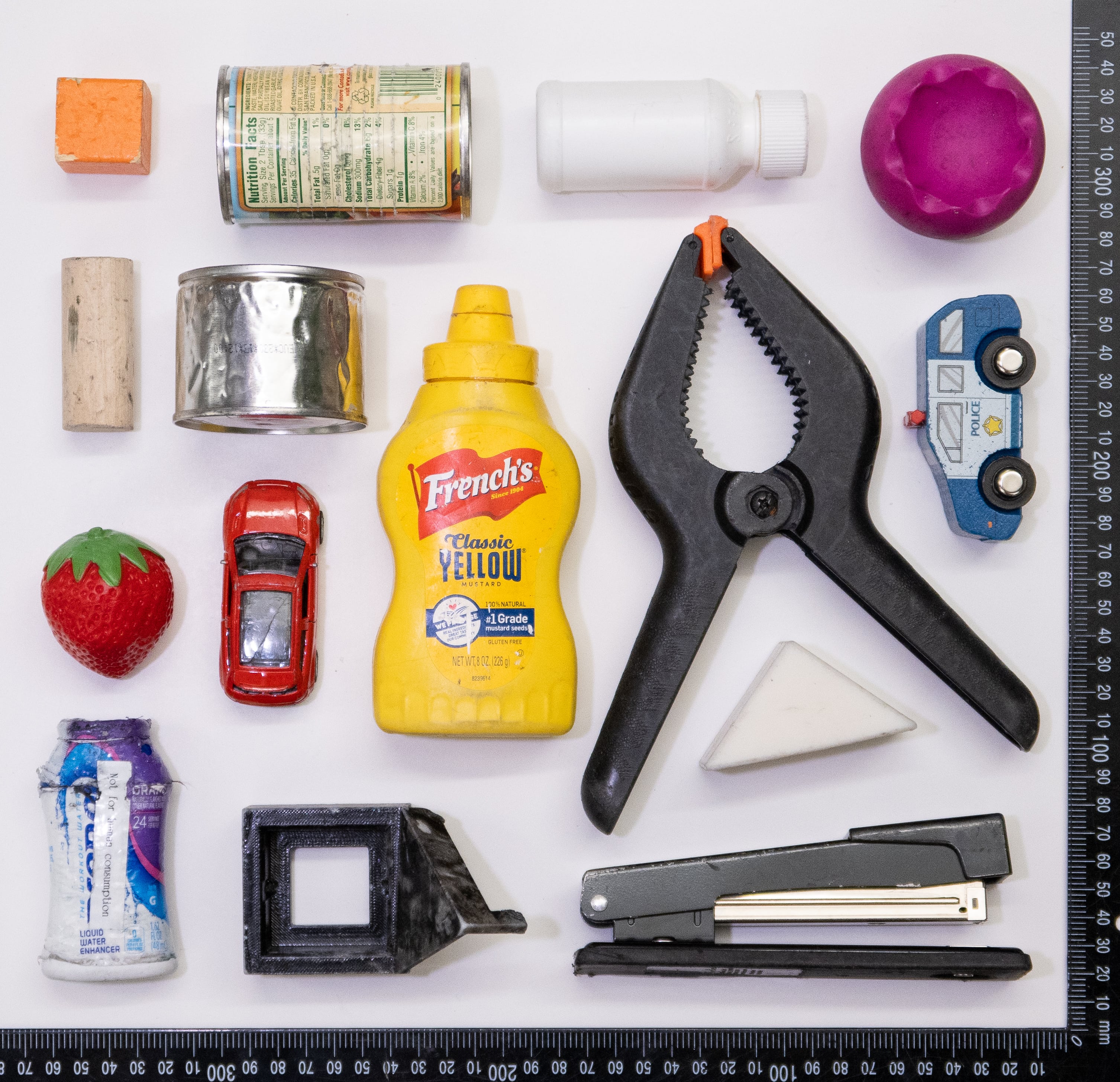}}
    \subfigure[Testing set, easy]{\includegraphics[width=0.142\textwidth]{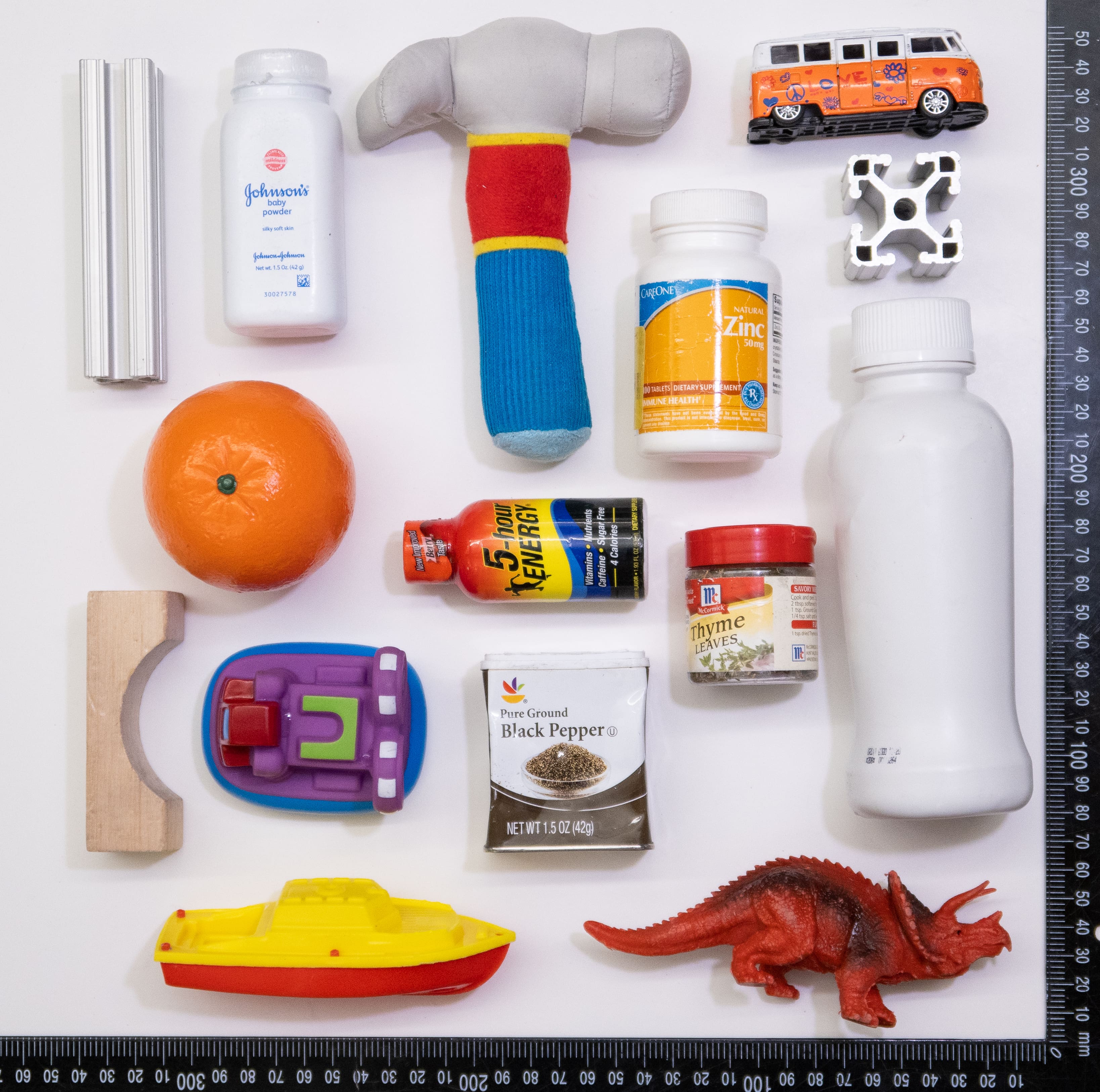}}
    \subfigure[Testing set, hard]{\includegraphics[width=0.177\textwidth]{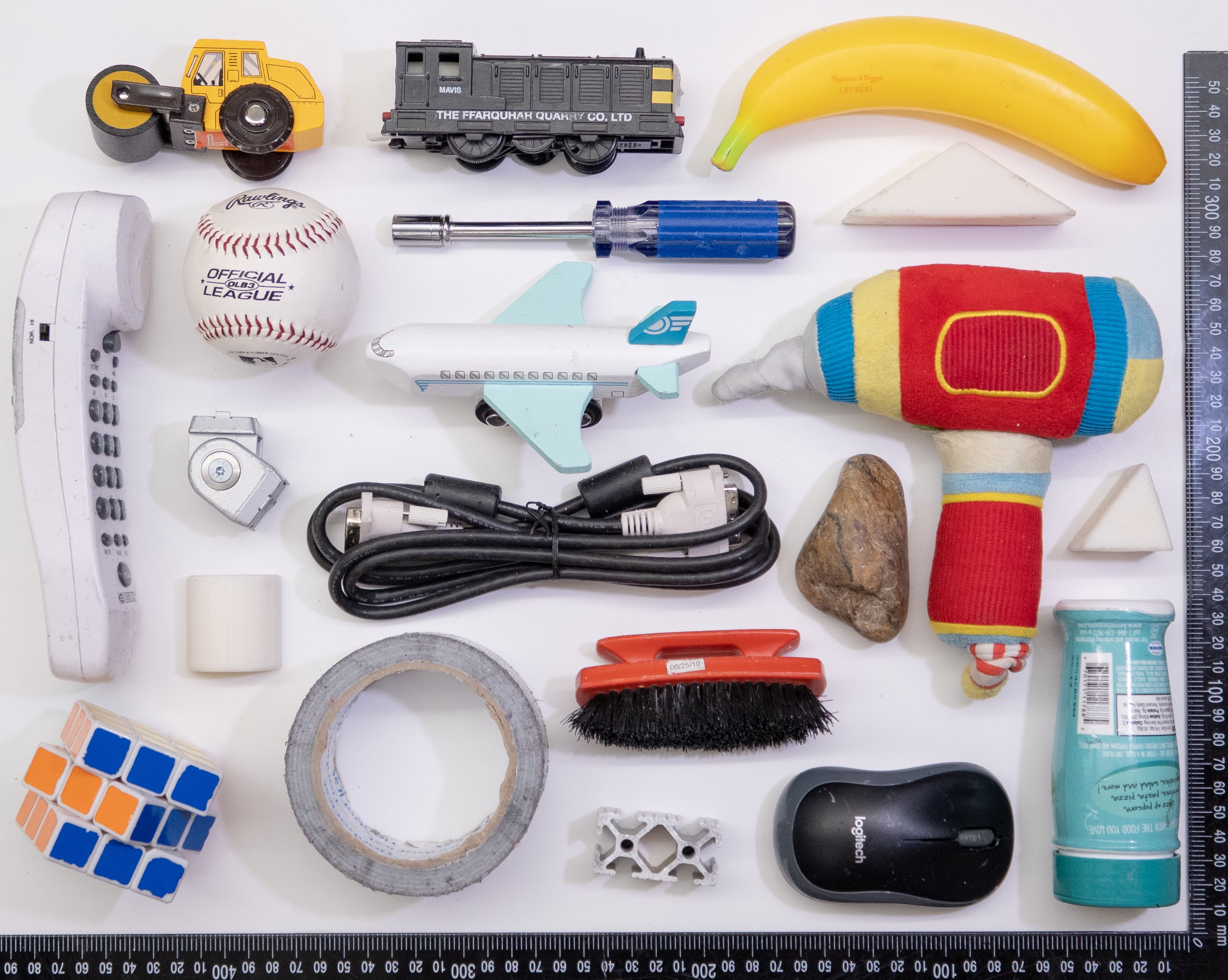}}
    \caption{Object sets used for training and testing. Both training and test set easy include 15 objects while test set hard has 20 objects. Objects were curated so that they were graspable by the Robotiq 2F-85 parallel jaw gripper from any configuration and visible to the Occipital Structure Sensor.}
    % The train/test object sets. Objects need to be graspable of the gripper at any configuration (smaller than the gripper open width and has enough height), and need be able been captured by the depth camera (no transparent nor has thing wall).}
    \label{fig:obj_sets}
\end{figure}

All training happens using the 15 objects shown in Figure~\ref{fig:obj_sets}a. After training, we evaluate grasp performance on both the ``easy'' test objects (Figure~\ref{fig:obj_sets}b) and the ``hard'' test objects (Figure~\ref{fig:obj_sets}c). Note that both test sets are novel with respect to the training set.

\subsubsection{Self-Supervised Training}

At the beginning of training, the 15 training objects (Figure~\ref{fig:obj_sets}a) are dropped into one of the two trays by the human operator. Then, we train by attempting to grasp these objects and place them in the other bin. All grasp attempts are generated by the contextual bandit. When all 15 objects have been transported in this way, training switches to attempting to grasp from the other bin and transport them into the first. Training continues in this way until 600 grasp attempts have been performed (that is 600 grasp \emph{attempts}, not 600 successful grasps). A grasp is considered to be successful if the gripper remains open after closing stops due to a squeezing force. To avoid systematic bias in the way the robot drops objects into a tray, we sample the drop position randomly from a Gaussian distribution centered in the middle of the receiving tray.

% Grasp execution during training is automated. Initially, 15 training objects are randomly placed into one of the two trays (the object sets used during training are shown in Figure~\ref{fig:obj_sets}a and b.) Then, the robot grasps each of the objects and transports it to the other tray. \hl{During transport, a successful grasp is recognized by thresholding gripper aperture.} This process repeats with all 15 objects being transported between one tray and the other until sufficient training has occurred. To avoid systematic bias in the way objects fall into a tray, we sample the drop position randomly from a Gaussian distribution centered in the middle of the receiving tray.

% \textbf{Optimizations for training on a physical robot:} 

\subsubsection{In-Motion Computation}

We were able to nearly double the speed of robot training by doing all image processing and model learning while the robotic arm was in motion. This was implemented in Python as a producer-consumer process using mutexs. As a result, our robot is constantly in motion during training and the training speed for our equivariant algorithm is completely determined by the speed of robot motion. This improvement enabled us to increase robot training speed from approximately 230 grasps per hour to roughly 400 grasps per hour.

\subsubsection{Model Details}

For all methods, prior to training on the robot, model weights are initialized randomly using an independent seed. No experiences from simulation are used, i.e. we train from scratch. The model and training parameters used in the robot experiments are the same as those used in simulation. For our algorithm, the $q_1$ network is defined using $D_4$-equivariant layers and the $q_2$ network is defined using $C_{16}/C_2$-equivariant layers. During training, we use Boltzmann exploration with a temperature of $0.01$. During testing, the temperature is reduced to $0.002$ (near-greedy). For more details, see Appendix~\ref{sec:evaluation_details}.

% We were able to significantly accelerate real-robot training by processing camera data and executing SGD steps while the robotic system was in motion \hl{(in real time)}. This was implemented as a producer-consumer process using mutexs. With this improvement, the robot is constantly in motion during training. As a result, training speed completely determined by the speed of robot motion. In our setup, this approach enabled us to increase robot training speed from approximately 230 grasps per hour to roughly 500 grasps per hour.

% The training process, which is agent-environment interaction, can executed in parallel to increase the mean grasp per hour. The genera idea is paralleling the manipulator executing action, sensor collecting and processing data, neural network forward/ backward path processes. The parallel training is implemented as a producer-consumer process using mutex. After paralleling, the only bottleneck of the training speed is the robot execution speed. The parallel training accelerates the agent-environment interaction from ~$230$ grasps per hour to ~$500$ grasps per hour.

% \begin{wrapfigure}[28]{r}{0.4\textwidth}
% % \vspace{-0.5cm}
%   \begin{center}
%   \includegraphics[width=0.4\textwidth]{figure/experiment/Training_with_500_grasps.png}
%     \caption{Results from 500-grasp experiment. All curves are averaged over $3$ runs.}
%     \end{center}
%     \label{fig:physical_experiments500}
% \end{wrapfigure}

\begin{table}
 \centering
    \begin{tabular}{ m{9.5em} m{5em} m{5em} m{5.5em}}
        \toprule
        Baseline &test set easy & test set hard & training $t_\text{SGD}$\\
        \midrule
        % Random &  39.5 \pm$2.12$ & 36.0 \pm$4.24$ & -\\
        $8\times$ RAD VPG &  61.8 $\pm$3.59 & 52.5 $\pm$5.33 & 12.1\\
        $8\times$ RAD FC-GQ-CNN &  82.8 $\pm$1.65 & 74.3 $\pm$3.04 & 1.55 \\
        Ours & 95.0 $\pm$1.47 & 87.0 $\pm$1.87 & 1.11\\
        \bottomrule
    \end{tabular}
    \caption{Evaluation success rate ($\%$), standard error, and training time per grasp (in seconds) in the hardware experiments. Results are an average of 100 grasps per training run averaged over four runs, performed on the held out test objects shown in Figure~\ref{fig:obj_sets}b and c.}      \label{tab:test_reuslts_600}
\end{table}

% In this experiment, we train each of three different methods over a period of 600 grasp trials for the set of 15 objects shown in Figure~\ref{fig:obj_sets}a on the UR5 system. At the beginning of each grasp run, we deposit these same 15 objects into one of the two trays and run 600 grasp trials during which the learning algorithm trains, actively selecting each successive grasp according to its model and its action selection mechanism. 

% As in the simulation experiments, our $q_1$ network is defined for the group $D_4$ and our $q_2$ network is defined for the group $C_{16}/C_2$. During training, training the Boltzmann temperature is $0.01$. After training, during evaluation, it is $0.002$. 

\subsubsection{Baselines}

In our robot experiments, we compare our method against \underline{$8\times$ RAD VPG}~\citep{synergy}~\cite{rad} and \underline{$8\times$ RAD FC-GQ-CNN}~\citep{4Dof}~\cite{rad}, the two baselines we found to perform best in simulation. As before, \underline{$8\times$ RAD VPG}, uses a fully convolutional network (FCN) with a single output channel. The $Q$ map for each gripper orientation is calculated by rotating the input image. After each grasp, we perform $8\times$ RAD data augmentation (8 optimization steps with a mini-batch containing randomly translated and rotated image data). \underline{$8\times$ RAD FC-GQ-CNN} also has an FCN backbone, but with eight output channels corresponding to each gripper orientation. It uses $8\times$ RAD data augmentation as well. All exploration is the same as it was in simulation except that the $\epsilon$-greedy schedule goes from $50\%$ to $10\%$ over 200 steps rather than over 500 steps.

% Our method uses Boltzmann exploration  with a temperature of $\tau=0.01$. The baselines~\cite{4dof,synergy} use $\epsilon$-greedy exploration starting with $\epsilon = 50\%$ and ending with $\epsilon = 10\%$ over 200 grasps (this follows the original implementation in~\cite{synergy}). 

\subsubsection{Evaluation procedure}

A key failure mode during testing is repeated grasp failures due to an inability of the model to learn quickly enough. To combat this, we use the procedure of~\cite{synergy} to reduce the chances of repeated grasp failures (we use this procedure only during testing, not training). After a grasp failure, we perform multiple SGD steps using that experience to ``discourage'' the model from selecting the same action a second time and then use that updated model for the subsequent grasp. Then, after a successful grasp has occurred, we discard these updates and return to the original network.

% The evaluation of baselines are greedy policy on the $q$ outputted by the baseline. Moreover, after each failure grasp, the baseline will take \hl{SGD steps} on this failure grasp and reload the network after the next success grasp, as VPG adopted in\cite{synergy}. \hl{See the detals in appendix}\ref{sec:evaluation_details}.

% We compare against two baselines: 1) \textit{$8\times$ RAD VPG}~\citep{synergy}~\cite{rad} where we have an FCN with a single channel output that estimates the $Q$ map for each rotation in the action space by rotating the input image. Moreover, after each grasp $8$ SGD steps with random \SE(2) perturbed mini-batch; 2) \textit{$8\times$ RAD FC-GQ-CNN}~\citep{4Dof}~\cite{rad} use FCN as backbone with $8$ SGD steps with random \SE(2) perturbed mini-batch.

\subsection{Results} 

% egreedy 50% - 10% over 500 grasps (200 for hardware)

\subsubsection{Results}

\begin{figure}
    \centering
    \includegraphics[height=0.25\textwidth]{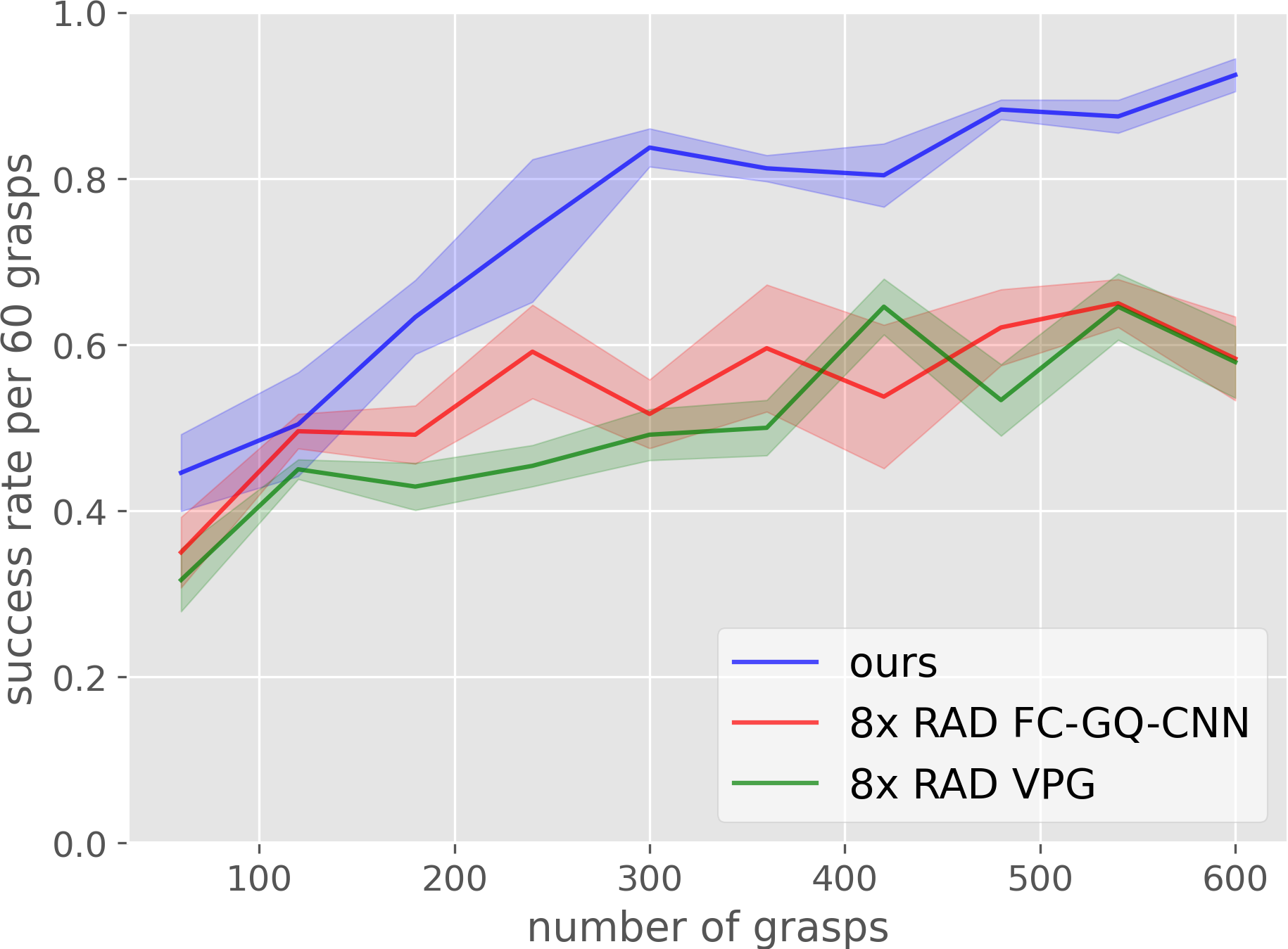}
    \caption{Learning curves for the hardware experiment. All curves are averaged over $4$ runs. Shading denotes standard error.}
    \label{fig:physical_experiments}
\end{figure}

Figure~\ref{fig:physical_experiments} shows the learning curves for the three methods during learning. Each curve is an average of four runs starting with different random seeds and random object placement. Each data point is the average grasp success over the last 60 grasp attempts during training. Table~\ref{tab:test_reuslts_600} shows the performance of all methods after the 600-grasp training is complete. All methods are evaluated by freezing the corresponding model and executing 100 greedy (or near greedy) test grasps for the easy-object test set (Figure~\ref{fig:obj_sets}b) and 100 additional test grasps for the hard-object test set (Figure~\ref{fig:obj_sets}c).

\subsubsection{Discussion}

Probably the most important observation to make is that the results from training on the physical robot (Figure~\ref{fig:physical_experiments}) matches the simulation training results (Figure~\ref{fig:simulation_baseline}a) closely. After 600 grasp attempts, our method achieves a success rate of $>90\%$ while the baselines are near $70\%$. The other observation is that since each of these 600-grasp training runs takes approximately 1.5 hours, it is reasonable to expect that this method could adapt to physical changes in the robot very quickly. Looking at Table~\ref{tab:test_reuslts_600}, our method significantly outperforms the baselines during testing as well, although performance is significantly lower on the ``hard'' test set. We hypothesize that the lower ``hard'' set performance is due to a lack of sufficient diversity in the training set. 

% At the end of training, the model is frozen and evaluated for the 15 test objects shown in Figure~\ref{fig:obj_sets}c. 

% Benefiting from our parallel design, the robot runs continuously without stopping for taking depth images or inferencing/performing SGD step on the network. 

\section{Experiments with the Jacquard Dataset}

% Since many grasp detection models have been evaluated on the Jacquard dataset, 

We also evaluated our model using a large online dataset known as Jacquard~\cite{jacquard} consisting of approximately 54k RGBD images with a total of 1.1M labeled grasps~\cite{jacquard}.

\subsubsection{Setup}

Here, we discard the bandit framework and simply evaluate our equivariant model on Jacquard in the context of supervised learning. Since our focus is on sample efficiency, we train our model on small randomly sampled subsets of 16, 64, 256, and 1024 images drawn randomly from the 54k Jacquard images. In all cases, we evaluated against a single set of 240 Jacquard images held out from all training images. We preprocess the Jacquard images by rectifying the image background in the depth channel so that it is parallel to the image plane. 

% Fist, the images' depth channel is rectified so that the background is parallel to the image plane. Second, the grasp angle is discretized to $0 \sim 15/16\pi$, corresponding to networks' $16$ channels output. Besides, these networks output $16$ channels tensor for gripper open width. 

% All baselines are trained with $50$ epcohs where each epoch is $1$k SGD steps (trained till converge). After each epoch, a validation is performed. The best validation result is choose.

\subsubsection{Baselines}

We evaluate our model against the following baselines: \underline{GG-CNN}~\cite{morrison2018closing}: a generative grasping fully convolutional network (FCN) that predicts grasp quality, angle, and width; \underline{GR-Conv-Net}~\cite{3d_generative_res_conv}: similar to GG-CNN, but it incorporates a generative residual FCN; \underline{VPG}~\cite{synergy}: an FCN with a single-channel output -- the model architecture used in VPG. The $Q$-value of different gripper orientations is evaluated by rotating the input image; \underline{FC-GQ-CNN}~\cite{4Dof}: an FCN with 16 channel output that associates each grasp rotation to a channel of the output. Since both GG-CNN and GR-Conv-Net were originally designed for the supervised learning setting, we use those methods as originally proposed. However, since VPG, FC-GQ-CNN, and our method are on-policy, we discard the on-policy aspects of those methods and just retain the model. For those methods that output discrete grasp angles, we discretize the grasp angle into 16 orientations between 0 and $\pi$ radians and 16 different values for gripper width.

\begin{figure}
    \centering
    \includegraphics[width=0.4\textwidth]{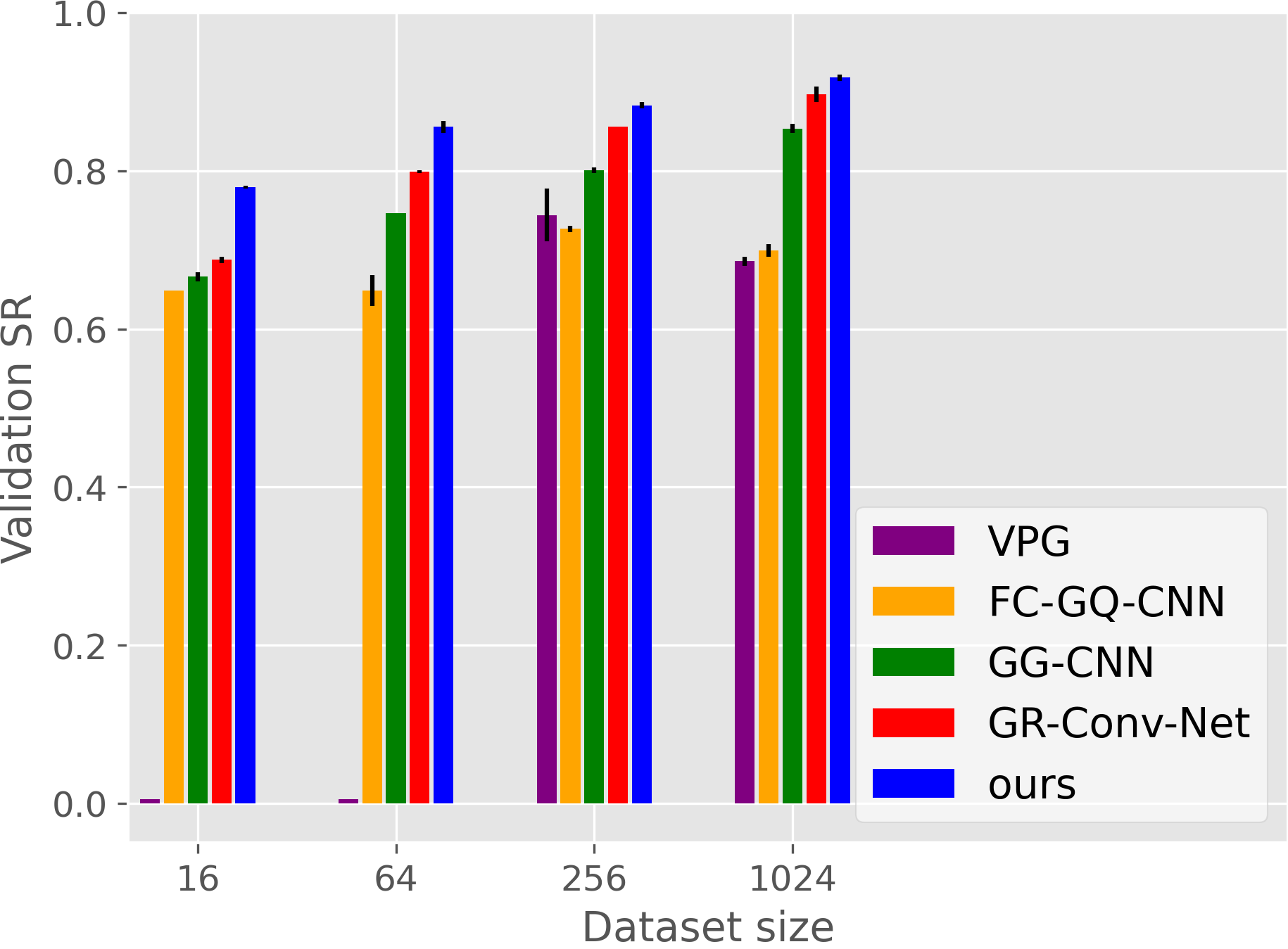}
    \caption{Grasp success rates on Jacquard dataset. All models were trained with four datasets sized 16, 64, 256, and 1024 images. Error bar denotes standard deviation. Results averaged over two runs with independent seeds.}
    \label{fig:jacquard_sl_results}
\end{figure}

\subsubsection{Results}

Figure~\ref{fig:jacquard_sl_results} evaluates the grasp success rates for our method in comparison with the baselines. As is standard in Jaquard, a grasp is considered a ``success'' when the IOU is greater than 25\% and the predicted grasp angle is within 30 degrees of a ground truth grasp. The horizontal axis of Figure~\ref{fig:jacquard_sl_results} shows success rates for the various methods as a function of the four training set sizes. Our proposed equivariant method outperforms in all cases.

% The loss functions are Binary cross entropy on grasp quality and gripper opening; cross entropy on $16$ discritized theta.

\section{Conclusions and Limitations}

This paper recognises that planar grasp detection where the input is an image of the scene and the output is a planar grasp pose is $\SE(2)$-equivariant.
% the grasp function that is learned in the \hl{$\SE(2)$} visual grasp detection problem is $\SE(2)$-equivariant. 
We propose using an $\SO(2)$-equivariant model architecture to encode this structure. The resulting method is significantly more sample efficient than other grasp learning approaches and can learn a good grasp function in less than 600 grasp samples. A key advantage of this increase in sample efficiency is that we are able to learn to grasp completely on the physical robotic system without any pretraining in simulation. This increase in sample efficiency could be important in robotics for a couple of reasons. First, it obviates the need for training in simulation (at least for some problems like grasping), thereby making the sim2real gap less of a concern. Second, it opens up the possibility for our system to adapt to idiosyncrasies of the robot hardware or the physical environment that are hard to simulate. A key limitation of these results, both in simulation and on the physical robot, is that despite the fast learning rate, grasp success rates (after training) still seems to be limited to the low/mid $90\%$ range. This is the same success rate seen in with other grasp detection methods~\cite{mahler2017dex,tenpas_ijrr2017,mousavian20196}, but it is disappointment here because one might expect faster adaptation to lead ultimately to better grasp performance. This could simply be an indication of the complexity of the grasp function to be learned or it could be a result of stochasticity in the simulator and on the real robot. 

% However, further exploration of ways to get closer to a perfect grasp success rate seems to be an important direction for future work.

%% Use plainnat to work nicely with natbib. 

\bibliographystyle{plainnat}
\bibliography{references}

\clearpage
\appendix

\subsection{Neural network architecture}
\label{sec:network_architecture}

\begin{figure*}
    \centering
    \subfigure[ours and no opt]{\includegraphics[width=0.6\textwidth]{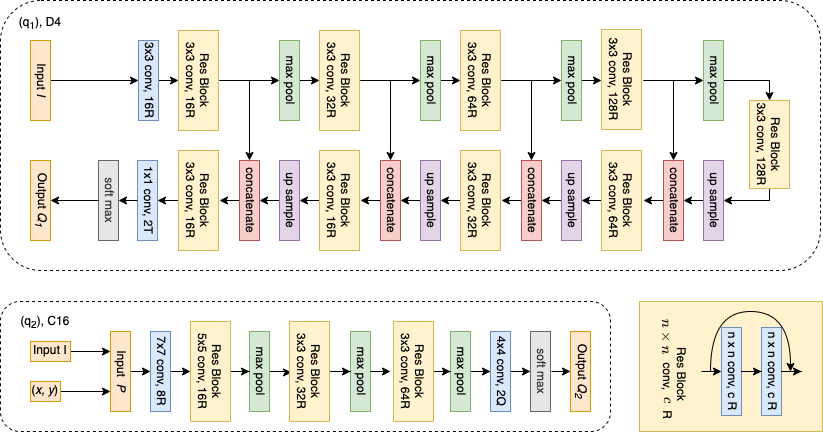}}
    \subfigure[no asr]{\includegraphics[width=0.6\textwidth]{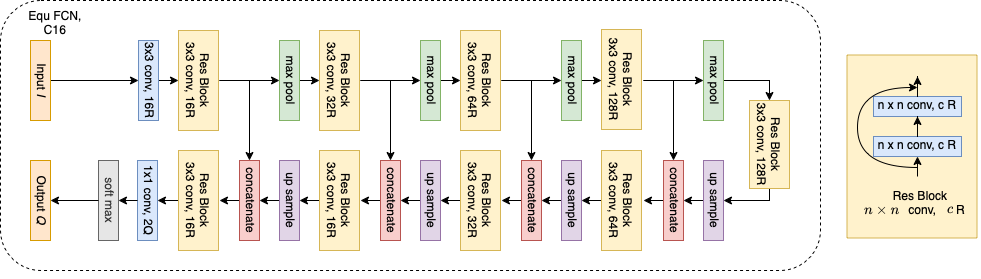}}
    \subfigure[no equ]{\includegraphics[width=0.6\textwidth]{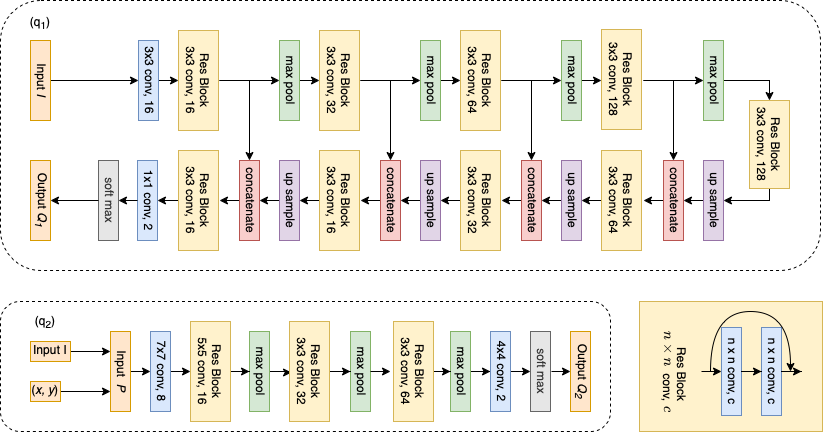}}
    \subfigure[Rot equ]{\includegraphics[width=0.6\textwidth]{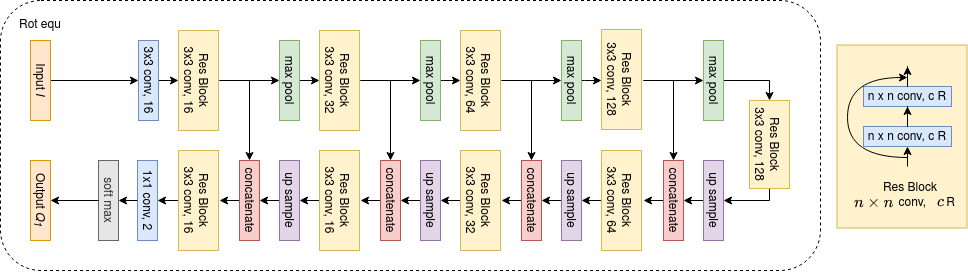}}
    \caption{The neural network architecture for ours and ablations. R means regular representation, T means trivial representation, Q means quotient representation.}
    \label{fig:q1q2_architecture}
\end{figure*}

Our network architecture is shown in Figure~\ref{fig:q1q2_architecture}a.
% The network architectures for our and no opt are shown in Figure~\ref{fig:q1q2_architecture}a.
The $q1$ network is a fully convolutional UNet \cite{u_net}. The $q2$ network is a residual neural network \cite{resnet}. These networks are implemented using PyTorch \cite{pytorch}, and the equivariant networks are implemented using the E2CNN library \cite{e2cnn}. Adam optimizer \cite{adam} is used for the SGD step. The ablation no opt has the same architecture as above. The ablation no asr (Figure~\ref{fig:q1q2_architecture}b) ablated the $q2$ network and is defined with respect to group $C16$. The ablation no equ (Figure~\ref{fig:q1q2_architecture}c) has a similar network architecture as ours with approximately the same number of free weights. However, the equivariant network is replaced with an FCN. The ablation rot equ (Figure~\ref{fig:q1q2_architecture}d) has a similar network architecture as no asr method with approximately the same number of free weights. However, the equivariant network is replaced with an FCN.

\subsection{Parameter choices}

The parameters we choose in simulation (Section~\ref{sec:simulation_experiments}) and in hardware (Section~\ref{sec:physical_robot_experiments}) are listed in tables~\ref{tab:all_method_parameters}, \ref{tab:ours_parameters}, and \ref{tab:baseline_parameters}.

\begin{table} [h]
    \caption{Parameter choices for all methods.}
    \centering
    \begin{tabular}{ m{6em} m{9em} m{9em}}
        \toprule
        environment & parameter & value \\
        \midrule
        \multirowcell{5}{in simulation,\\in hardware}
        & $bs$ batch size& 8 (2 for VPG) \\
        & number of rotations& 8 \\
        & augment buffer& $8$ times \\
        & augment buffer type& random $\SE(2)$, flip \\
        & $d_\text{dilation}$ & 4 pixels \\
        \midrule
        \multirowcell{4}{in simulation} 
        & $s_\text{threshold}$& 0.5cm \\
        & workspace size& $0.3\times 0.3$m \\
        & state $s$ size& $128^2$ pixel \\
        & action range& $96^2$ pixel \\
        \midrule
        \multirowcell{4}{in hardware}
        & $s_\text{threshold}$& 1.5cm \\
        & workspace size& $0.25\times 0.25$m \\
        & state $s$ size& $112^2$ pixel \\
        & action range& $80^2$ pixel \\
        \bottomrule
    \end{tabular}
    \label{tab:all_method_parameters}
\end{table}

\begin{table}[h]
    \caption{Parameter choices for ours.}
    \centering
    \begin{tabular}{ m{6em} m{9em} m{9em}}
        \toprule
        environment & parameter & value \\
        \midrule
        \multirowcell{9}{in simulation,\\in hardware}
        & train SGD step after& the 20th grasp \\
        & policy & Boltzmann \\
        & $crop(s, x)$ size& 32 \\
        & learning rate& 1e-4 \\
        & weight decay & 1e-5 \\
        & $k$ in $L_1^{''}$& 10 \\
        & $\tau$ in $L_1^{''}$& 1 \\
        & $\tau_\text{test}$& 0.01 \\
        & $\tau_\text{train}$& 0.002 \\
        & SGD step per grasps& 1 \\
        \bottomrule
    \end{tabular}
    \label{tab:ours_parameters}
\end{table}

\begin{table} [h]
    \caption{Parameter choices for baselines.}
    \centering
    \begin{tabular}{ m{6em} m{9em} m{10em}}
        \toprule
        environment & parameter & value \\
        \midrule
        \multirowcell{4}{in simulation,\\in hardware}
        & train SGD step after& the 1st grasp \\
        & policy & $\epsilon$-greedy \\
        & $\epsilon_\text{initial}$& 0.5 \\
        & $\epsilon_\text{final}$& 0.1 \\
        & $\epsilon$ linear schedule & 200 grasps in simulation \\
        & & 500 grasps in hardware \\
        \bottomrule
    \end{tabular}
    \label{tab:baseline_parameters}
\end{table}

\subsection{Augmentation baseline choices}
\label{sec:aug_baseline}

The data augmentation strategies are: \underline{$n\times$ RAD:} The method from~\cite{rad} where we perform $n$ SGD steps after each grasp sample, where each SGD step is taken over a mini-batch $bs$ of samples that have been randomly translated and rotated by $g\in \hat{\SE}(2)$.
% \hl{for both observation and action: $(\mathrm{T}s, \mathrm{T}a), \mathrm{T} \in \hat{\SE}(2)$. This augmentation assumes a perturbed sample is valid $r(s, a) = r(\mathrm{T}s, \mathrm{T}a)$.}
\underline{$n\times$ soft equ:} a data augmentation method \cite{wang2021equivariant} that performs $n$ soft equivariant SGD steps per grasp, where each SGD step is taken over $n$ times randomly $\hat{\SE}(2)$ augmented mini-batch.
% $(\mathrm{T}s, \mathrm{T}a)$.
% Specifically, we sample one transition from the replay buffer and augment it $n$ times, then perform one SGD step
Specifically, we sample $bs/n$ samples ($bs$ is the batch size), augment it $n$ times and train on this mini-batch. We perform this SGD steps $n$ times so that $bs$ transitions are sampled. This augmentation aims at achieving equivariance in the mini-batch. 

\begin{figure*}
    \centering
    \subfigure[]{\includegraphics[width=0.22\textwidth]{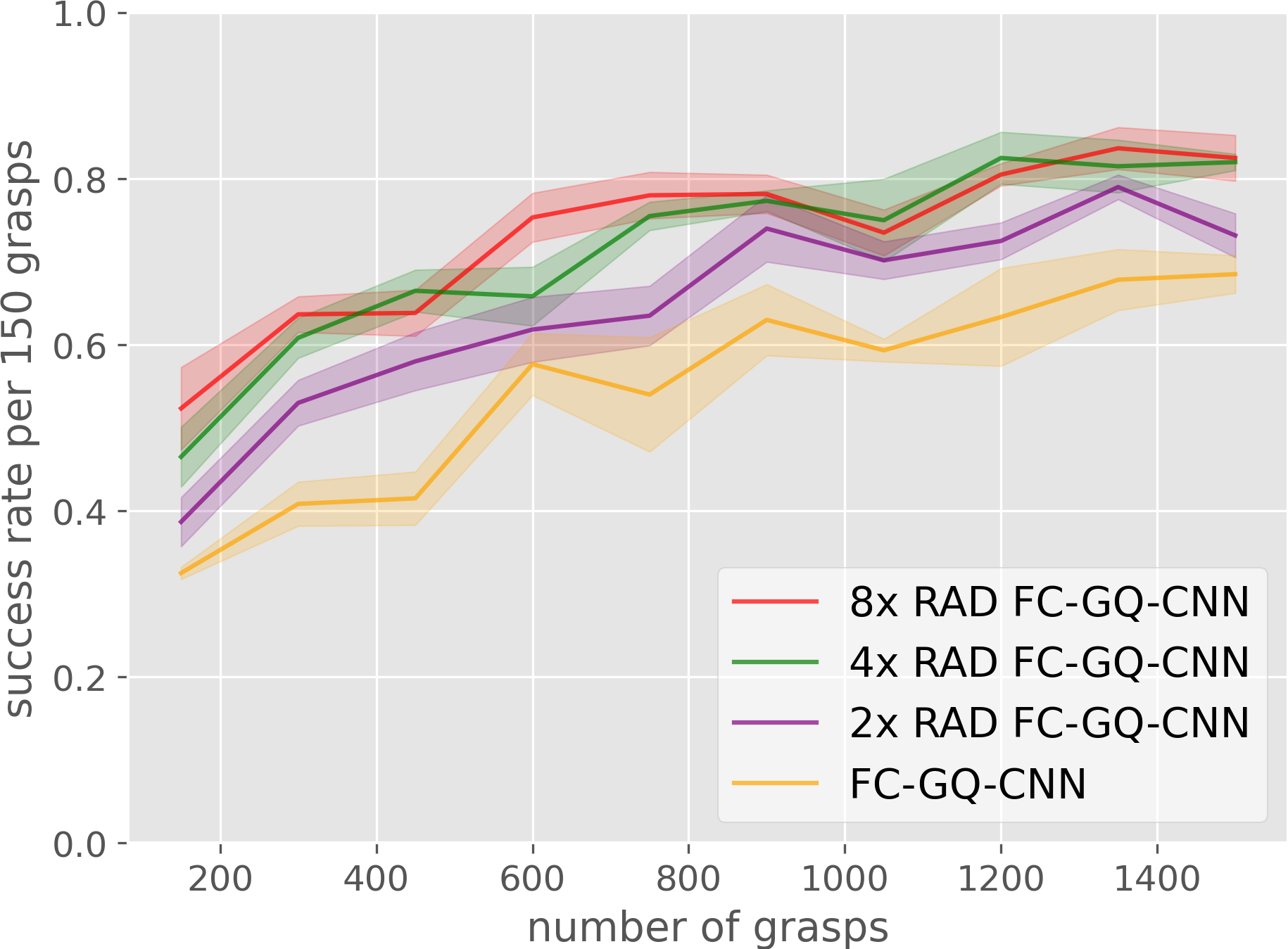}}
    \subfigure[]{\includegraphics[width=0.22\textwidth]{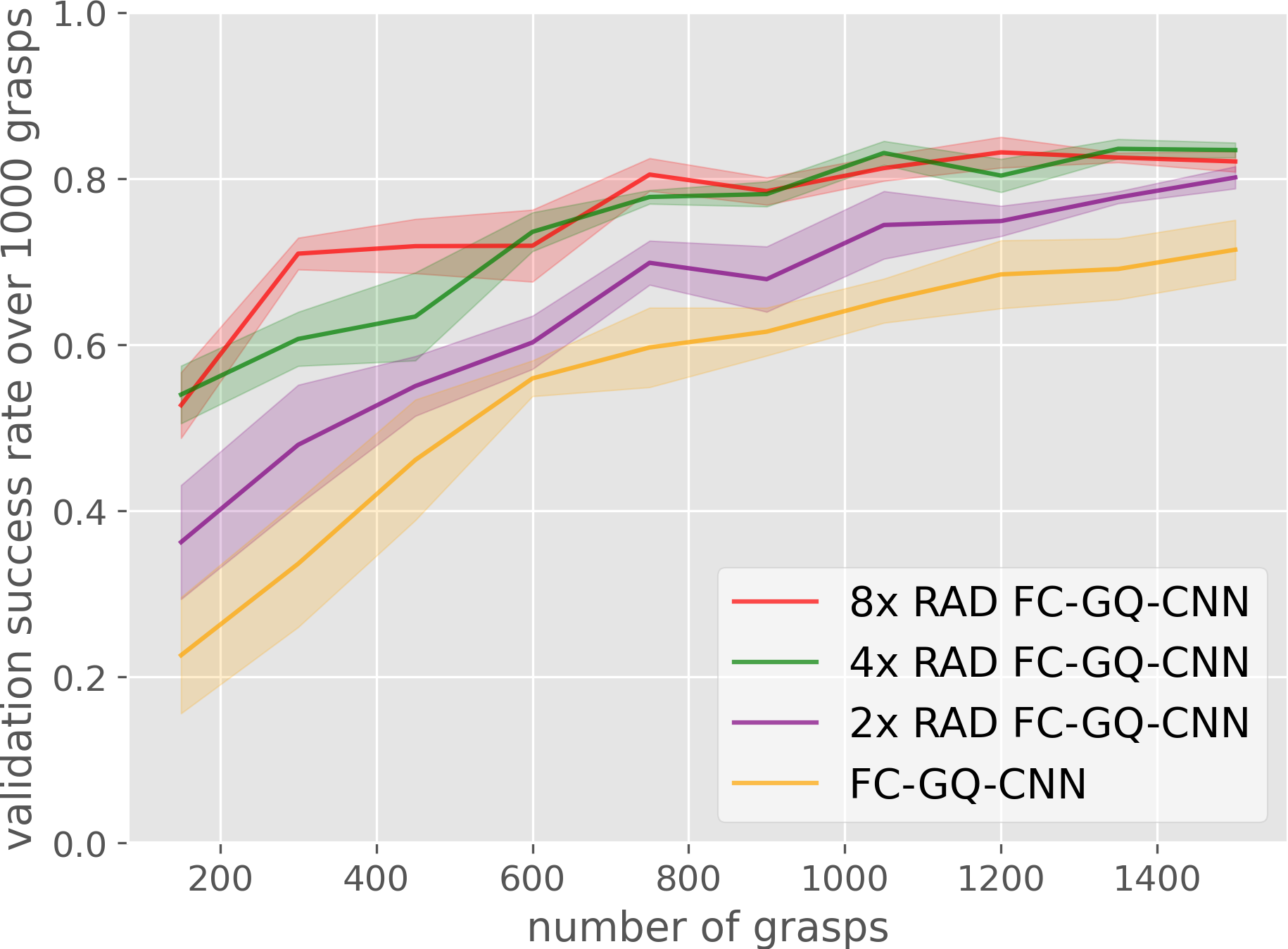}}
    \subfigure[]{\includegraphics[width=0.22\textwidth]{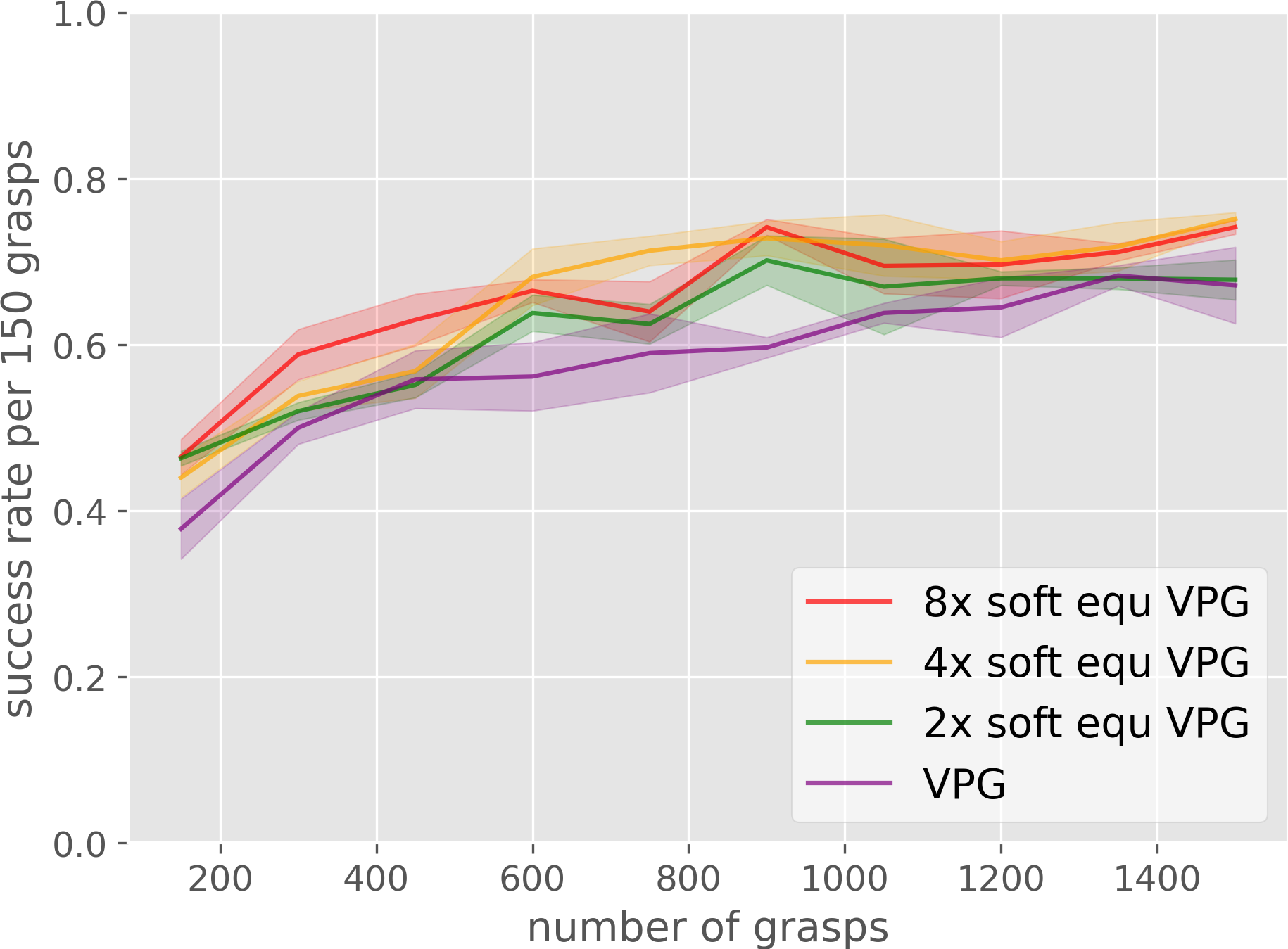}}
    \subfigure[]{\includegraphics[width=0.22\textwidth]{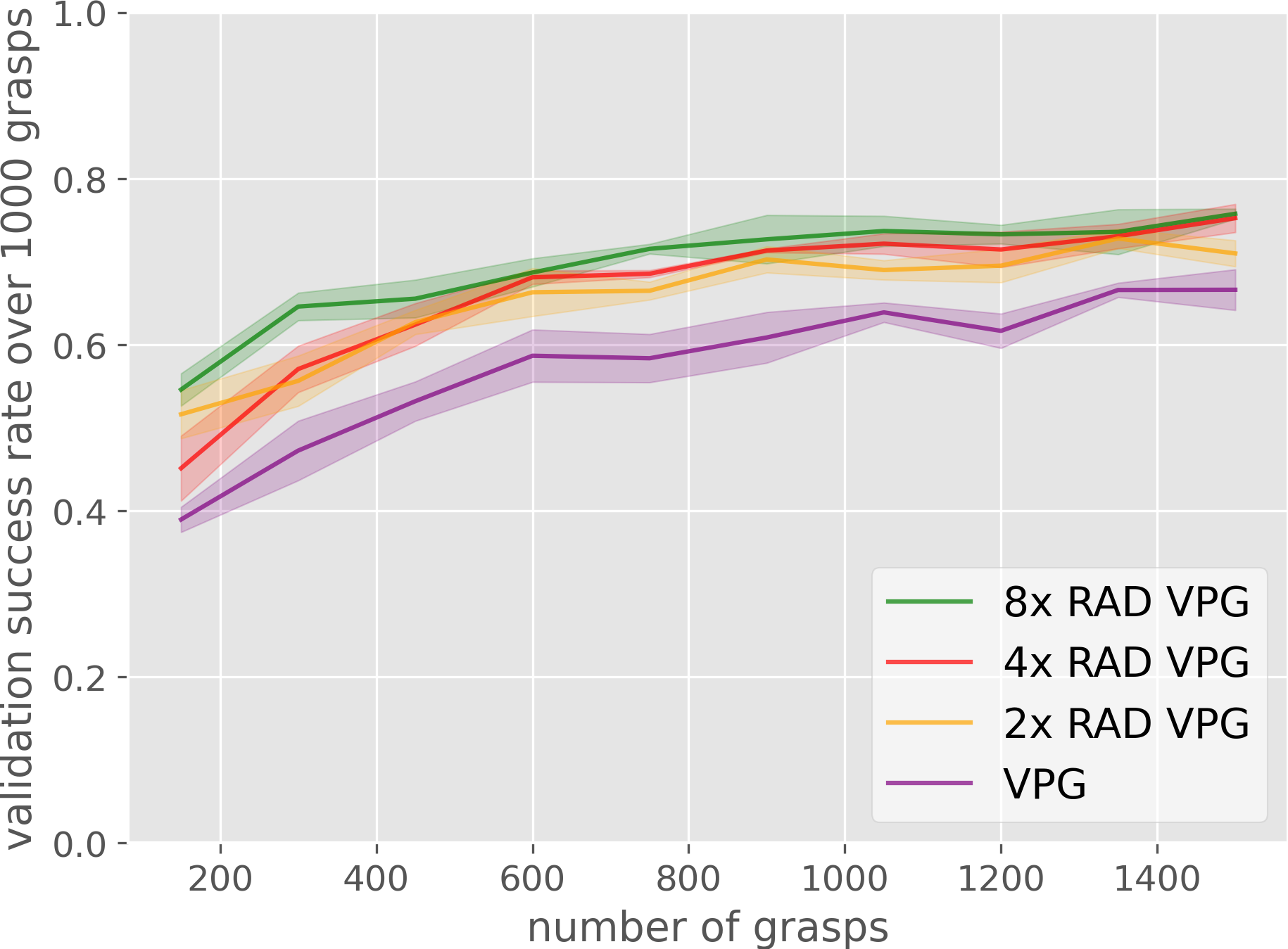}}
    \subfigure[]{\includegraphics[width=0.22\textwidth]{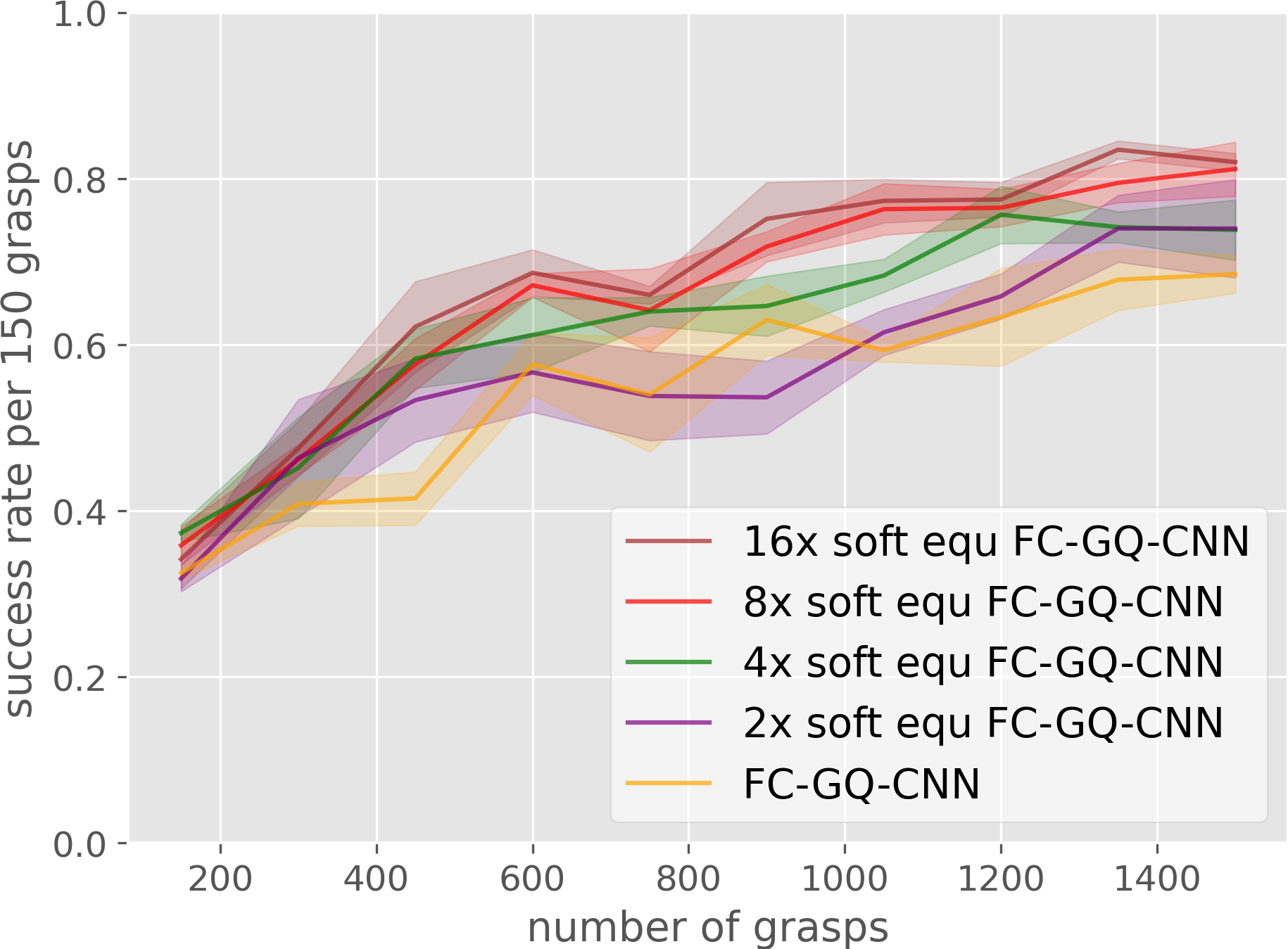}}
    \subfigure[]{\includegraphics[width=0.22\textwidth]{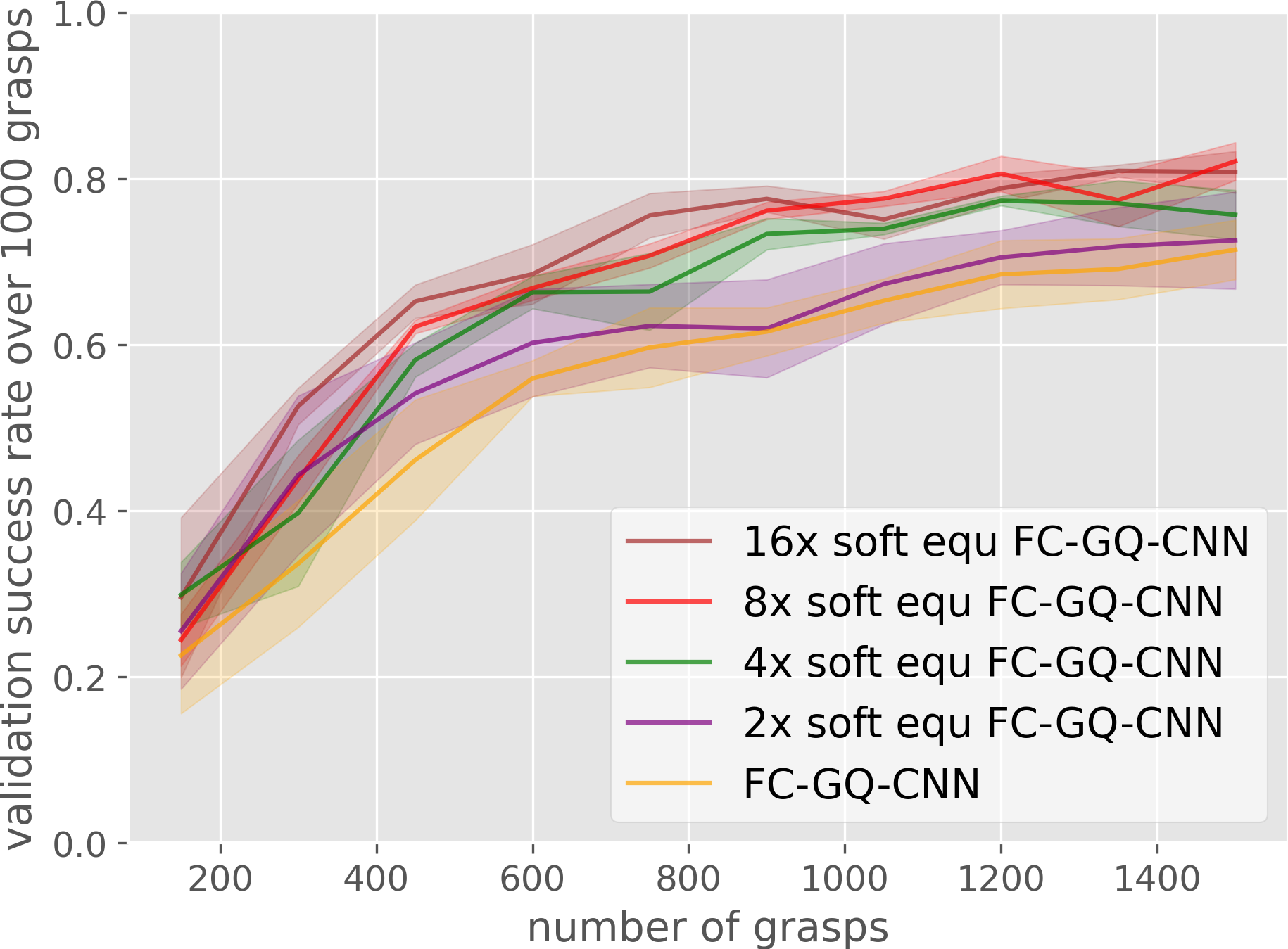}}
    \subfigure[]{\includegraphics[width=0.22\textwidth]{figure/appendix/n_soft_VPG.png}}
    \subfigure[]{\includegraphics[width=0.22\textwidth]{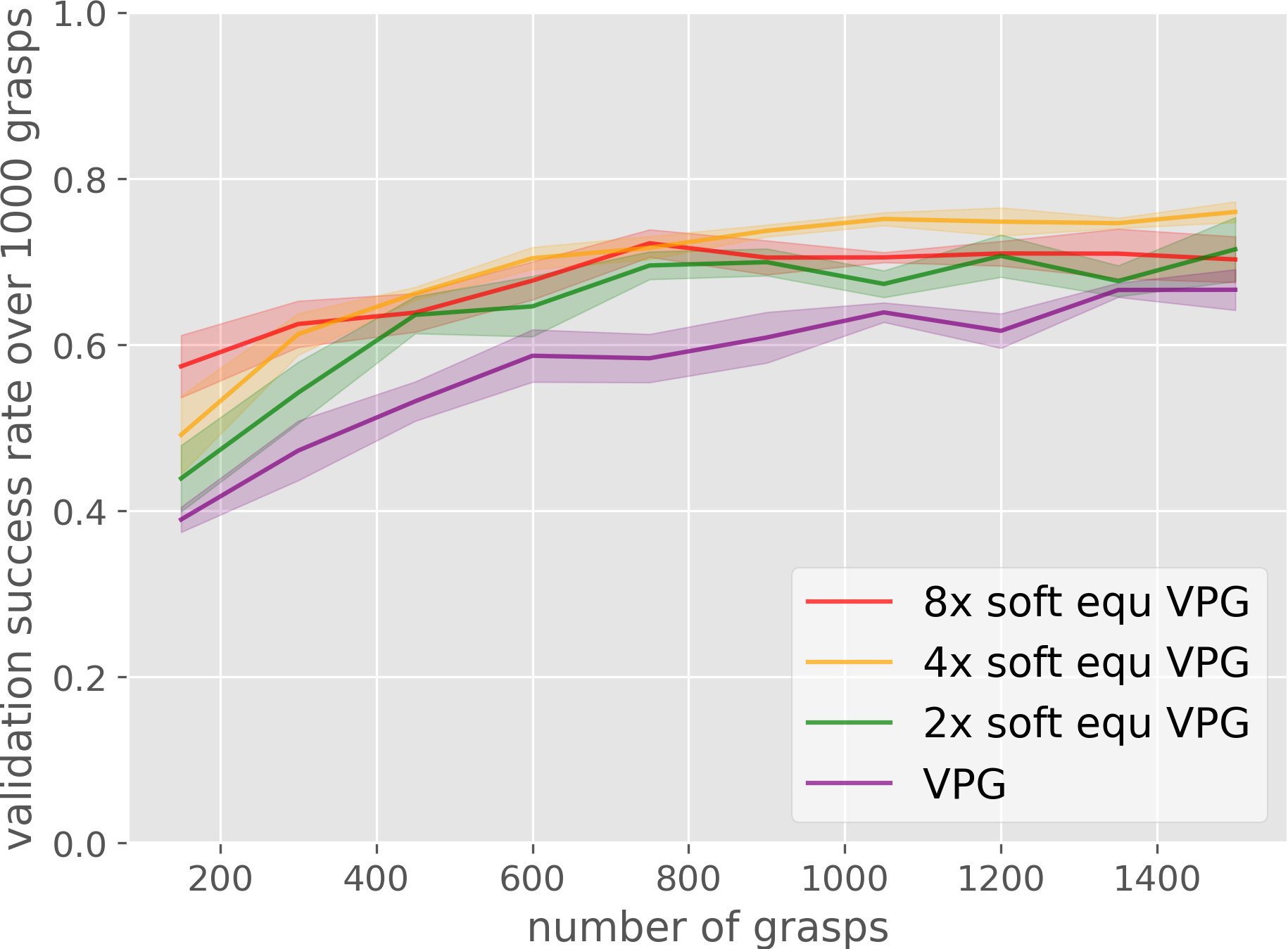}}
    \caption{Baseline with data augmentation. The first two columns are FC-GQ-CNN while the last two columns are VPG. The first row is $n\times$ RAD augmentation whereas the last row is $n\times$ soft equ augmentation. The baselines without augmentation prefix are the baselines without augmentation.}
    \label{fig:baseline_with_augmentation}
\end{figure*}

We apply $n\times$ RAD and $n\times$ soft equ data augmentation to both VPG and FC-GQ-CNN baselines, with $n={2, 4, 8}$. Figure~\ref{fig:baseline_with_augmentation} shows the results. Observe that all data augmentation choices improve the baselines, but an increase in $n$ leads to a saturation effect in learning while causing more computation overhead. The best data augmentation parameters $n$ are chosen for each baseline in the comparison in Figure~\ref{fig:simulation_baseline}

\begin{figure}
    \centering
    \subfigure[Training]{\includegraphics[width=0.22\textwidth]{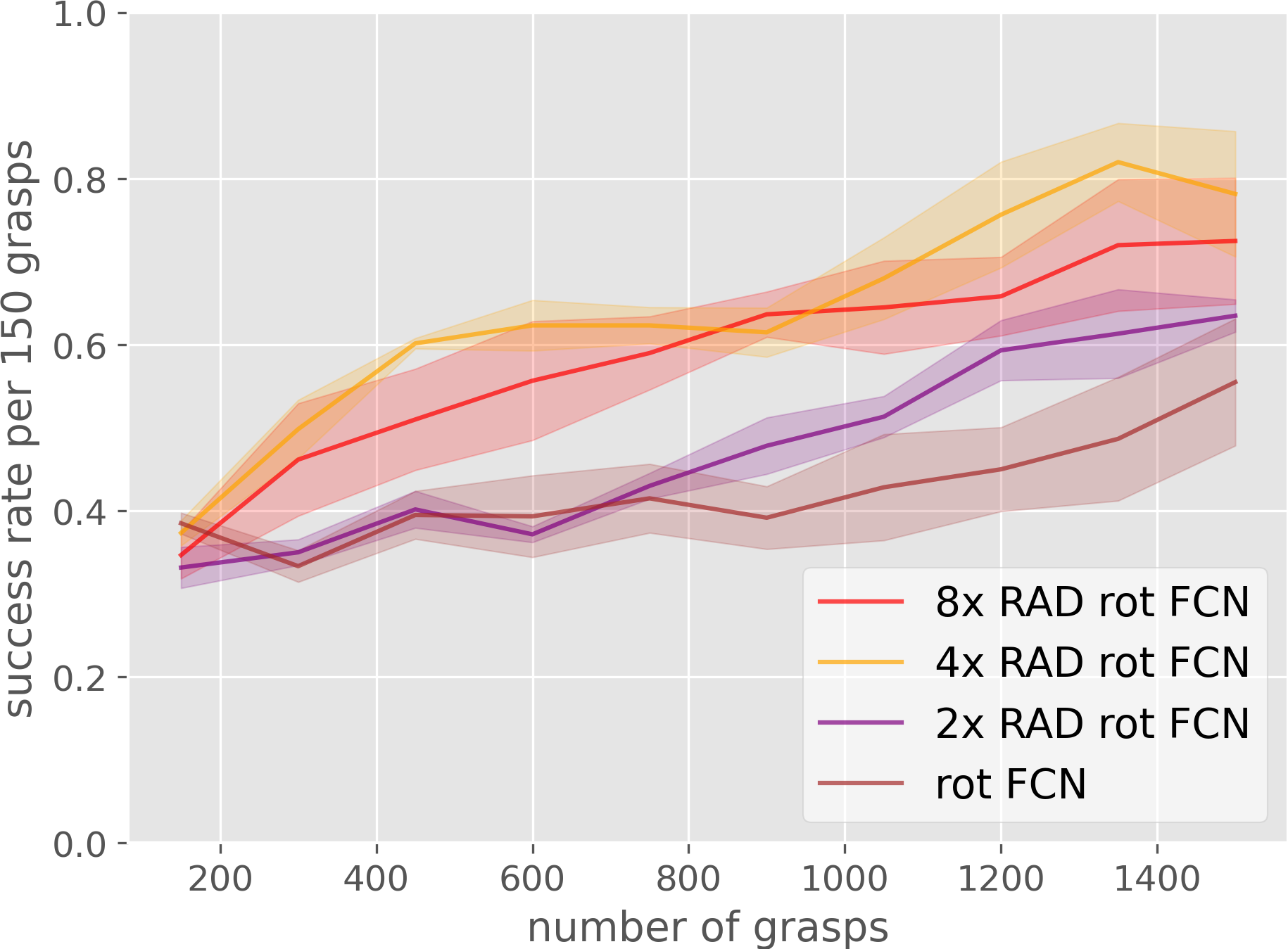}}
    \subfigure[Testing]{\includegraphics[width=0.22\textwidth]{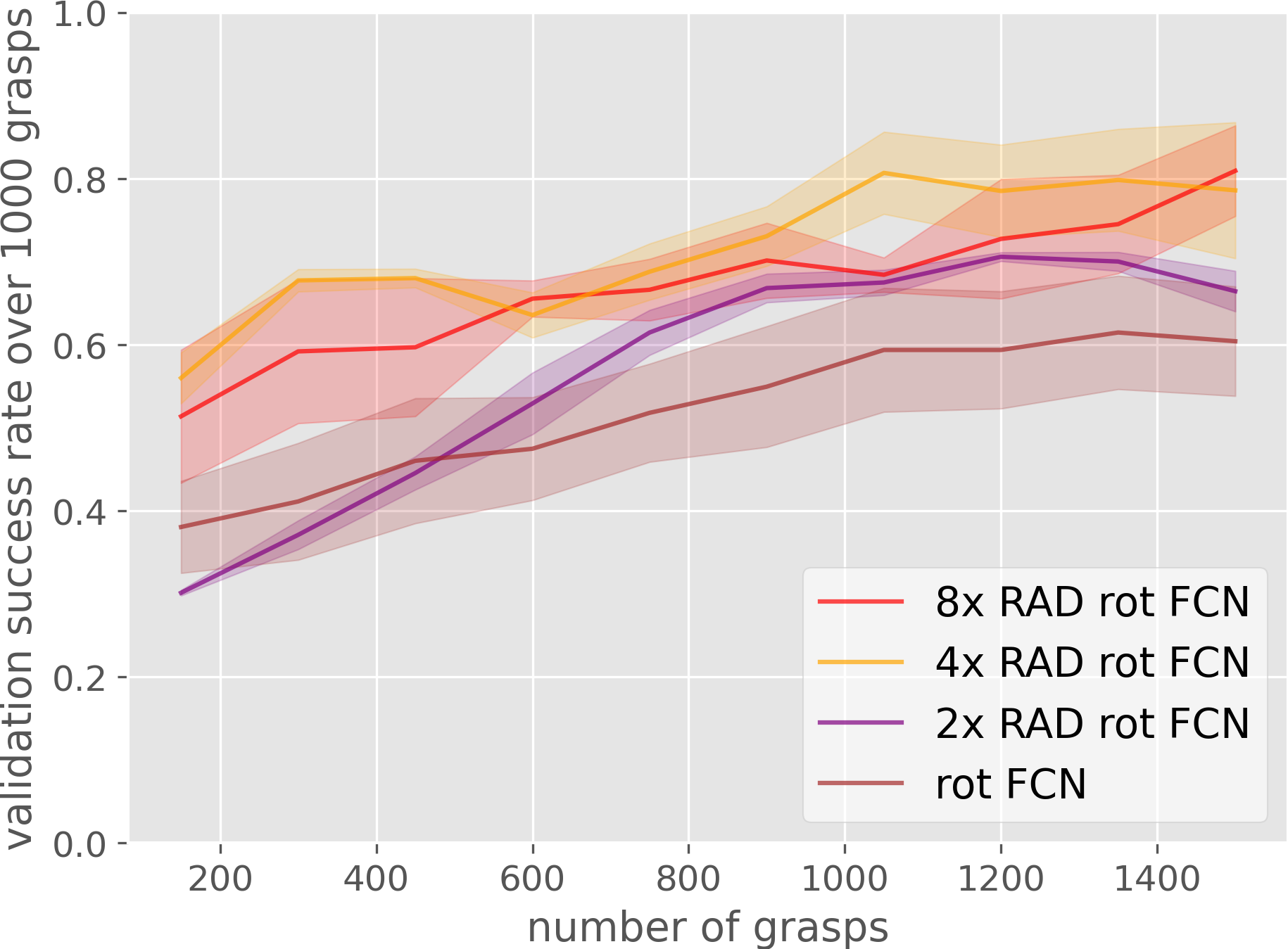}}
    \caption{Baseline comparisons for Rot equ. We refer rot equ as the best learning curve in (b), i.e, $4\times$ RAD rot FCN.}
    \label{fig:rot_fcn_augmentation}
\end{figure}

For rot equ ablation baseline, we choose the best learning curve in Figure~\ref{fig:rot_fcn_augmentation}b, i.e, $4\times$ RAD rot FCN. This baseline aims to achieve equivariance through the combination of data augmentation and rotation encoding of an FCN.

\subsection{Success and failure modes}

We list typical success and failure modes to evaluate our algorithm's performance.

\begin{figure*}
    \centering
    \subfigure[]{\includegraphics[width=0.2\textwidth]{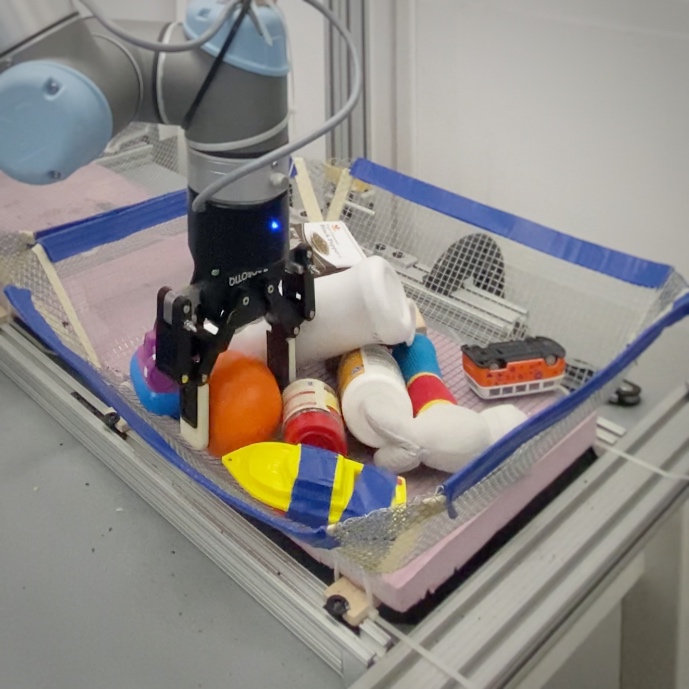}}
    \subfigure[]{\includegraphics[width=0.2\textwidth]{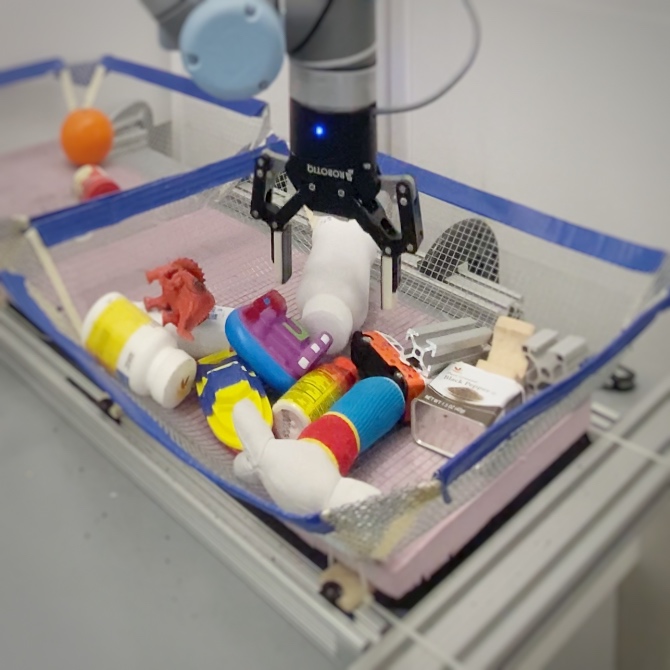}}
    \subfigure[]{\includegraphics[width=0.2\textwidth]{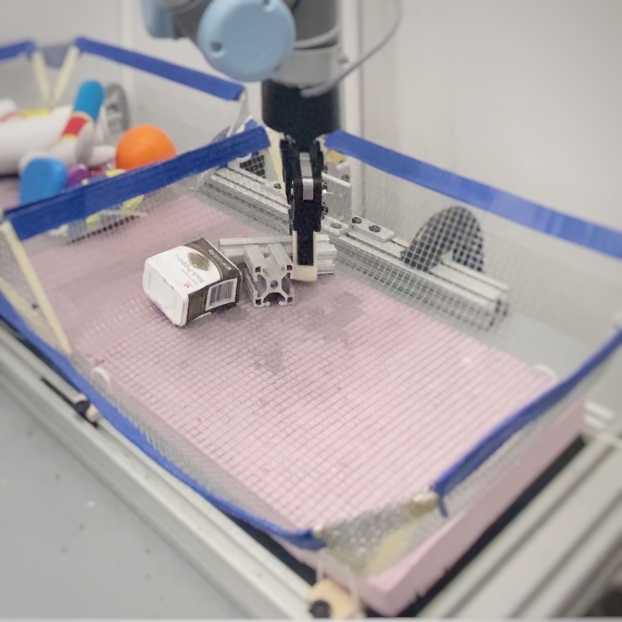}}
    \subfigure[]{\includegraphics[width=0.2\textwidth]{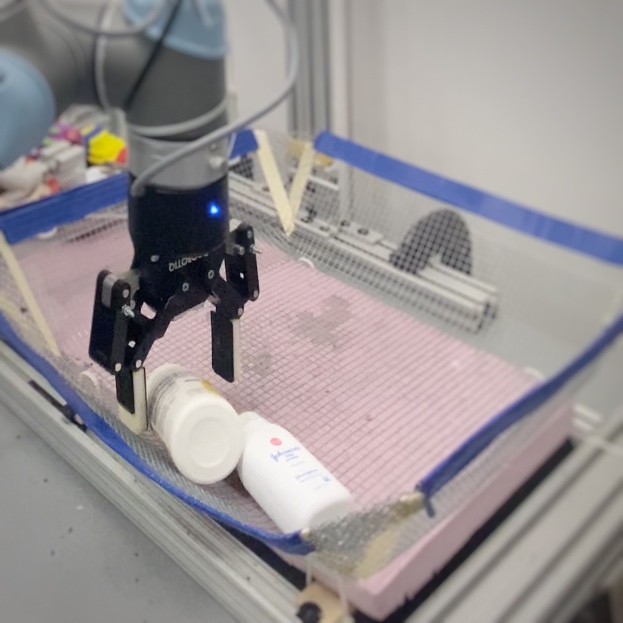}}
    \caption{Success modes in the test set easy, which has 15 hold out objects.}
    \label{fig:success_modes}
\end{figure*}

For success modes, the learned policy of our method showcases its intelligence. At the densely cluttered scene, our method prefers to grasp the relatively isolated part of the objects, see Figure~\ref{fig:success_modes}a, b. At the scene where the objects are close to each other, our method can find the grasp pose that doesn't cause a collision/interference with other objects, see Figure~\ref{fig:success_modes}c, d.

\begin{figure*}
    \centering
    \subfigure[Wrong $q_1$ (10/20)]{\includegraphics[width=0.22\textwidth]{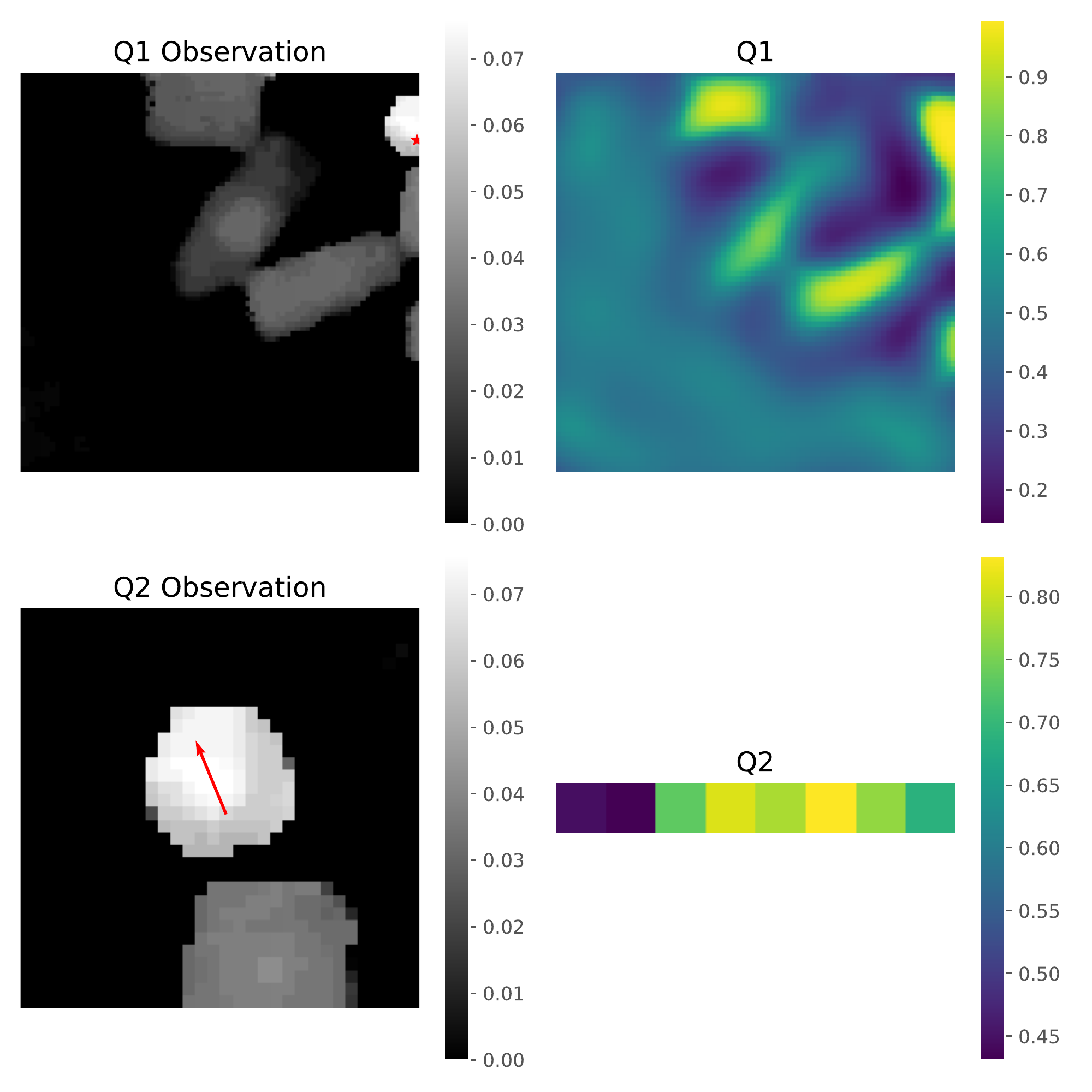}}
    \subfigure[Wrong $q_2$ (3/20)]{\includegraphics[width=0.22\textwidth]{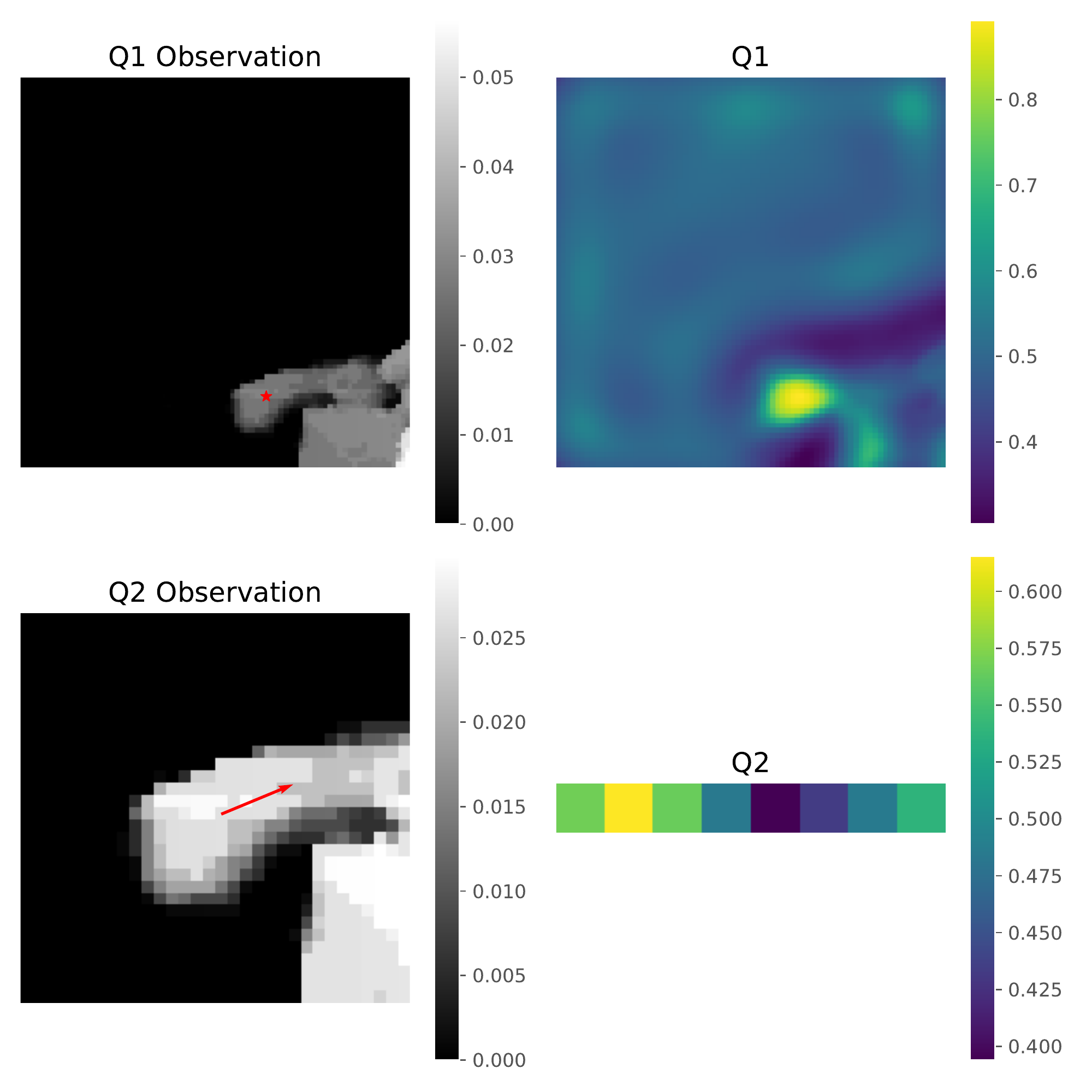}}
    \subfigure[Challenging scenes (3/20)]{\includegraphics[width=0.22\textwidth]{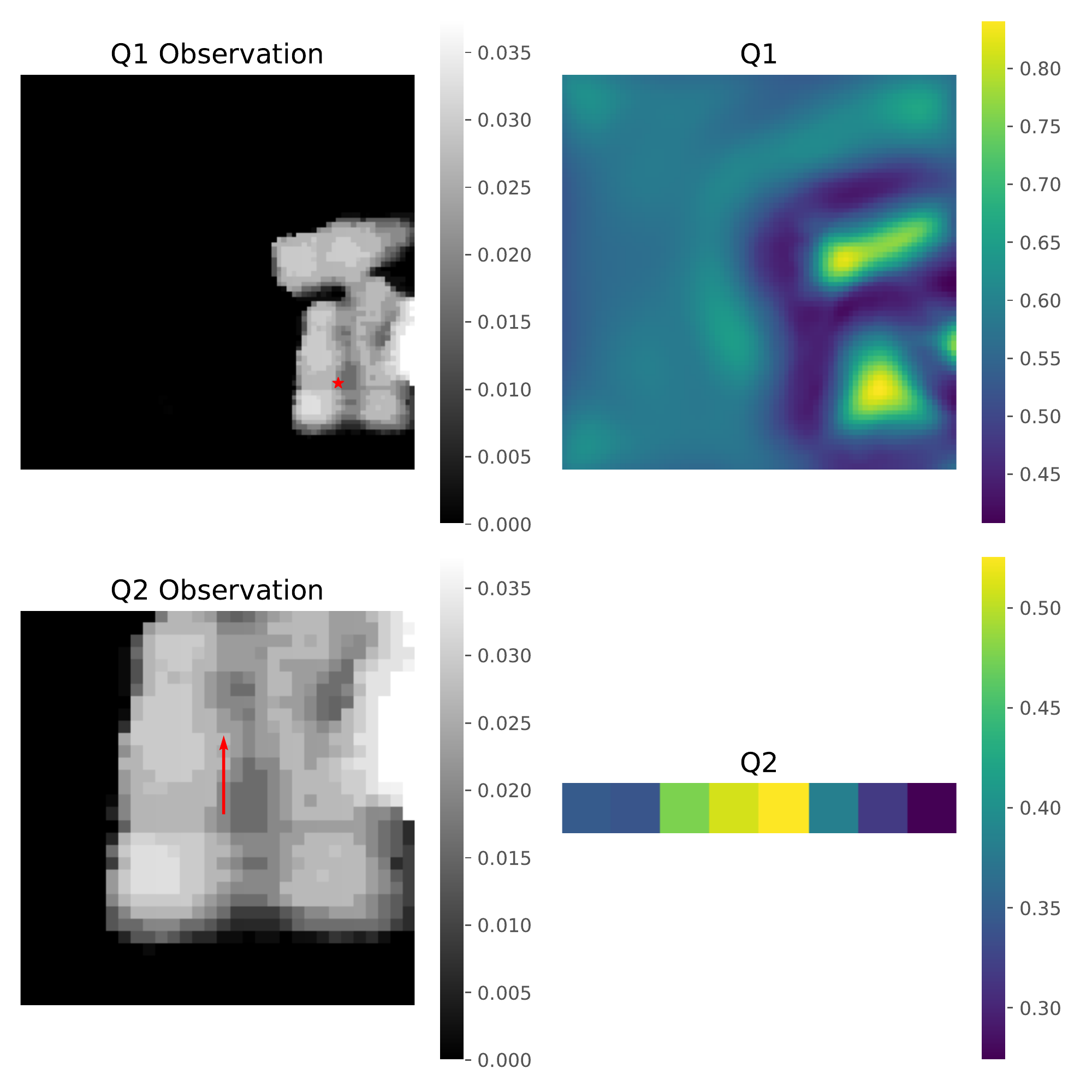}}
    \subfigure[Reasonable grasps (3/20)]{\includegraphics[width=0.22\textwidth]{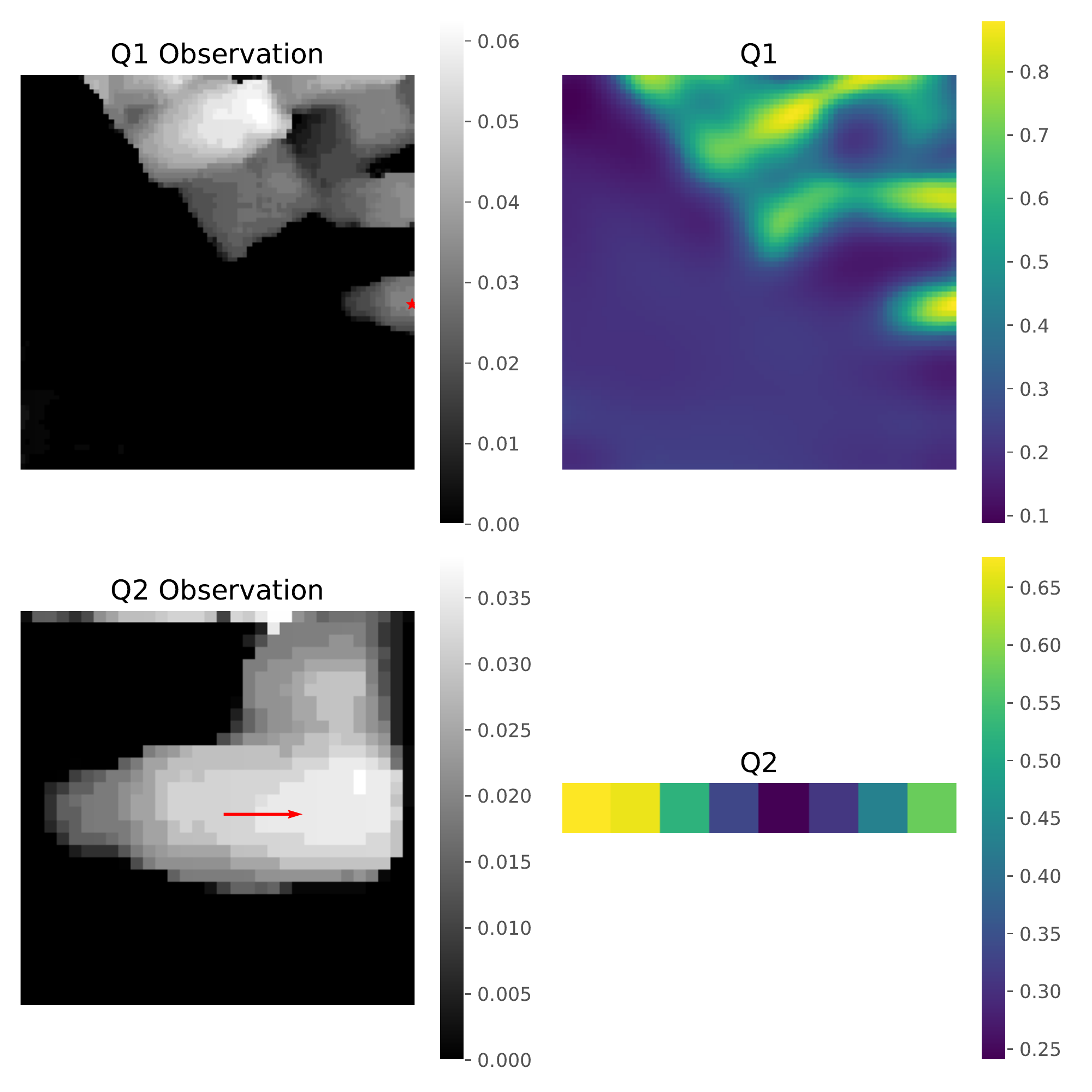}}
    \subfigure[Wrong $q_1$ (17/52)]{\includegraphics[width=0.22\textwidth]{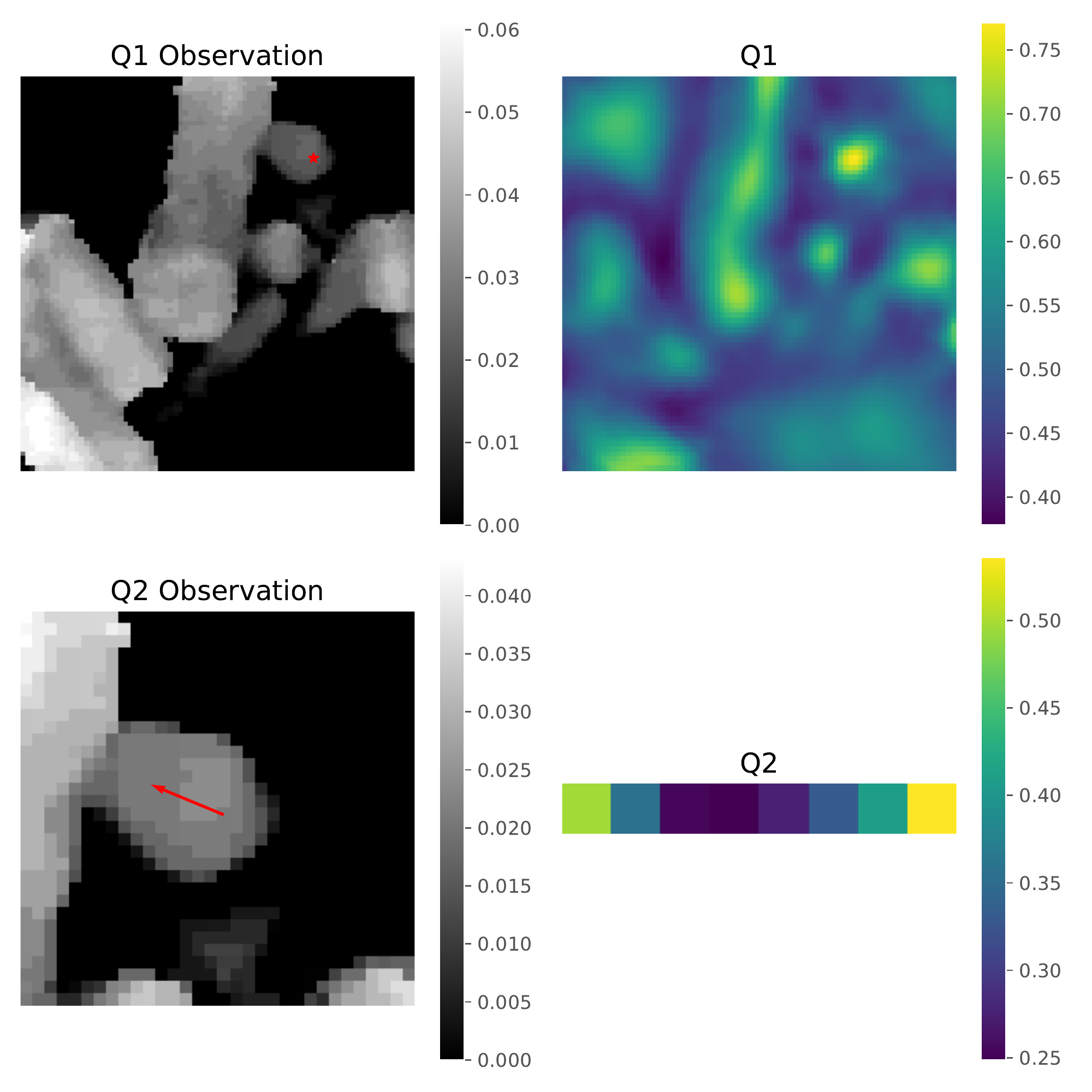}}
    \subfigure[Reasonable grasps (13/52)]{\includegraphics[width=0.22\textwidth]{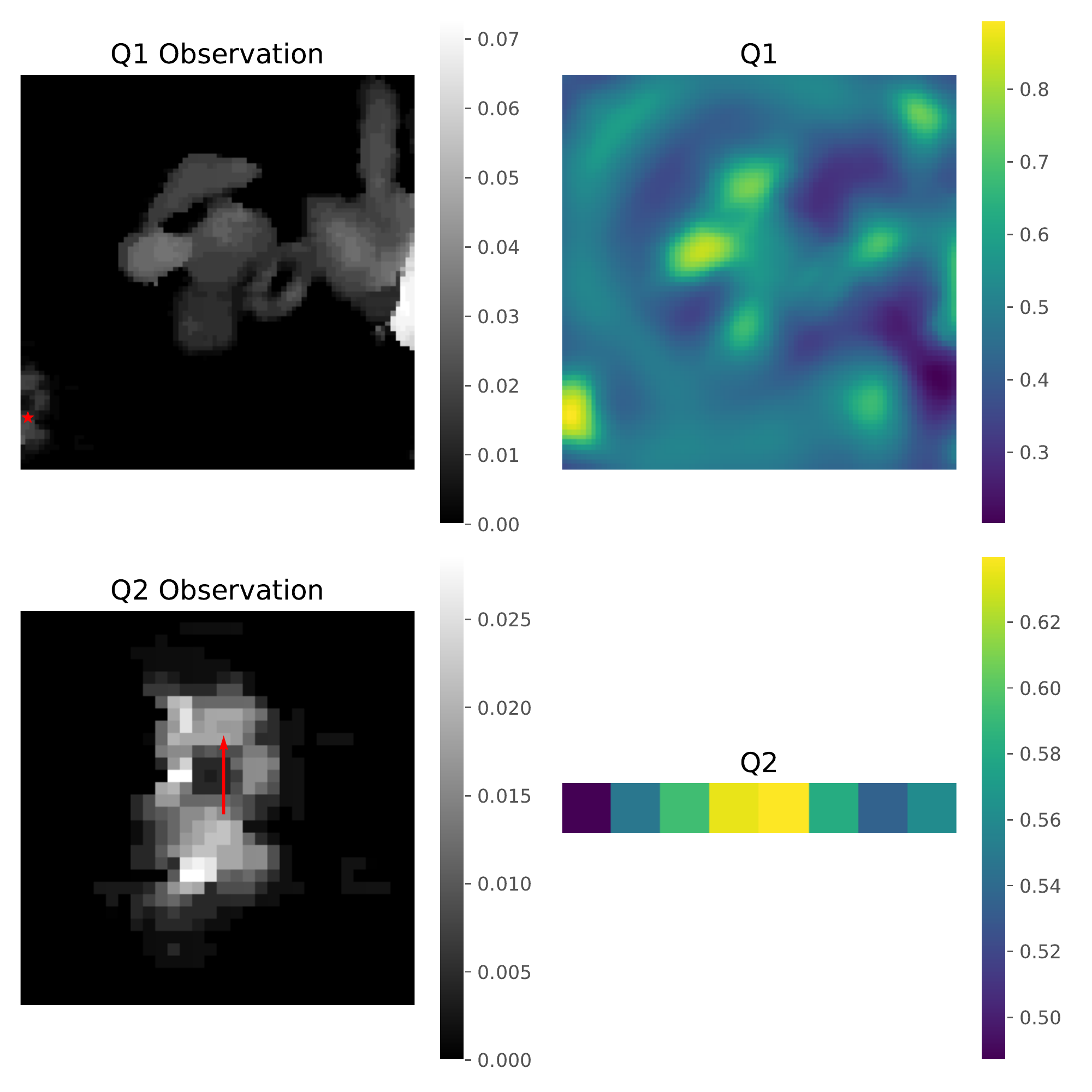}}
    \subfigure[Challenging scenes (11/52)]{\includegraphics[width=0.22\textwidth]{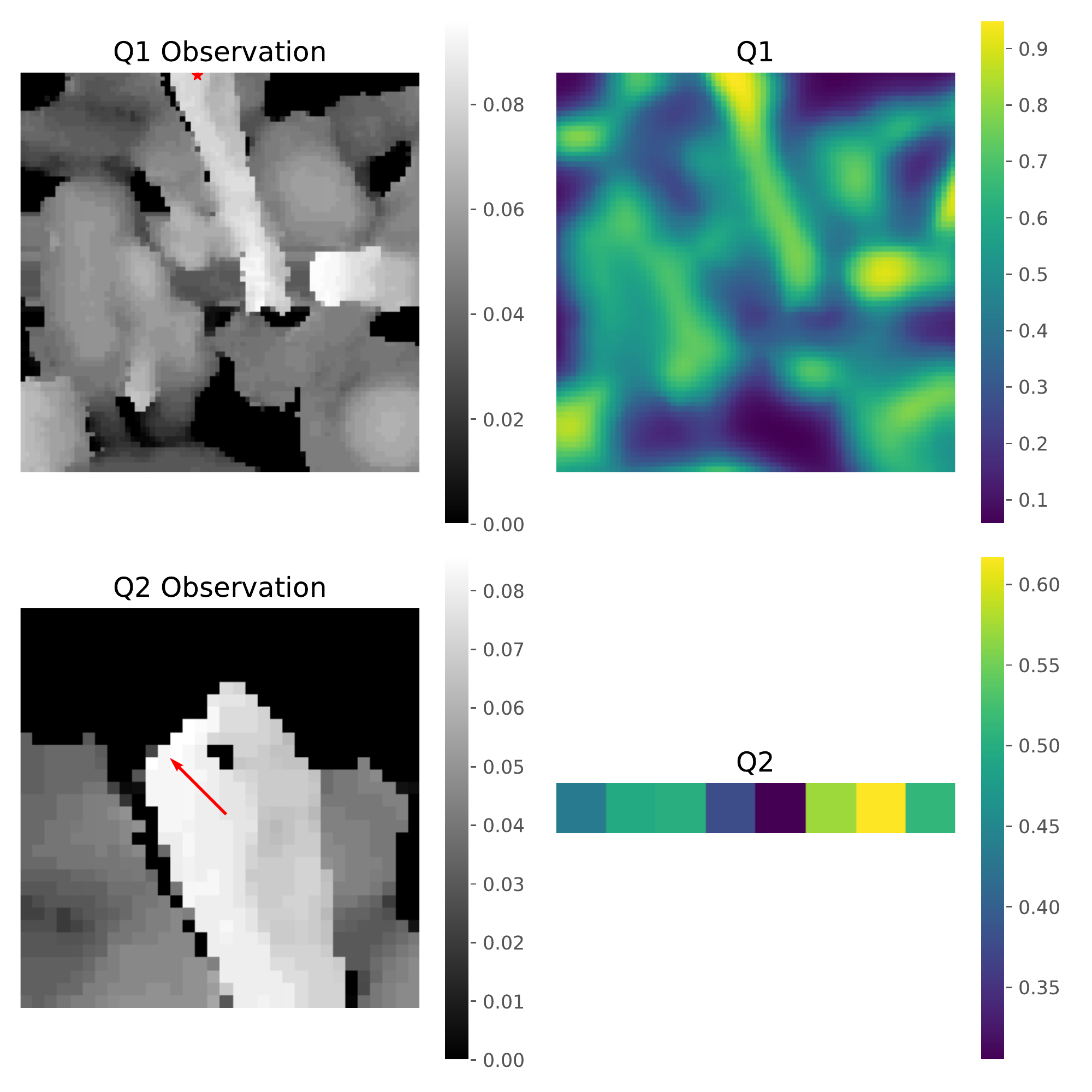}}
    \subfigure[Sensor distortion (8/52)]{\includegraphics[width=0.22\textwidth]{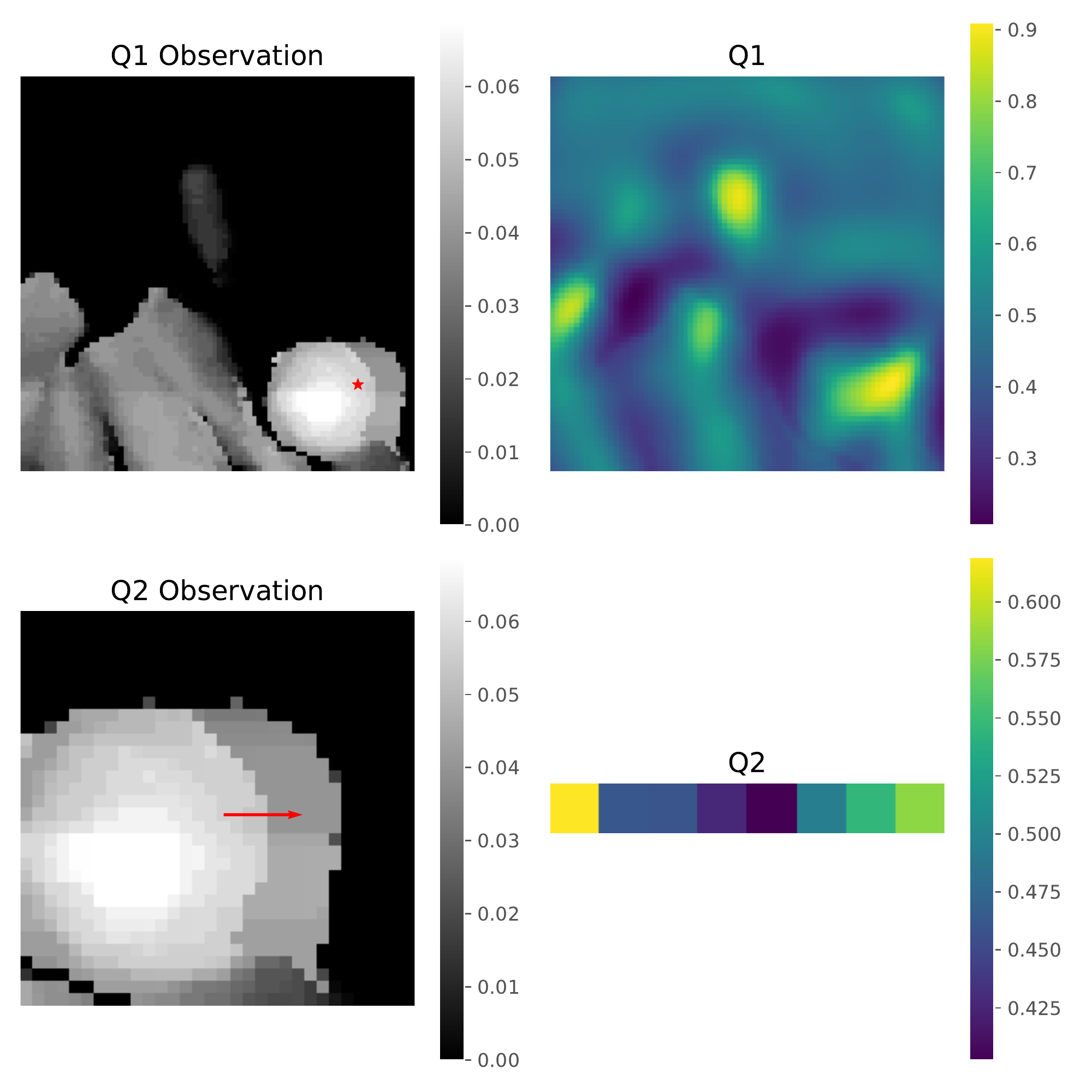}}
    \caption{Failure modes. The brackets show the failure times divided by the total number of failures in all four runs. The first row is test in the test set easy while the second row is test in the test set hard.}
    \label{fig:failure_modes}
\end{figure*}

For failure modes, we identify several typical scenarios: Wrong action selection (Figure~\ref{fig:failure_modes}a, b, and e) indicates that there is a clear gap between our method and optimal policy, this might be caused by the biased dataset collected by the algorithm. Reasonable grasps failure (Figure~\ref{fig:failure_modes}d, f) means that the agent selects a reasonable grasp, but it fails due to the stochasticity of the real world, i.e., sensor noise, contact dynamics, hardware flaws, etc. Challenging scenes (Figure~\ref{fig:failure_modes}c, g) is the nature of densely cluttered objects, it can be alleviated by learning an optimal policy or executing higher DoFs grasps. The sensor distortion (Figure~\ref{fig:failure_modes}h) is caused by an imperfect sensor. Among all failure modes, wrong action selection takes the most part (65\% failure in the test set easy and 33\% failure in the test set hard). It is followed by reasonable grasps, challenging scenes, and then sensor distortion.

\begin{figure}
    \centering
    \includegraphics[width=0.3\textwidth]{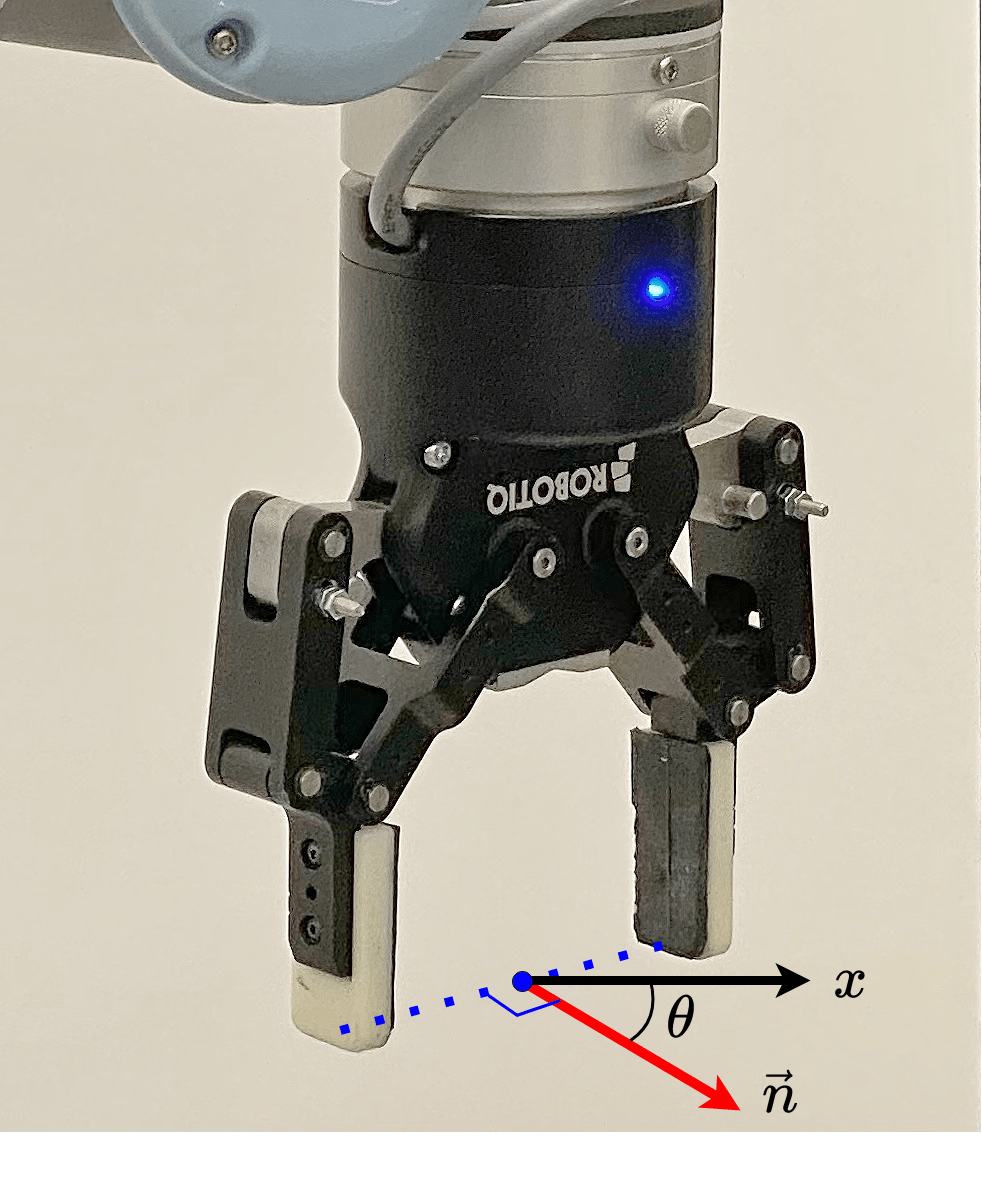}
    \caption{The definition of $\theta$. $x$ is the $x$-axle of the workspace while $\vec{n}$ is the normal of the gripper.}
    \label{fig:gripper}
\end{figure}

\subsection{Action space details}

The action $\theta$ is defined as the angle between the normal vector $\vec{n}$ of the gripper and the $x$-axle, see Figure~\ref{fig:gripper}.

\begin{figure*}
    \centering
    \subfigure[]{\includegraphics[width=0.13\textwidth]{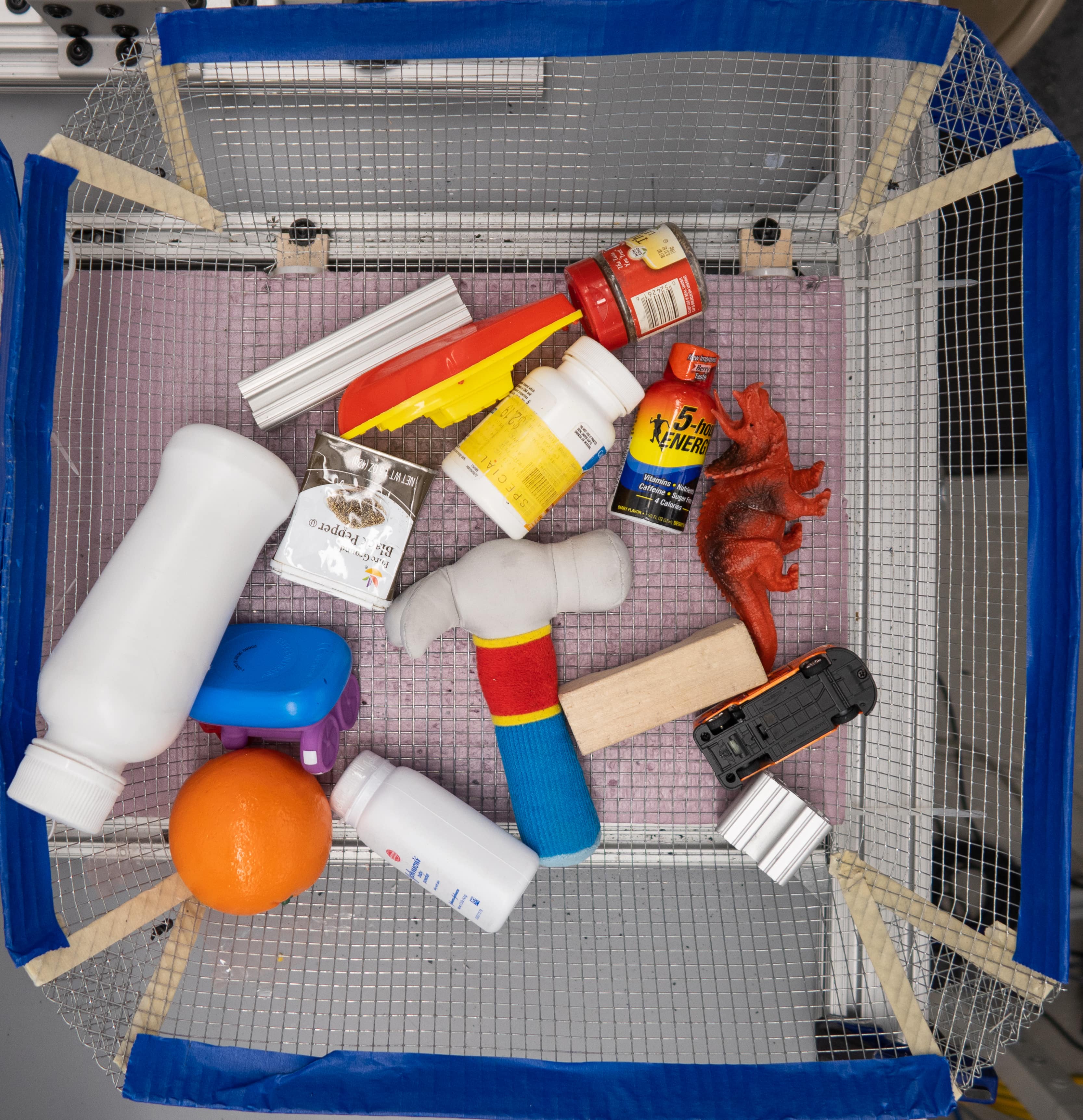}}
    \subfigure[]{\includegraphics[width=0.16\textwidth]{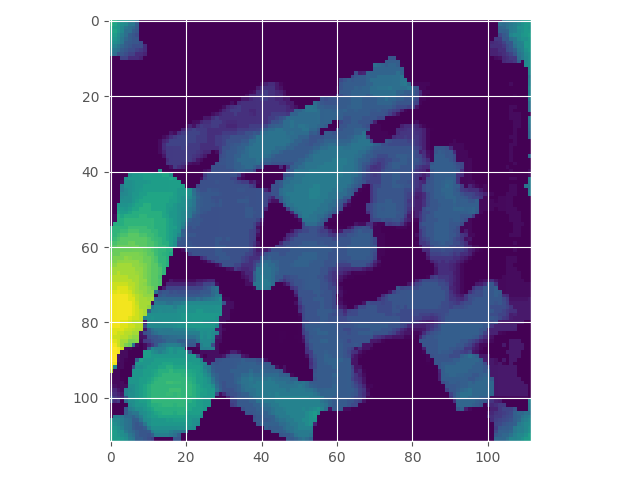}}
    \subfigure[]{\includegraphics[width=0.16\textwidth]{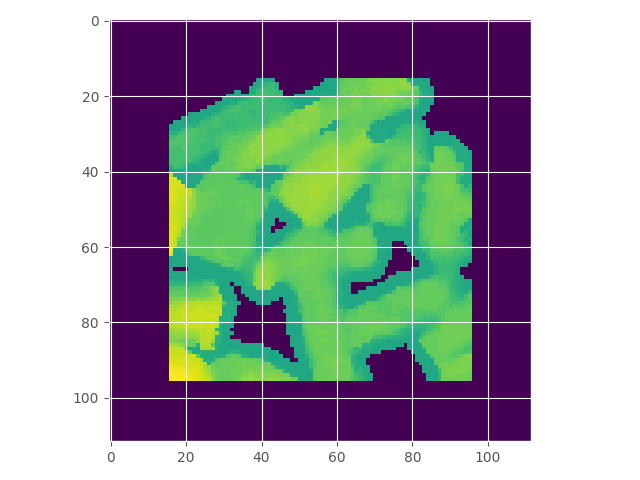}}
    \subfigure[]{\includegraphics[width=0.15\textwidth]{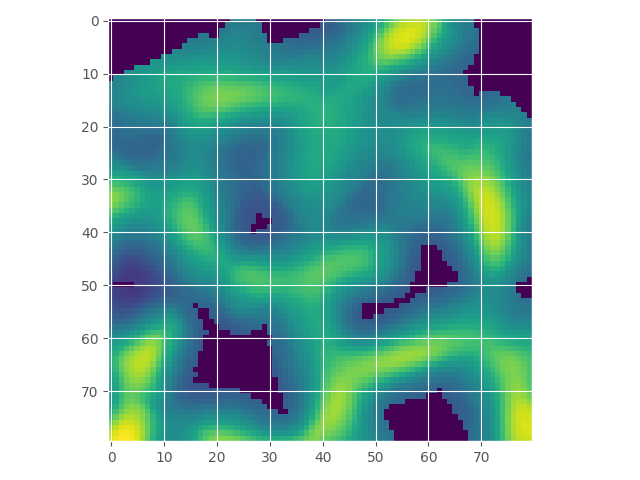}}
    \subfigure[]{\includegraphics[width=0.13\textwidth]{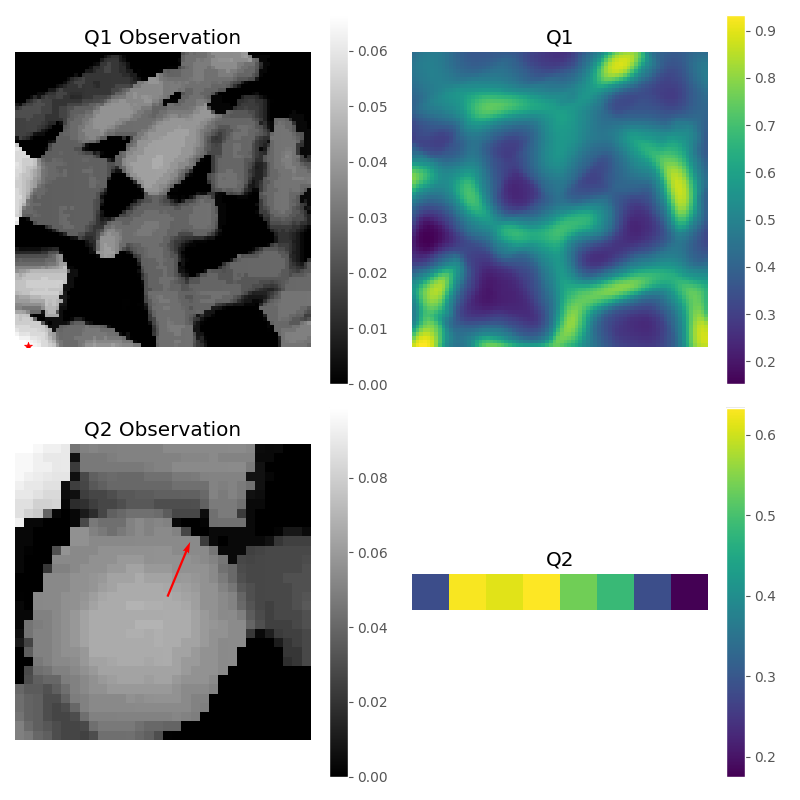}}
    \subfigure[]{\includegraphics[width=0.18\textwidth]{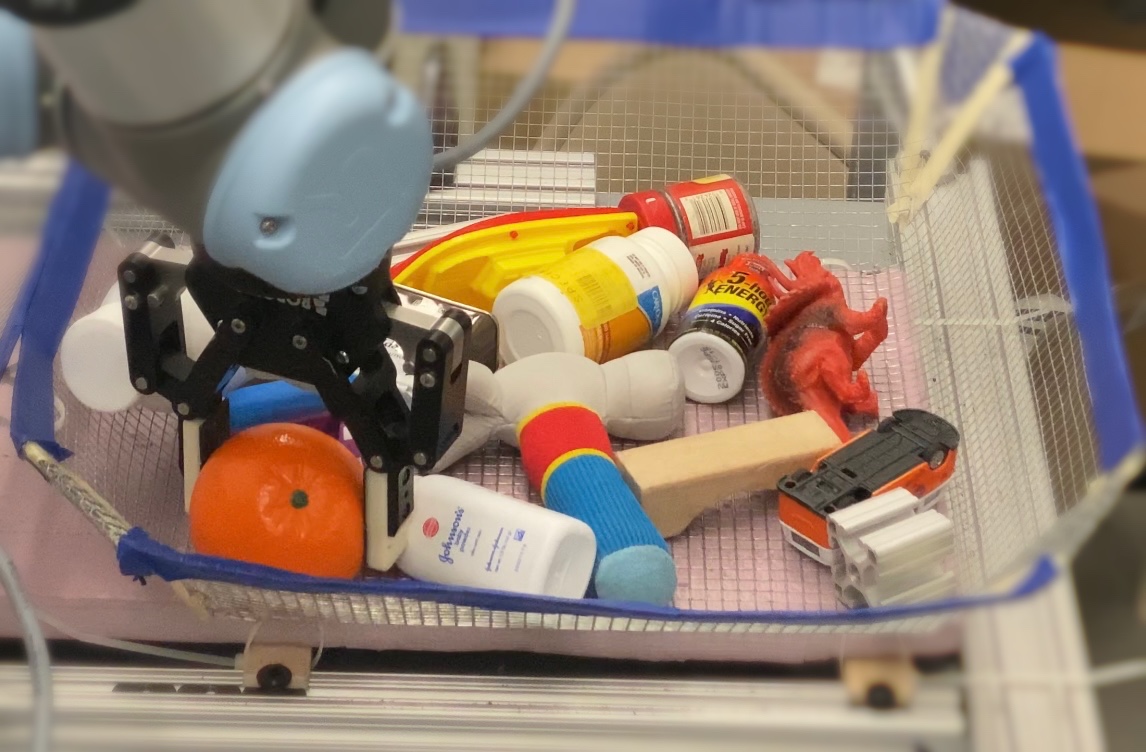}}
    \caption{Action space constraint for action selection. (a) Test set easy cluttered scene. (b) The state $s$. (c) The action space $x_\text{positive}$, it overlays the binary mask $x_\text{positive}$ with the state $s$ for visualization. (d) The $Q$-values within the action space. (e) Selecting an action. (f) Executing a grasp.}
    \label{fig:action_space}
\end{figure*}

To prevent the grasps in the empty space where there is no object in $s$, we constrain the action space to $x_\text{positive} \in X$ to exclude the empty space, see Figure~\ref{fig:action_space}c. The constrain $x_\text{positive}$ is achieved by first thresholding the depth image: $s_\text{positive} = s > s_\text{threshold}$ ($s_\text{threshold}$ is 0.5cm in simulation and 1.5cm in hardware), then dilate this binary map $s_\text{positive}$ by radius $d_\text{dilation} = 4$ pixels. The parameter $s_\text{threshold}$ is selected according to sensor noise where $d_\text{dilation}$ is related to the half of the gripper aperture. Moreover, we constrain the action space within the tray to prevent collision.

\subsection{Evaluation details in hardware}
\label{sec:evaluation_details}

The evaluation policy and environment are different from that of training in the following aspects. First, for all methods, the robot arm moves slower than that during training in the environment. This helps form stable grasps. Second, for our method, the evaluation policy uses a lower temperature ($\tau_\text{test}=0.002$) than training. After a failure grasp, ours performs $2$ SGD steps on this failure experience. The network weight will be reloaded after recovery from the failure~\cite{synergy}. For the baselines, the evaluation policy uses a greedy policy. After a failure grasp, baselines perform $8$ RAD SGD steps on this failure experience. The network weight will be reloaded after recovery from the failure~\cite{synergy}.

\end{document}